%% file: SSRN_2026.tex
\documentclass[msom,nonblindrev]{informs4}

\usepackage{graphics,booktabs,color,enumerate,subfigure}
\usepackage{xcolor}
\usepackage{amssymb,amsmath,bm}
\usepackage{mathrsfs}
\usepackage{algorithm}
\usepackage{subfigure}
\usepackage{algorithmicx}
\usepackage{algpseudocode}
\definecolor{DarkBlue}{rgb}{0,0.08,0.45}
\usepackage[backref = false, bookmarks, colorlinks = true, plainpages = false, hypertexnames = false, citecolor = DarkBlue , urlcolor = DarkBlue, filecolor = DarkBlue, linkcolor = DarkBlue]{hyperref}
\usepackage{natbib}
\usepackage{pgfplots}
\usepackage{tikz}
\usepackage{setspace}
\usepackage{clipboard}
\newclipboard{myclipboard}
\OneAndAHalfSpacedXI

\makeatletter
\let\@currsize\normalsize
\makeatother
\bibpunct[, ]{(}{)}{,}{a}{}{,}%
\def\bibfont{\small}%

\newcommand{\mbe}{\mathbb{E}}
\newcommand{\E}{\mathbb{E}}
\newcommand{\mbp}{\mathbb{P}}

\newcommand{\R}{\mathbb{R}}

\newcommand{\Nscr}{\mathcal{N}}

\newcommand{\ep}{\epsilon}

\newcommand{\Var}{\mathrm{Var}}

\newcommand{\mI}{\mathbb I}

\newcommand{\diag}{\mbox{diag}}
\newcommand{\bmu}{\bm\mu}
\newcommand{\bsig}{\bm\Sigma}
\newcommand{\bth}{\bm\theta}

\newcommand{\dv}{\mathrm{d}\bm v}

\newcommand{\blue}[1]{\textcolor{black}{#1}}

\newcommand{\bA}{\bm A}
\newcommand{\bXn}{\bm X_n}

\newcommand{\Lscr}{\mathcal{L}}

\usepackage{nameref}

\definecolor{darkorange}{RGB}{215,95,0}

\usepackage[draft]{changes}   
\definechangesauthor[name={Fanni}, color=blue]{sfn}
\definechangesauthor[name={Fanni}, color=darkorange]{sfn2}
\definechangesauthor[name={Jin}, color=red]{jin}
\setauthormarkuptext{name}
\setauthormarkup{\textsuperscript{#1}}     
\newenvironment{sfnadded}{
    \color{black}
}{
    \normalfont\color{black}
}

\newenvironment{lightgray}{
    \color{black}
}{
    \normalfont\color{black}
}

\usepackage{natbib}
\bibpunct[, ]{(}{)}{,}{a}{}{,}%
\def\bibfont{\small}%
\theoremstyle{TH}%
\newtheorem{thm}{Theorem}
\newtheorem{lem}{Lemma}
\newtheorem{prop}{Proposition}

\newtheorem{defin}{Definition}

\theoremstyle{EX}
\newtheorem{rem}{Remark}
\newtheorem{assumption}{Assumption}
\newtheorem{exmp}{Example}

\usepackage{array}
\newcolumntype{L}[1]{>{\raggedright\let\newline\\\arraybackslash\hspace{0pt}}m{#1}}
\newcolumntype{C}[1]{>{\centering\let\newline\\\arraybackslash\hspace{0pt}}m{#1}}
\newcolumntype{R}[1]{>{\raggedleft\let\newline\\\arraybackslash\hspace{0pt}}m{#1}}

\ECRepeatTheorems

\EquationsNumberedThrough    

\MANUSCRIPTNO{}

\usepackage[most]{tcolorbox}
\usepackage{tikz}
\usetikzlibrary{arrows, positioning}
\tikzset{
	>=stealth',
	banki/.style={
		rectangle,
		rounded corners,
		draw=black, very thick,
		text width= 3.5em,
		minimum height=2em,
		text centered},
	debti/.style={
		->,
		thick,
		shorten <=2pt,
		shorten >=2pt,}
}
\newtcolorbox{myquote}[1][]{%
    fontupper={\OneAndAHalfSpacedXI},
    colback=black!5,
    colframe=black!5,
    notitle,
    sharp corners,
    borderline west={2pt}{0pt}{red!80!black},
    enhanced,
    breakable,
    coltext=blue,
}

\begin{document}


\ARTICLEAUTHORS{%
\AUTHOR{Ningyuan Chen\textsuperscript{1}, Setareh Farajollahzadeh\textsuperscript{2}, Qingwei Jin\textsuperscript{3}, Fanni Shen\textsuperscript{3}, Guan Wang\textsuperscript{1}}
\AFF{\textsuperscript{1} University of Toronto; \textsuperscript{2} McGill University; \textsuperscript{3} Zhejiang University}
}

\TITLE{Learning Customer Preferences from Bundle Sales Data}

\ABSTRACT{
\emph{Problem definition:} \Copy{rev:writing0}{
This paper studies the problem of estimating consumer preferences from bundle sales data. Product bundling is a widely used pricing strategy in retail markets.
}
To set profitable bundle selection and prices, the seller needs to learn the distribution of consumers' valuations for individual products from the transaction data. When customers purchase bundles or multiple products, classical methods such as discrete choice models cannot be used to estimate consumers'
valuations. 
In this paper, we propose an approach to learn the distribution of consumers' valuations toward the products using bundle sales data. 
\emph{Methodology/results:} Our approach is to define a utility model for customer choices and estimate the parameters of a valuation distribution that maximizes the likelihood of observing the transaction data. Our approach reduces this problem to an estimation problem where the samples are censored by polyhedral regions on the valuation space of customers. Using the EM algorithm and Monte Carlo simulation, our approach can recover the distribution of consumers' valuations. We extend the framework to allow for unobserved no-purchases, clustered market segments
\blue{and to incorporate non-additive bundle utilities with synergy effects.}
In addition, we provide theoretical results on the identifiability of the probability model and \blue{sufficient conditions for local convergence} of the EM algorithm. Moreover, the performance of the approach is also demonstrated numerically with synthetic and real datasets. 
\emph{Managerial implications:} This study demonstrates the challenge to leverage the transaction data of bundle sales to learn customers' preferences. The proposed algorithm provides a practical guidance for retailers.}
\maketitle


\input{1_Intro_LR_Problem}
\input{2_Theoretical}
\input{3_Extensions}
\input{4_Numerical_Base}
\input{4_Numerical}
\section{Conclusion}

We investigate the practical problem of estimating customers' valuations in the presence of bundle sales promotions, which is essential for designing bundle sales promotions and pricing strategies. We formulate this problem as a statistical problem of fitting a distribution (multivariate Gaussian distribution in the base model) when observations are censored within various partitions, and we propose an Expectation-Maximization algorithm. We study the properties of the problem and show that under separate selling and offering each single product at two different prices, the estimation problem is identifiable. We also provide sufficient conditions under which the population EM operator converges locally to the true parameters when initialized within a basin of attraction.
\blue
{Building upon the base model, we develop a Product Synergy model to capture complementarity and substitutability among products within a bundle, allowing the utility of a bundle to depend not only on the valuations of its constituent products but also on their interactions. }
Furthermore, we extend the base model to censored-demand settings, where no-purchase observations may be unobserved, and to Gaussian mixture models that capture latent customer segments. We demonstrate the efficacy of our algorithm through synthetic numerical experiments and sales data from JD.com.

However, there are still unexplored problems in the literature, such as considering competitive prices when estimating the demand distribution given that the sales data of the competitor is censored for one's sales data. Additionally, statistical questions related to calculating the error of Monte Carlo samples and the convergence error of the EM algorithm need further exploration. Moreover, identifying the number of components when performing the Gaussian Mixture model is an interesting direction for future research.

%

\bibliographystyle{informs2014}
\bibliography{Reference_Bundle}
\include{Appendix_Bundle}
\include{EC_Bundle}

\hypertarget{loc}{}

\end{document}

%% file: 1_Intro_LR_Problem.tex
\section{Introduction}
Bundles are pervasive in online retailing, fast-food restaurants, insurance, airlines, and the video game industry.
Firms often combine products or services into discounted packages to stimulate demand and increase transaction value.
Examples range from McDonald's combo meals and bundled home-and-auto insurance policies to airline packages that combine fares, baggage allowances, and meals.
Due to the popularity of bundles, how to set profitable bundle pricing strategies has been a focus of the literature since \cite{stigler1963united}.
A number of simple bundle pricing strategies have shown promising theoretical and empirical performance, such as grand bundles (bundling all products together at a discount) and bundle size pricing (pricing the bundle based on its size only). See \citet*{chu2011bundle} for a comprehensive empirical comparison.

A prerequisite for successful bundle pricing is an accurate estimate of consumers' reservation prices, or valuations.
These valuations determine how consumers respond to menus of individual products and bundles, and hence determine aggregate demand.
A natural source for learning such information is the firm's historical transaction data.
However, compared to the number of studies on how to set prices for bundles assuming \emph{known} market demand or valuation distributions, there is scant literature on how to learn the valuations from the transaction data of bundle sales.
This is a stark contrast to the industry practice: according to \cite{fisher2020empirical}, companies make more than 90\% of the effort estimating demand and 10\% setting optimal prices.

\Copy{rev2:writing}{
{This study provides a data-driven framework for learning consumer valuations from the transaction data of bundle sales, which can be used to inform and support the development of bundle pricing strategies.}
}
The firm observes the behavior of past customers: the products and bundles that are offered at the time, the charged prices, and the final purchase decisions. Based on the information, the firm attempts to learn the valuations of the customers in the data to forecast the shopping behavior of future customers.

{
    The challenges of this learning problem arise from missing valuations, censored demand, and the bundled products. We summarize each of these challenges and explain why existing approaches are insufficient.
    First, customers’ valuations for products are only partially observed in transaction data. While purchase decisions are recorded, the underlying willingness-to-pay (WTP) is not directly observed. Instead, each observed purchase provides only a lower bound on the customer’s valuation: if a customer buys a product at price \$5, the firm can infer that the customer’s valuation is at least \$5, but the exact valuation remains unknown. This leads to a form of interval-censored or partially identified data, which complicates the estimation of the valuation distribution.
    Second, the data suffer from demand censoring, as customers who do not make any purchase are typically not observed. As a result, the dataset over-represents purchasing customers and omits those with low valuations, introducing selection bias into the estimation problem.
    The first two challenges are common in estimation problems based on transaction data. In settings without bundles, discrete choice models such as the multinomial logit (MNL) model address these issues by linking observed purchase probabilities to model parameters, thereby bypassing the need to observe valuations directly.
    However, the presence of bundles introduces an additional layer of difficulty. A common simplification in the literature is to treat each bundle as an independent product and apply standard discrete choice models. While this approach enables tractable estimation, it ignores the structural relationship between bundles and their component products. This simplification is often adopted precisely because bundles introduce additional complexity: when products are sold jointly, their valuations become entangled, and it is no longer clear how to attribute the observed purchase decision to the valuation of individual products. As a result, the firm cannot separately identify product-level valuations using standard methods.
    These challenges highlight the limitations of existing approaches and motivate the need for a framework that can jointly handle partial observability, censoring, and the structural dependence induced by bundle offerings.
}
{
In this paper, we develop a parsimonious yet expressive model of consumer valuations and propose an estimation procedure based on transaction data. In particular, we assume that consumers' valuations for products follow a multivariate Gaussian distribution, and that the valuation of a bundle is given by the sum of the valuations of its component products. This additive utility specification is standard in the bundle pricing literature and provides a tractable foundation for analysis.
Building on this formulation, we cast the estimation problem as a maximum likelihood problem with incomplete data, where individual-level valuation realizations are unobserved. To address this challenge, we design an estimation procedure that integrates the expectation–maximization (EM) algorithm with Monte Carlo simulation, enabling the recovery of the mean vector and covariance matrix of the underlying valuation distribution.
}
The contribution of the paper is threefold.
\begin{itemize}
    \item We reformulate bundle transaction data as a missing-data estimation problem over IC polyhedra.
    Each purchase observation does not reveal the customer's valuation vector; instead, it identifies a polyhedral region, defined by linear IC inequalities, in which that vector must lie.
    A special case of this problem, interval-censored data, is commonly seen in survival analysis \citep{lindsey1998methods}.
    Our formulation generalizes the interval to polyhedra in high dimensions.
    \item The framework and algorithm we propose are flexible enough to accommodate many practical considerations.
    \blue{We extend the framework to non-additive product valuations in bundles, in order to capture complementarity and substitutability.}
    The model also allows for Gaussian mixture models (GMM) and thus clustered valuations due to market segmentation.
    Our framework can tackle censored demand.
    \Copy{rev:synthetic_datasets}{
    \blue{Computationally, we demonstrate the algorithm on synthetic datasets with
    {up to $6$ products and $10,000$ transactions, across different product categories, sample sizes, and bundle discounts.}
    }
    }
    This is because Monte Carlo simulation allows us to provide closed-form updates in the \textit{M-step}.
    No known algorithms can handle the estimation problem for bundle sales of this scale.
    \item We study two theoretical questions: when the model is identifiable, and under what conditions the EM algorithm converges to the true parameters.
    Both questions have not been answered before for the estimation problem with bundle sales.
    For the first question, we identify a simple pricing policy used by many firms that leads to identifiable transaction data.
    \Copy{rev:technical-contribution}{For convergence, we adapt the framework of \citet*{balakrishnan2017statistical} and show that, in a population setting with known covariance and a common IC partition, the EM operator is locally contractive around the true mean under a nondegeneracy condition.}
\end{itemize}

\section{Literature Review}\label{sec:literature}
Bundle pricing mechanisms have been studied extensively in the literature,
including their theoretical and computational properties.
We list a number of representative papers below.
\cite*{bakos1999bundling,abdallah2021large} analyze the asymptotic performance of pure bundling and bundle size pricing.
\cite*{chu2011bundle} conduct a comprehensive numerical experiment to compare the performance of four commonly studied bundle pricing mechanisms.
\cite*{hanson1990optimal,wu2008customized,jiang2011optimizing,li2020convex} provide mixed-integer programming or convex optimization formulations to compute the optimal bundle pricing policies.
\cite*{hart2017approximate,babaioff2020simple} derive theoretical performance bounds for simple mechanisms.
A few recent studies \citep*{ma2021reaping,sun2021product,chen2022component, fattahi2022customer} propose new implementable pricing mechanisms and analyze their properties.
These papers usually rely on one crucial assumption: the distribution of consumers' valuations is known or a number of realized samples are given.
In practice, however, such information needs to be learned from the sales data.
Our work provides a framework to achieve this goal and further justifies the practical feasibility of the pricing mechanisms studied in these papers.
As pointed out by \cite*{fisher2020empirical}, the estimation of demand (customers' valuations) is $90\%$ of the work in practice.

Our work is related to the stream of literature that focuses on the estimation of discrete choice models, especially the well-known multinomial logit (MNL) model, from transaction data.
\cite{kok2008assortment} provide a comprehensive literature review of the earlier studies on such models in revenue management.
Building on \cite{talluri2004revenue}, \citet{vulcano2012estimating} combine the MNL model with a non-homogeneous Poisson arrival process over multiple periods and use the EM algorithm to estimate model parameters as well as the number of no-purchases.
\cite{newman2014estimation} incorporate price and product features into their estimation approach.
\cite{abdallah2021demand} study the identifiability of this type of model and propose a computational approach to optimize the likelihood function.
\cite{subramanian2021demand} use a mixed-integer program to estimate the parameters under the loss-minimization objective function.
The estimation of other types of discrete choice models is also studied, including the rank-based choice model \citep*{farias2013nonparametric,van2015market}
and the Markov chain choice model \citep*{csimcsek2018expectation}.
Our contribution relative to this stream is to model transaction data in which customers may choose discounted bundles whose utilities are linked through shared component products.
This setting is not directly captured by standard discrete-choice specifications that treat alternatives as unrelated products.
We therefore develop an estimation framework for product-level valuations under bundle menus.

This study proposes a framework to estimate customer valuations from transaction data for multiple products when the firm may offer a price menu for products and bundles.

\Copy{rev:Hierarchical_Bayesian}{\blue
{A prominent stream of the marketing literature models bundle choice using hierarchical Bayesian approaches. Early work by \citet{Farquhar_Rao_1976} provides a theoretical foundation for modeling bundle preferences, emphasizing complementarity and substitutability across products. Building on this foundation, \citet{Bradlow_Rao_2000} introduce Bayesian estimation into bundle choice models, allowing for consumer heterogeneity. Subsequent work by \citet{Chung_Rao_2003} develops a full hierarchical Bayesian framework in which consumer valuations are modeled as draws from a multivariate distribution, enabling flexible modeling of both heterogeneity and correlation across products.
\cite{jedidi2003measuring} study a similar problem and propose a hierarchical Bayesian approach to infer the parameters of a multivariate Gaussian distribution that is used to model the valuations for the products.
Compared with this line of literature, our paper focuses on transaction data rather than survey or stated-preference data.
We develop a likelihood-based algorithm and study theoretical questions such as identifiability, whereas the aforementioned papers primarily use Bayesian approaches for computation.
In the numerical experiments, we compare our method with a Bayesian implementation based on \cite{jedidi2003measuring} and demonstrate the strong performance of our approach.
}}

In another related paper, \cite{letham2014latent} approximate the polyhedral region with rectangular regions, which allows them to simplify the likelihood function.
In contrast, we use the EM algorithm to solve the original estimation problem without approximations.
\cite*{ma2016constructing} consider a slightly different setting: the bundle discount is only offered when a customer purchases all products.
They do not assume a parametric form for the valuation distribution and focus on a
few quantities of interest, in order to reconstruct linear demand curves of each product.
They show their estimators are consistent.
Our setting assumes the Gaussian distribution and allows for an arbitrary price menu.
In addition to the different setups and methodologies,
we extend our approach to the censored demand (unobserved no-purchases).
This is not considered in \cite{jedidi2003measuring,letham2014latent,ma2016constructing}.

The methodology used in this paper, the EM algorithm combined with Monte Carlo simulation, has been used in previous studies, including \citet{natarajan2000monte,lee2012algorithms}.
In particular, Monte Carlo simulation is used to approximate the E-step when the expectation does not have a closed form.
Because our study is focused on a specific practical problem,
we provide the concrete steps of the EM algorithm and the Monte Carlo simulation.
Moreover, the theoretical properties are derived for the specific application.
\Copy{rev:EM-2017paper}{In addition to the convergence to \blue{a stationary point} of the sample likelihood function shown in the seminal papers \cite{dempster1977maximum,wu1983convergence},  \citet*{balakrishnan2017statistical} provide conditions and statistical guarantees for the EM algorithm to converge to the global maximum corresponding to the true parameters, when initialized in a neighborhood.
\blue{Note that the classical results in \cite{dempster1977maximum,wu1983convergence} only guarantee the convergence to a stationary point, but do not characterize the ``basin of attraction'' of the initial point of the EM algorithm.
In other words, under the classical results, the EM algorithm is not guaranteed to converge to the true parameters even when the initialization is arbitrarily close.
\cite{balakrishnan2017statistical} characterize the basin of attraction of the true parameters with a set of technical conditions.
However, the conditions are abstract
and not directly applicable to a specific problem.
}
We adapt the conditions proposed by \citet{balakrishnan2017statistical} to the bundle setting.}

\section{Bundle Sales Data and Estimation}\label{sec:base-model}
In this section, we first introduce the format of the dataset recording the bundle transactions.
Suppose there are $I$ products and we use $[I]\triangleq \left\{1,\dots,I\right\}$ to denote the product space.
There are $J$ candidate bundles offered on the menu and we use $j\in \left\{0,1\right\}^I$ to denote a bundle.
That is, product $i\in [I]$ is included in bundle $j$ if and only if $j_i = 1$.
Note that not all the bundles are offered to each customer and a bundle may include only one product.
With slight abuse of notation,
we also use $j$ as the bundle index in the menu $[J]\triangleq \left\{1,\dots,J\right\}$ and $i\in j$ when bundle $j$ includes product $i$.

The dataset records the choices of $N$ customers:
customer $n\in [N]\triangleq \left\{1,\dots,N\right\}$ chooses alternative $c_n\in [J]\cup\left\{0\right\}$,
where $c_n=0$ denotes no purchase.
The case of censored demand is studied in Section~\ref{sec:censored-demand}.
Moreover, the price of bundle $j$ faced by customer $n$ is denoted by $p^j_n$, which is also recorded in the dataset.
Therefore, the dataset consists of the tuples $\{(p^1_n,\dots,p^J_n,c_n)\}_{n=1}^N$.
We do not require every bundle in the menu to be offered to every customer.
When a customer is only offered a subset of the available bundles, we can artificially set the prices of bundles that are not offered to infinity.

To be able to utilize the dataset and estimate customer preferences, we specify a utility model.
Suppose customer $n$ is endowed with a vector of product valuations, $\bm v_n=(v_{n1},\dots,v_{nI})$.
Under additive utility, the valuation of bundle $j$ for customer $n$ is $\sum_{i\in j} v_{ni}$.
Note that additive utility is a standard assumption in the literature (e.g., see \citealt{chu2011bundle} and references therein), although other types of valuation functions have been proposed and studied recently \citep{tulabandhula2020multi}.
Under additive utility, statistical dependence across products can be captured through correlations in product valuations.
\blue{In Section~\ref{sec:synergy}, we introduce a more general non-additive model with product synergy, which directly captures complementarity and substitutability within bundles.}

Under this utility model, the choice $c_n=j$ for $j\neq 0$ corresponds to the following region of the valuation vector $\bm v$, given by the incentive-compatible constraints:
\begin{equation}\label{eq:cn=j}
c_n=j\iff \left\{\sum_{i\in j}v_i-p^{j}_n\ge \max_{j'\in [J]}\sum_{i\in j'}v_i-p^{j'}_n ,\;\sum_{i\in j}v_i-p^{j}_n\ge 0\right\}.
\end{equation}
Similarly, we have
\begin{equation}\label{eq:cn=0}
c_n=0 \iff\left\{\max_{j'\in [J]}\sum_{i\in j'}v_i-p^{j'}_n\le 0 \right\}.
\end{equation}
Suppose the random valuation $\bm v\sim \bm V$ of each customer is independently drawn from an $I$-dimensional multivariate normal distribution:
\begin{equation}\label{eq:pdf-normal}
    \bm V\sim \Nscr(\bm \mu,\bm \Sigma).
\end{equation}
Note that we focus on the somewhat restrictive normal distribution for the purpose of exposition.
In Appendix~\ref{app:GMM},
we relax it to a Gaussian mixture model, which is flexible enough to approximate any probability distributions (for example, see \citealt[p.65]{goodfellow2016deep}).

\begin{rem}\label{rem:outside_option}
\Copy{rev:outside_option}{
\blue{In our formulation, the utility of the outside option is normalized to $u_0 = 0$ without randomness. This specification is standard to resolve \emph{shift invariance} in random utility models. In the Multinomial Probit (MNP) model, for a model with $I$ products and an outside option, a full $(I+1)$-dimensional covariance matrix is not identifiable from observed choices.
The standard econometric resolution is to difference all utilities against the outside option, such that the random shock of the ``no-purchase" decision is effectively absorbed into the $I$-dimensional differenced covariance structure. Our model follows this idea by parameterizing only the product utilities. Consequently, the estimated $I$-dimensional covariance matrix $\Sigma$ fully captures the relative uncertainty of choosing a product or bundle over the outside option.
This treatment is also adopted in, e.g., \cite{jedidi2003measuring}.}
}
\end{rem}

Given the dataset $\{(p^1_n,\dots,p^J_n,c_n)\}_{n=1}^N$, the firm's objective is to estimate the distribution of customer valuations, in particular $\bm \mu$ and $\bm \Sigma$,
with the knowledge that the incentive-compatible constraints~\eqref{eq:cn=j} and~\eqref{eq:cn=0} relate the valuations to the observations in the dataset.
Note that because the parameters $\bm \mu$ and $\bm \Sigma$ are linked to the observed data only indirectly through the IC constraints, this estimation problem is not standard in the statistics literature.
For example, if $\bm v_n$ were observed, then the estimation of $\bm \mu$ and $\bm \Sigma$ would be trivial using the empirical average and covariance matrix of $\{\bm v_n\}_{n=1}^N$.
Next we explain how the problem can be viewed as an estimation problem with missing data and present an algorithm for the problem.

\subsection{An Estimation Problem with Missing Data}
The key observation to simplify the estimation problem is that given $(p^1_n,\dots,p^J_n,c_n)$,
the incentive-compatible constraints \eqref{eq:cn=j} and \eqref{eq:cn=0} specify a polyhedron in which the valuation vector $\bm v_n$ falls in.
More precisely, define
\begin{align}\label{eq:ic-polytope}
    R^j_n&\triangleq \left\{\bm v: \sum_{i\in j}v_i-p^{j}_n\ge \max_{j'\in [J]}\sum_{i\in j'}v_i-p^{j'}_n ,\;\sum_{i\in j}v_i-p^{j}_n\ge 0\right\}, \quad j\neq 0,\notag\\
    R^0_n&\triangleq \left\{\bm v: \max_{j'\in [J]}\sum_{i\in j'}v_i-p^{j'}_n\le 0 \right\},
\end{align}
which we refer to as \emph{IC polyhedra}.
The observation $c_n=j$ is equivalent to $\bm v_n\in R^j_n$.
As a result, for customer $n$, the valuation space $\R^I$ can be partitioned into the $J+1$ regions: $\R^I = \cup_{j=1}^J R^j_n \cup R^0_n$.
The observation $(p^1_n,\dots,p^J_n,c_n)$ characterizes the partition as well as the IC polyhedron in which $\bm v_n$ falls.
Therefore, instead of observing $\bm v_n$ exactly, the firm observes a polyhedron $R_n^{c_n}$ for all potential $\bm v_n$ that is consistent with the observation.
This is sometimes referred to as ``censored'' observations in statistics.
To avoid ambiguity, we do not use the term and reserve it for demand censoring, which is the subject of Section~\ref{sec:censored-demand}.
With this interpretation, we view the estimation problem through the lens of inference with missing data.
\begin{description}
    \item[Parameters:] we are interested in estimating $(\bm \mu, \bm \Sigma)$, the distribution of the valuations for the products among consumers.
    By convention, we use $\bm\theta\triangleq (\bm \mu, \bm \Sigma)$ to denote the parameters.
    \item[Observed data:] the firm observes the polyhedra $\left\{R_n^{c_n}\right\}_{n=1}^N$ given the data $\{(p^1_n,\dots,p^J_n,c_n)\}_{n=1}^N$ in \eqref{eq:ic-polytope}.
        We use {$\bm D \triangleq \left\{R_n^{c_n}\right\}_{n=1}^N$} for the observed data.
        Note that by converting the data to polyhedra, we do not lose any information regarding the possible valuations $\left\{\bm v_n\right\}_{n=1}^N$.
    \item[Missing data:] we treat valuations $\bm v_n$, $n=1,\dots,N$, as the missing data.
        They are related to the observed data by the simple fact that $\bm v_n\in R_n^{c_n}$.
        We use $\bm Z\triangleq\{\bm v_n\}_{n=1}^N$ to denote the missing data.
\end{description}
There are two common approaches to missing data: imputation and maximum likelihood estimation (MLE).
Since our estimation problem is model-based, we adopt the latter.
In particular, the likelihood function based on the observed data is:
$\Lscr(\bm\theta;\bm D) = \int p(\bm D, \bm Z|\bm \theta)\mathrm{d}\bm Z = \prod_{n=1}^N\int_{R^{c_n}_n} f(\bm v|\bm\mu,\bm\Sigma) \mathrm{d}\bm v$,
where
$f(\bm v|\bmu,\bsig) = \tfrac{1}{\sqrt{(2\pi)^I|\bsig|}}\exp\left(-\frac{1}{2}(\bm v-\bmu)^\top \bsig^{-1}(\bm v-\bmu)\right)$
is the probability density function (PDF) of the multivariate normal distribution \eqref{eq:pdf-normal}.

\Copy{rev:directmax}{
\blue{
A natural approach would be to directly maximize the likelihood function over $\bth$.
However, in our setting, the likelihood involves truncated high-dimensional Gaussian integrals over polyhedral regions defined by the IC constraints.
These integrals do not admit closed-form expressions and have to be approximated via Monte Carlo sampling.
The resulting Monte-Carlo-approximated objective function is discontinuous.
Its integer programming formulation is challenging for standard solvers even for a small-scale problem.
}}
Taking logarithms improves numerical scaling but does not remove the main difficulty: the objective still contains Gaussian integrals over IC polyhedra, which do not have closed-form expressions in general.
As a popular alternative, the EM algorithm iteratively maximizes the likelihood function by focusing on the \emph{complete-data} likelihood. Next, we elaborate on the implementation of the EM algorithm in this problem.

\subsection{The EM Algorithm} \label{sec: The_EM_Algorithm}
Instead of focusing on the (log-)likelihood function of the observed data with missing values,
the EM algorithm focuses on the complete-data log-likelihood function, i.e.,
\begin{equation}\label{eq:complete-data-llhx}
    \ell (\bth;\bm D, \bm Z) = \sum_{n=1}^{N} \log f(\bm v_n|\bm \mu,\bm \Sigma).
\end{equation}
Note that as an objective function,
{$\ell(\bth;\bm D, \bm Z)$} is easier to handle than
{$\ell(\bth;\bm D)$}.
Given complete valuation data, the Gaussian log-likelihood has standard closed-form maximizers for $\bmu$ and $\bsig$.
However, $\bm Z$ is not observed in the transaction data.

The EM algorithm circumvents the issue by estimating $\bth^{(t)}$ iteratively.
In each iteration, suppose $\bth^{(t)}=\left(\bm \mu^{(t)},\bm \Sigma^{(t)}\right)$ is given.
We first evaluate the expected complete-data log-likelihood over $\bm Z$, i.e.,
{
\begin{align}\label{eq:Q-func}
    Q\left(\bm \theta|\bm \theta^{(t)}\right)&\triangleq \E\left[\ell(\bm\theta;\bm D,\bm Z)\bigg|\bm D,\bm \mu^{(t)},\bm \Sigma^{(t)}\right]\notag\\
                                  &= \sum_{n=1}^{N} \int_{\bm v\in R_n^{c_n}} \frac{f\left(\bm v|\bm \mu^{(t)},\bm \Sigma^{(t)}\right)}{\int_{R_n^{c_n}}f\left(\bm \xi|\bm \mu^{(t)},\bm \Sigma^{(t)}\right)\mathrm{d}\bm \xi }\log f(\bm v|\bm \mu,\bm \Sigma)\mathrm{d}\bm v.
\end{align}
}
This is the so-called \emph{E-step} and the expectation is taken with respect to the missing data $\bm Z$.
Given $\left(\bm D,\bmu^{(t)},\bsig^{(t)}\right)$, the missing variable $\bm v_n$ follows a multivariate normal distribution with mean $\bmu^{(t)}$ and covariance $\bsig^{(t)}$, truncated to the polyhedron $R_n^{c_n}$.
Its PDF is thus $f\left(\bm v|\bm \mu^{(t)},\bm \Sigma^{(t)}\right)/\int_{R_n^{c_n}}f\left(\bm \xi|\bm \mu^{(t)},\bm \Sigma^{(t)}\right)\mathrm{d}\bm \xi $ for $\bm v\in R_n^{c_n}$, which is the first term in the integral.
The second term $\log f(\bm v|\bm \mu,\bm \Sigma)$ is the log-likelihood function
{$\ell(\bm\theta;\bm D,\bm Z)$} and is integrated with the conditional PDF for its expected value.


To update $\bth$ in the $t$-th iteration, the \emph{M-step} maximizes $Q\left(\bth|\bth^{(t)}\right)$ and sets
$\bth^{(t+1)}=\argmax_{\bth}Q\left(\bth|\bth^{(t)}\right)$.
The main difficulty is that the truncated-normal expectations in \eqref{eq:Q-func} do not have closed forms.
To deal with this challenge, we use Monte Carlo simulation.
More precisely, for all $n=1,\dots,N$, we generate $L$ samples according to the same distribution as $\bm v|R_n^{c_n},\bmu^{(t)},\bsig^{(t)}$.
This is the conditional multivariate normal distribution inside $R_{n}^{c_n}$.
The standard acceptance-rejection method can be used and we provide the details in
Algorithm~\ref{alg:acc-rej}.
\begin{algorithm}[t!]
    \caption{Acceptance-rejection method}\label{alg:acc-rej}\blue{
    \begin{algorithmic}[1]
        \State \textbf{Input:} Feasible region $R_n^{c_n}$, current parameters $\bmu^{(t)}, \bsig^{(t)}$, number of samples $L$
        \For{$l = 1$ to $L$}
            \Repeat
            \State Draw candidate $\bm v \sim \Nscr(\bmu^{(t)}, \bsig^{(t)})$
            \Until{$\bm v \in R_n^{c_n}$}\label{step:acceptance} \Comment{Accept only if $\bm v$ lies in the feasible region}
            \State $\bm v_n^{(l)} \gets \bm v$
        \EndFor
        \State \Return $\left\{\bm v_n^{(1)}, \dots, \bm v_n^{(L)}\right\}$
    \end{algorithmic}}
\end{algorithm}

Once the Monte Carlo samples {$\{\bm v_{n}^{(l)}\}$ for $n=1,\dots,N$ and $l=1,\dots,L$} have been generated,
we may replace $Q(\bth|\bth^{(t)})$ in \eqref{eq:Q-func} by
{
\begin{align}\label{eq:approx-Q}
    \hat Q(\bm \theta|\bm \theta^{(t)})&= \frac{1}{L}\sum_{n=1}^{N} \sum_{l=1}^L \log f(\bm v_{n}^{(l)}|\bm \mu,\bm \Sigma)\notag\\
                                       &= C+ \frac{1}{L}\sum_{n=1}^N\sum_{l=1}^L \left(- \frac{1}{2}\left(\bm v_n^{(l)}-\bmu\right)^\top \bsig^{-1}\left(\bm v_{n}^{(l)}-\bmu\right) - \frac{1}{2}\log |\bsig| \right),
\end{align}}
where $C$ is constant with respect to $\bmu$ and $\bsig$.
Note that this is the standard MLE for multivariate normal distribution with $NL$ samples.
Therefore, we have
{
\begin{align}
    \bmu^{(t+1)} &=  \frac{1}{NL}\sum_{n=1}^N\sum_{l=1}^L \bm v_{n}^{(l)},\label{eq:update_mu_t1}\\
    \quad\bsig^{(t+1)} &= \frac{1}{NL} \sum_{n=1}^N\sum_{l=1}^L \left(\bm v_n^{(l)}-\bmu^{(t+1)}\right)\left(\bm v_n^{(l)}-\bmu^{(t+1)}\right)^{\top}\label{eq:update_sig_t1}.
\end{align}
    }
This completes the $t$-th iteration of the EM algorithm.

The EM algorithm is terminated when the estimation $\bth^{(t)}$ has converged.
In practice, we may impose a small tolerance level $\epsilon>0$ and terminate the algorithm once $\|\bth^{(t+1)}-\bth^{(t)}\|\le \epsilon$ for a chosen norm $\|\cdot\|$.
We summarize the steps in Algorithm~\ref{alg:em-basic}.
\begin{algorithm}
    \caption{The EM algorithm}\label{alg:em-basic}
    \blue{
    \begin{algorithmic}[1]
        \State \textbf{Input:} Observed data $\{(p^1_n,\dots,p^J_n,c_n)\}_{n=1}^N$, tolerance $\epsilon$, Monte Carlo sample number $L$, initial parameters ($\bmu^{(0)}$, $\bsig^{(0)}$)
        \State Construct $\{R_n^{c_n}\}_{n=1}^N$ using \eqref{eq:ic-polytope}
        \State $error\gets \infty$, $t\gets 0$
        \While{$error>\epsilon$}\label{step:em-iteration}
            \State Run Algorithm~\ref{alg:acc-rej} for $\bmu^{(t)}, \bsig^{(t)},\left\{R_n^{c_n}\right\}_{n=1}^N$
            \State Update $\bmu^{(t+1)}$ and $\bsig^{(t+1)}$ using \eqref{eq:update_mu_t1} and \eqref{eq:update_sig_t1}
            \State $error\gets \|\bmu^{(t+1)}-\bmu^{(t)}\|_1+\|\bsig^{(t+1)}-\bsig^{(t)}\|_1$, $t\gets t+1$
        \EndWhile
        \State \Return $\bmu^{(t)}, \bsig^{(t)}$
    \end{algorithmic}
    }
\end{algorithm}


\textbf{Sampling efficiency of Monte Carlo simulation.}
In practice, the major computational complexity of Algorithm~\ref{alg:em-basic} is caused by the Monte Carlo simulation.
when the region $R_n^{c_n}$ is small or distant from $\bmu^{(t)}$, Algorithm~\ref{alg:acc-rej} may be inefficient as it requires a large number of samples in Step~\ref{step:acceptance}  to draw an accepted sample in $R_n^{c_n}$.
\label{rev:willma}
\Copy{rev:willma}{
\blue{
Unlike settings with highly structured bundle discounts in \citet{ma2016constructing}, where bundle prices are additive except for a single full-bundle discount, we allow for arbitrary bundle prices. This general price menu makes the IC regions general polyhedra rather than separable item-level rectangles.
}}
\label{rev:smallJ}
\Copy{rev:smallJ}{
\blue{
    In particular, the dependence on $J$ arises through the IC constraints that define the region $R_n^{c_n}$.
    As $J$ grows, the number of affine inequalities increases, which may shrink the feasible region and reduce the acceptance rate of the sampling procedure.
    This leads to a substantial increase in computational cost, especially when $J$ grows exponentially in $I$.
}}
We provide three remedies to improve the sampling efficiency.

First, in each iteration of Algorithm~\ref{alg:em-basic} (Step~\ref{step:em-iteration}), we can generate a large number of samples from $\Nscr\left(\bmu^{(t)},\bsig^{(t)}\right)$ and store them.
When running Algorithm~\ref{alg:acc-rej} for each $R_n^{c_n}$, the algorithm can use the same set of samples for acceptance/rejection.
This saves the computation time to generate a large number of samples for each $n=1,\dots,N$.

Second, we may not require the same number of samples $L$ in all regions $R_n^{c_n}$.
For example, we may generate $L_n$ Monte Carlo samples for region $R_n^{c_n}$, choosing $L_n$ adaptively based on the acceptance rate or the empirical variance of the sampled log-likelihood terms.
As long as we use $L_n^{-1}\sum_{l=1}^{L_n} \log f(\bm v_{n}^{(l)}|\bm \mu,\bm \Sigma)$ in $\hat Q(\bm \theta|\bm \theta^{(t)})$, the Monte Carlo approximation remains unbiased for the corresponding conditional expectation.
However, small values of $L_n$ can increase Monte Carlo noise and may slow or destabilize convergence.

Third, for low-probability $R_n^{c_n}$, we may use \emph{importance sampling} to increase the sampling efficiency.
In particular, consider \eqref{eq:Q-func} and some $n$ that $f\left(\bm v|\bm \mu^{(t)},\bm \Sigma^{(t)}\right)\approx 0$ for $\bm v\in R_n^{c_n}$.
In this case, it is computationally challenging as Step~\ref{step:acceptance} is repeated many times before acceptance.
Instead, we can consider a different normal distribution $\Nscr(\bmu',\bsig')$ such that $\bm v\in R_n^{c_n}$ with high probability.
This can be achieved, for example, by choosing $\bmu'$ close to the center of $R_n^{c_n}$ when $R_n^{c_n}$ is bounded.
As a result, we have
\begin{align*}
    &\int_{\bm v\in R_n^{c_n}} \frac{f\left(\bm v|\bm \mu^{(t)},\bm \Sigma^{(t)}\right)}{\int_{R_n^{c_n}}f\left(\bm \xi|\bm \mu^{(t)},\bm \Sigma^{(t)}\right)\mathrm{d}\bm \xi }\log f(\bm v|\bm \mu,\bm \Sigma)\mathrm{d}\bm v\\
    =& \int_{\bm v\in R_n^{c_n}} \frac{f(\bm v|\bmu',\bsig')}{\int_{R_n^{c_n}}f(\bm \xi|\bmu',\bsig')\mathrm{d}\bm \xi }\frac{f\left(\bm v|\bm \mu^{(t)},\bm \Sigma^{(t)}\right)\int_{R_n^{c_n}}f(\bm \xi|\bmu',\bsig')\mathrm{d}\bm \xi }{f(\bm v|\bmu',\bsig')\int_{R_n^{c_n}}f\left(\bm \xi|\bm \mu^{(t)},\bm \Sigma^{(t)}\right)\mathrm{d}\bm \xi }\log f(\bm v|\bm \mu,\bm \Sigma)\mathrm{d}\bm v.
\end{align*}

By simulating $L$ samples from a truncated normal distribution $\Nscr(\bmu',\bsig')$ on region $R_n^{c_n}$ (run Algorithm~\ref{alg:acc-rej} with the new distribution), denoted as
{$\bm v_{n}^{(1)},\dots,\bm v_n^{(L)}$}, we can approximate the above term by
{
\begin{align}
	\frac{1}{\frac{\int_{R_n^{c_n}}f\left(\bm \xi|\bm \mu^{(t)},\bm \Sigma^{(t)}\right)\mathrm{d}\bm \xi }{\int_{R_n^{c_n}}f\left(\bm \xi|\bmu',\bsig'\right)\mathrm{d}\bm \xi }} &\frac{1}{L}   \sum_{l=1}^L\frac{f\left(\bm v_n^{(l)}|\bm \mu^{(t)},\bm \Sigma^{(t)}\right)}{f\left(\bm v_n^{(l)}|\bmu',\bsig'\right)} \log f\left(\bm v_n^{(l)}|\bmu,\bsig\right) \notag \\
    &=\frac{1}{\sum_{l=1}^L \frac{f\left(\bm v_n^{(l)}|\bm \mu^{(t)},\bm \Sigma^{(t)}\right)}{f\left(\bm v_n^{(l)}|\bmu',\bsig'\right)}}\sum_{l=1}^L\frac{f\left(\bm v_n^{(l)}|\bm \mu^{(t)},\bm \Sigma^{(t)}\right)}{f\left(\bm v_n^{(l)}|\bmu',\bsig'\right)} \log f\left(\bm v_n^{(l)}|\bmu,\bsig\right), \label{eq:imp_samp_approx}
\end{align}
}
where the equality follows from approximating the denominator outside the sum:
$$\frac{\int_{R_n^{c_n}}f\left(\bm \xi|\bm \mu^{(t)},\bm \Sigma^{(t)}\right)\mathrm{d} \bm \xi }{\int_{R_n^{c_n}}f(\bm \xi|\bmu',\bsig') \mathrm{d} \bm \xi} = \int_{R_n^{c_n}} \frac{f\left(\bm \xi|\bm \mu^{(t)},\bm \Sigma^{(t)}\right) }{f(\bm \xi|\bmu',\bsig')  } \frac{f(\bm \xi|\bmu',\bsig') }{\int_{R_n^{c_n}}f(\bm \xi|\bmu',\bsig') \mathrm{d} \bm \xi } \mathrm{d}\bm \xi = \frac{1}{L} \sum_{l=1}^L \frac{f\left(\bm v_n^{(l)}|\bm \mu^{(t)},\bm \Sigma^{(t)}\right)}{f\left(\bm v_n^{(l)}|\bmu',\bsig'\right)}.$$
{
By defining the normalized importance weights $w_{nl} \triangleq \frac{f\left(\bm v_n^{(l)}|\bm \mu^{(t)},\bm \Sigma^{(t)}\right)}{f\left(\bm v_n^{(l)}|\bmu',\bsig'\right)}/\sum_{l=1}^L \frac{f\left(\bm v_n^{(l)}|\bm \mu^{(t)},\bm \Sigma^{(t)}\right)}{f\left(\bm v_n^{(l)}|\bmu',\bsig'\right)}$,
the approximate $Q$ function \eqref{eq:Q-func} can be cast in the following form using the Monte Carlo samples, and \eqref{eq:imp_samp_approx}:
}
$\hat Q\left(\bm \theta|\bm \theta^{(t)}\right)= \sum_{n=1}^{N} \sum_{l=1}^L w_{nl}\log f\left(\bm v_{n}^{(l)}|\bm \mu,\bm \Sigma\right)
$.
The $M$-step thus leads to the following update
\begin{align}
    \bmu^{(t+1)} = & \frac{\sum_{n=1}^N\sum_{l=1}^L w_{nl}\bm v_{n}^{(l)}}{\sum_{n=1}^N\sum_{l=1}^L w_{nl}},\label{eq:update_mut1_imp}\\
    \bsig^{(t+1)} = & \frac{\sum_{n=1}^N\sum_{l=1}^L w_{nl}\left(\bm v_n^{(l)}-\bmu^{(t+1)}\right)\left(\bm v_n^{(l)}-\bmu^{(t+1)}\right)^{\top}}{\sum_{n=1}^N\sum_{l=1}^L w_{nl}}.\label{eq:update_sigt1_imp}
\end{align}
Therefore, using importance sampling results in a similar procedure to Algorithm~\ref{alg:em-basic} but reduces the computation in Algorithm~\ref{alg:acc-rej}.
We rewrite the complete EM algorithm with importance sampling in detail in Algorithm~\ref{alg:em-imp}.
\begin{algorithm}[ht]
    \caption{The EM algorithm with Importance Sampling}\label{alg:em-imp}
    \blue{
    \begin{algorithmic}[1]
        \State \textbf{Input:} Observed data $\{(p^1_n,\dots,p^J_n,c_n)\}_{n=1}^N$, tolerance $\epsilon$, Monte Carlo sample size $L$, initial parameters $(\bmu^{(0)},\bsig^{(0)})$
        \State Construct $\{R_n^{c_n}\}_{n=1}^N$ using \eqref{eq:ic-polytope}
        \State Choose the proposal {$\Nscr(\bmu'_n,\bsig'_n)$}  \Comment{the proposal distribution can be centered at the polyhedron}
        \State $\text{error} \gets \infty$, $t \gets 0$
        \While{$\text{error}>\epsilon$}
            \For{$n=1,\dots,N$} \Comment{\textbf{E-step:} generate samples}
                    \For{$l=1,\dots,L$}
                        \State Repeat $\bm v \sim \Nscr(\bmu'_n,\bsig'_n)$ until $\bm v\in R_n^{c_n}$
                        \State $\bm v_{n}^{(l)}\gets \bm v$
                    \EndFor
                \EndFor
            \State Compute normalized importance weights for each $n=1,\dots,N$, $l=1,\dots,L$,
            \begin{align*}
                   w_{nl} \gets \frac{f\left(\bm v_n^{(l)}|\bm \mu^{(t)},\bm \Sigma^{(t)}\right)}{f\left(\bm v_n^{(l)}|\bmu'_n,\bsig'_n\right)}/\sum_{l=1}^L \frac{f\left(\bm v_n^{(l)}|\bm \mu^{(t)},\bm \Sigma^{(t)}\right)}{f\left(\bm v_n^{(l)}|\bmu'_n,\bsig'_n\right)}
            \end{align*}
            \State \textbf{M-step:} Update $\bmu^{(t+1)}$ and $\bsig^{(t+1)}$ using \eqref{eq:update_mut1_imp} and \eqref{eq:update_sigt1_imp}
            \State $\text{error} \gets
                \|\bmu^{(t+1)}-\bmu^{(t)}\|_1+
                \|\bsig^{(t+1)}-\bsig^{(t)}\|_1
            $, $t \gets t+1$
        \EndWhile
        \State \textbf{Return:} $\bmu^{(t)},\bsig^{(t)}$
    \end{algorithmic}
    }
\end{algorithm}



%% file: 2_Theoretical.tex
\section{Theoretical Properties}\label{sec:theory}
In this section, we study statistical properties of the base model in Section~\ref{sec:base-model} and the associated EM algorithm.
We first establish conditions under which the valuation distribution is identifiable from bundle transaction data. We then analyze local convergence of the EM algorithm.

\subsection{Identifiability}\label{sec:ident}

\Copy{rev:typo_J_K}{
Identifiability addresses the fundamental question of whether the underlying model parameters $\bm\theta=(\bmu, \bsig)$ can be uniquely recovered from observable data.
In our context, consider a collection of regions
$\mathscr P\triangleq \{\mathcal P_1,\dots,\mathcal P_P\}$, where $\mathcal P_\rho\subset \R^I$.
We define identifiability of our model as follows.
}

\begin{defin}[{\sc Identifiability}] \label{def_iden}
Let $\mathcal{F} = \{f(\cdot|\bm \theta) :\bm  \theta \in \Theta\}$ be a family of multivariate normal distributions in $\R^I$, where
$\bm \theta = (\bmu, \bsig)$ denotes the parameter vector in the parameter space $\Theta$ and $f(\bm v|\bm\theta) = \frac{1}{\sqrt{(2\pi)^I|\bsig|}}\exp\left(-\frac{1}{2}(\bm v-\bmu)^\top \bsig^{-1}(\bm v-\bmu)\right)$. The model is identifiable over $\mathscr{P}$ if
the mapping $\Psi: \Theta \to [0, 1]^P$
defined by the vector of probabilities
$\Psi(\bm\theta) = \left( \int_{\mathcal P_1} f(\bm v|\bm \theta)\mathrm d\bm v, \dots, \int_{\mathcal P_P} f(\bm v|\bm \theta)\mathrm d\bm v \right)$
is injective.
Formally, for any $\bm\theta, \bm\theta' \in \Theta$, the model is identifiable over $\mathscr{P}$ if:
$$
\int_{\mathcal P_\rho} f(\bm v|\bth)  \dv
=
\int_{\mathcal P_\rho} f(\bm v|\bth') \dv
\quad \forall \rho \in \{1, \dots, P\}
\implies
\bth =\bth'.
$$
\end{defin}
\begin{lightgray}{
}
\end{lightgray}
Identifiability is a prerequisite for consistent parameter recovery: if there are multiple parameters generating the same distribution of the data, then the parameter cannot be estimated consistently.
\blue
{\label{rev:I=1mnp}\Copy{rev:I=1mnp}{
Note that when $I=1$, the model reduces to a classical single-product discrete-choice setting, where identification issues are well understood. }}
However, identifiability is not a technical requirement that is always satisfied in this model.
Consider the following two examples.
\begin{exmp}\label{exm:two-prod}
    The firm always bundles two products together in the price menu.
    That is, they are never sold separately.
    In this case, the model is not identifiable over the resulting IC polyhedra because it is not possible to estimate the marginal mean of either product.
\end{exmp}
\begin{exmp}\label{exm:1price-not-enough}
\Copy{rev:1price-not-enough}{
    Consider a single product $I=J=1$ always offered at price $\$10$.
    In this case,
    {the induced partition contains two intervals: $\mathcal P_1=(-\infty,10)$ and $\mathcal P_2=[10,+\infty)$.}
    The model is not identifiable because $\bm\theta_1$ and $\bm\theta_2$ satisfying $(\mu_1-10)/\sigma_1=(\mu_2-10)/\sigma_2$ always leads to
    the same probability
    {$\mbp(\mathcal P_1|\bm\theta_1)=\mbp(\mathcal P_1|\bm\theta_2)$ and $\mbp(\mathcal P_2|\bm\theta_1)=\mbp(\mathcal P_2|\bm\theta_2)$.}
    }
\end{exmp}
Therefore, we need to establish conditions on $\mathscr P$ that guarantee identifiability.
Intuitively, the model has $I+I(I+1)/2$ free parameters: $I$ mean parameters and $I(I+1)/2$ covariance parameters.
Thus, identifiability requires sufficiently many informative region probabilities. However, because these probabilities depend nonlinearly on $(\bmu,\bsig)$, it is nontrivial to verify either sufficiency or necessity from a simple counting argument.

\blue{\label{rev:easily_satisfied}\Copy{rev:easily_satisfied}{
Based on the application, we provide the following sufficient condition that is common in retail environments.
Roughly speaking, the condition states that items are frequently sold separately without bundle discounts and they are sold at least two different prices each.
This requirement is plausible in many retail settings.
Identification only requires separate-selling observations in addition to bundle sales; it does not require a market with only separate selling.
Moreover, temporary promotions, discounts, and dynamic pricing policies naturally generate price variation.
These sources of variation are common in online retail.
The JD.com dataset \citep{shen2020jd} analyzed in Section \ref{subsec:realdata} contains both separate selling and bundle purchase options.
Bundle purchases account for 25.54\% of all purchase transactions.
In terms of price variation, we find that 90\% of single products appear with multiple observed prices, consistent with frequent promotional price changes.  }}

\begin{sfnadded}
    \begin{prop}[\sc{Identifiability under separate selling}] \label{prop:identifiability}
    Let $\bm p^{(0)} = (p_1, \dots, p_I)^\top \in \R^I$ be the vector of regular separate-selling prices. Suppose that for each product $i \in \{1, \dots, I\}$, there exists at least one promotion price $p'_i \neq p_i$ such that the separate-selling price menu $\bm p^{(i)} = (p_1, \dots, p'_i, \dots, p_I)^\top$ is offered infinitely often. Then, the model is identifiable over the collection of regions $\mathscr{P}$ induced by these $I+1$ price menus.
    \end{prop}
\end{sfnadded}
\Copy{rev:seperate_selling}{
Proposition~\ref{prop:identifiability} gives a sufficient condition for identification based on separate-selling observations with price variation. The proof also explains why a single separate-selling price per product is insufficient within this identification argument.
}
Note that
it
does not require all combinations of the two prices of the products to be offered infinitely often, which would result in $2^I$ price menus.
Instead, a simple and realistic scenario with $I+1$ price menus would suffice:
a price menu of regular prices of the products $(p_1,\dots,p_I)$ and $I$ price menus in which only one product is on sale, i.e., $(p_1,\dots,p_{i-1},p_i',p_{i+1},\dots,p_I)$ for product $i$.
If the above $I+1$ price menus are offered infinitely often, then $\mathscr P$ includes the following regions, whose probabilities turn out to be sufficient for identifiability:
\blue{
\begin{align*}
&\left\{\bm v\in \R^{I}, v_i\le p_i\right\}, \;
\left\{\bm v\in \R^{I}, v_i\le p_i'\right\}, \;
\forall i=1,\dots,I\\
&\left\{\bm v\in \R^{I}, v_{i_1}\le p_{i_1}, v_{i_2}\le p_{i_2}\right\}, \;
\left\{\bm v\in \R^{I}, v_{i_1}\le p'_{i_1}, v_{i_2}\le p_{i_2}\right\},\;
\left\{\bm v\in \R^{I}, v_{i_1}\le p_{i_1}, v_{i_2}\le p'_{i_2}\right\}, \;\forall i_1,i_2=1,\dots,I
\end{align*}
}
\begin{sfnadded}
\Copy{rev:2price}{
Different from \citet{ma2016constructing}, we allow for correlated valuations and arbitrary price menus.
As illustrated by Example~\ref{exm:1price-not-enough}, one price per product is generally insufficient for identification in this more general setting, which motivates the two-price requirement in Proposition~\ref{prop:identifiability}.
}
\end{sfnadded}

\begin{sfnadded}
    We provide a brief outline of the proof for Proposition~\ref{prop:identifiability}.
    For simplicity, consider a firm offering two products, indexed by $1$ and $2$.
    Customer valuations are drawn from a multivariate Gaussian distribution with mean $\bmu = (\mu_1,\mu_2)^\top$
    and covariance matrix
    $\bsig=
    \begin{bmatrix}
    \Sigma_{11} & \Sigma_{12}\\
    \Sigma_{12} & \Sigma_{22}
    \end{bmatrix}$.

    The necessity is immediate. If product $i$ is observed at only one price level $p_i$, then the observable probability
    $\mbp(V_i\le p_i)=\Phi((p_i-\mu_i)/{\sqrt{\Sigma_{ii}}})$
    provides only one equation in the two unknowns $(\mu_i,\Sigma_{ii})$. Hence $(\mu_i,\Sigma_{ii})$ cannot be uniquely determined, and the model is not identifiable.

    To prove sufficiency, we first identify the parameters of each product. Under separate selling, each product $i\in\{1,2\}$ is observed at two distinct price levels $p_i$ and $p_i'$. Therefore, from the two observable probabilities
    $\mbp(V_i\le p_i)$ and $\mbp(V_i\le p_i')$,
    we obtain two equations involving $(\mu_i,\Sigma_{ii})$. Since $\Phi(\cdot)$ is strictly increasing, these two equations uniquely determine $\mu_i$ and $\Sigma_{ii}$. Hence the marginal distributions of both products are identified.
    Once $\mu_1,\mu_2,\Sigma_{11},\Sigma_{22}$ are known, the joint probability
    under the regular menu can be written as
    \begin{equation*}
    \mbp(V_1\le p_1,\ V_2\le p_2)
    =
    \Phi_2\left(
    \frac{p_1-\mu_1}{\sqrt{\Sigma_{11}}},
    \frac{p_2-\mu_2}{\sqrt{\Sigma_{22}}};
    \frac{\Sigma_{12}}{\sqrt{\Sigma_{11}\Sigma_{22}}}
    \right),
    \end{equation*}
    where $\Phi_2(\cdot,\cdot;\rho)$ denotes the bivariate standard normal CDF with correlation $\rho$. For fixed thresholds, $\Phi_2(c_1,c_2;\rho)$ is strictly increasing in $\rho$. It follows that the observed joint probability uniquely determines the correlation parameter, and hence uniquely determines $\Sigma_{12}$.

    Therefore, all entries of $\bmu$ and $\bsig$ are uniquely determined. This establishes identifiability. The detailed proof for Proposition~\ref{prop:identifiability} can be found in the online appendix~\ref{appendix:identifiability}.
\end{sfnadded}

\subsection{Convergence of the EM Algorithm}\label{sec:EM-conv}
Classical EM theory establishes convergence to stationary points of the likelihood function under suitable regularity conditions \citep{dempster1977maximum,wu1983convergence}.
Convergence to the statistically meaningful fixed point corresponding to the true parameters requires additional conditions, especially because the likelihood can have multiple stationary points.
Recent advances such as \citet{balakrishnan2017statistical} provide a framework for proving local contractivity of the population EM operator around the true parameter.
In this section, we use the approach in \cite{balakrishnan2017statistical} to analyze our problem.
To simplify the analysis, we focus on the population-level $Q$-function, following the first step in \cite{balakrishnan2017statistical}; equivalently, we study the idealized limit in which the sample size is infinite.
We also assume that all consumers face the same price menu, so the IC polyhedra are common across observations.
{As a result, we can focus on a partition of $\R^I$, denoted $\{\mathcal P_\rho\}_{\rho=1}^{P}$.}
For tractability, we treat the true covariance matrix $\bsig^*$ as known and analyze the EM update for the mean parameter $\bmu$.
The $Q$-function in this simplified setting can be expressed as
\begin{align}\label{eq:q-func-theory}
  Q(\bmu'\mid \bmu) =  \sum_{\rho=1}^{P} \int_{\bm \xi \in \mathcal P_\rho}f(\bm \xi|\bmu^*) d\bm \xi \int_{\bm v\in \mathcal P_\rho} \frac{f(\bm v \mid \bmu)}{\left(\int_{\bm \xi\in \mathcal P_\rho}f(\bm \xi|\bmu) d\bm \xi\right)} \log f(\bm v|\bm \mu')  \dv.
\end{align}
Compared to \eqref{eq:Q-func}, we remove the dependence on $\bsig$ and use the same partition.
Moreover, because the function is defined at the population level, the probability
{$\int_{\bm \xi \in \mathcal P_\rho}f(\bm \xi|\bmu^*)\mathrm d\bm\xi$ }
in each region replaces the samples in \eqref{eq:Q-func}.
The population-level EM algorithm updates $\bmu$ iteratively according to
the operator:
\begin{align*}
  M(\bmu) \triangleq \argmax_{\bmu'} Q(\bmu' \mid \bmu).
\end{align*}
More precisely, given initialization $\bmu^{(0)}$, the EM algorithm uses $\bmu^{(t+1)}=M\left(\bmu^{(t)}\right)$ to obtain a sequence $\left\{\bmu^{(t)}\right\}_{t=0}^{\infty}$.
The theoretical question is whether we have $\bmu^{(t)}\to\bmu^*$.
Note that the true value $\bmu^*$ always maximizes the population-level likelihood function \citep{van2000asymptotic}.
Moreover, it satisfies the \emph{self-consistency} condition \citep{mclachlan2007algorithm}:
$\bmu^* = \argmax_{\bmu'} Q(\bmu' \mid \bmu^*)$.
Therefore, $\bmu^*$ is a fixed point of the EM operator $M$.
\citet{balakrishnan2017statistical} establish conditions under which $M$ is a contraction mapping in a neighborhood of $\bmu^*$, denoted by $\mathbb{B}(r;\bmu^*)$ with Euclidean radius $r$.
Since iterates of a contraction mapping converge to its fixed point, our goal is to verify analogous conditions for the bundle estimation problem.

\blue{Following \cite{balakrishnan2017statistical}, convergence is guaranteed if $M(\cdot)$ is a contraction mapping in a neighborhood of $\bmu^*$.
This is ensured by two conditions: (1) Concavity of $Q(\bmu\mid\bmu^*)$.
There exists $\lambda > 0$ such that for all $\bmu_1, \bmu_2$ in a neighborhood $\mathbb{B}(r;\bmu^*)$, we have $q(\bmu_1) - q(\bmu_2) - \langle \nabla q(\bmu_2), \bmu_1 - \bmu_2 \rangle  \le -\tfrac{\lambda}{2} \|\bmu_1 - \bmu_2\|_2^2$, where $q(\bmu) \triangleq Q(\bmu \mid \bmu^*)= \sum_{\rho=1}^{P} \int_{\bm v\in \mathcal P_\rho} f(\bm v \mid \bmu^*)\log f(\bm v|\bm \mu)  \dv$. (2) First-order stability. There exists $\gamma < \lambda$ such that $\| \nabla Q(M(\bmu) \mid \bmu^*) - \nabla Q(M(\bmu) \mid \bmu) \|_2  \le \gamma \| \bmu - \bmu^* \|_2$.
}

Note that neither condition can be easily checked in our model.
The main contribution of this section is to identify reasonable assumptions for the conditions to hold.
\blue{To verify these conditions, we introduce a transformed random vector $\bm V' = (\bsig^*)^{-1/2}(\bm V - \bmu^*)$, where $\bm V\sim\Nscr(\bmu^*,\bsig^*)$. Thus, $\bm V' \sim \mathcal N(\bm 0, \bm I)$. Correspondingly, define the whitened regions ${\mathcal P_\rho}' = (\bsig^*)^{-1/2}(\mathcal P_\rho - \bmu^*)$,
    which form a partition of $\mathbb R^I$ in the $\bm V'$-space. Let $\mathcal P'$ be the categorical random region label taking value ${\mathcal P_\rho}'$ on the event $\{\bm V' \in {\mathcal P_\rho}'\}$, so that
    $ \mbp(\mathcal P' = {\mathcal P_\rho}') = \mbp(\bm V' \in {\mathcal P_\rho}')$ for all $\rho = 1, \dots, P.    $
    The conditional mean $\mbe[\bm V' \mid \mathcal P']$ is then a discrete $I$-dimensional random vector taking value $\mbe[\bm V' \mid \bm V' \in {\mathcal P_\rho}']$ on the event $\{\mathcal P' = {\mathcal P_\rho}'\}$, supported on $P$ points.
    The assumption below provides a sufficient condition for the EM algorithm to converge.
    \begin{assumption} \label{minum-eig-assumption}
    There exists $\epsilon > 0$ such that $\lambda_{\min}\!\big(\Var(\mbe[\bm V' \mid \mathcal P'])\big) \geq \ep$,
    where $\lambda_{\min}(\cdot)$ denotes the minimum eigenvalue of a positive semi-definite matrix.
\end{assumption}
Assumption~\ref{minum-eig-assumption} states that the conditional means $\{\mbe[\bm V'\mid \bm V'\in {\mathcal P_\rho}']\}_{\rho=1}^P$ are nondegenerate after whitening: their variation has positive variance in every direction of $\R^I$.}

To provide some intuition for the assumption, consider the case $J=0$, i.e., no bundle is offered and the only feasible action is no-purchase. In this case, $\mbe[\bm V' \mid \mathcal P']$ takes a single value and $\Var(\mbe[\bm V' \mid \mathcal P'])$ is a degenerate $I \times I$ zero matrix, so Assumption~\ref{minum-eig-assumption} is clearly not satisfied. It is not surprising that the EM algorithm cannot recover the true parameters in this degenerate case, because the model is not identifiable, i.e., no information about $\bmu^*$ can be learned from observing only no-purchase decisions.

Now consider the other extreme. Suppose the price menu induces a partition $\{{\mathcal P_\rho}'\}_{\rho=1}^P$ that is so fine that each ${\mathcal P_\rho}'$ almost degenerates to a single point. In this case, $\mbe[\bm V' \mid \bm V' \in {\mathcal P_\rho}']$ effectively traces out the entire support of $\bm V'$, so $\Var(\mbe[\bm V' \mid \mathcal P']) \approx \bm I$, satisfying Assumption~\ref{minum-eig-assumption}. The two examples illustrate the intuition that the EM algorithm performs well when the price menu induces many small regions, which is consistent with the discussion after Example~\ref{exm:1price-not-enough}.
\blue{Assumption~\ref{minum-eig-assumption} guarantees the convergence of the EM algorithm to the true parameter locally.
\begin{thm} \label{thm:contra-map}
    Suppose Assumption~\ref{minum-eig-assumption} holds.
    There exists a neighborhood $\mathbb B(r;\bmu^*)$ of $\bmu^*$ such that for $\bmu^{(0)}\in \mathbb B(r;\bmu^*)$
    the EM algorithm guarantees
    $\|\bmu^{(t)} - \bmu^* \|_2 \leq \left( 1-\ep/2 \right)^t \| \bmu^{(0)} - \bmu^* \|_2$.
\end{thm}}

The detailed proof for Theorem~\ref{thm:contra-map} can be found in the online appendix ~\ref{appendix:EM-conv}.

\begin{lightgray}
{
}
\end{lightgray}

%% file: 3_Extensions.tex
\section{Model Extensions}\label{sec:extension}
In this section, we consider two extensions of the base model in Section~\ref{sec:base-model}.

\subsection{\blue{Product Synergy in a Bundle}}\label{sec:synergy}
In Section~\ref{sec:base-model}, the bundle utility is modeled as the sum of the utilities of the included products, with the covariance structure of product valuations capturing implicit relationships across products. 
In practice, however, products within a bundle may exhibit explicit complementarity or substitutability that renders the overall utility of the bundle non-additive. 
These \emph{synergy effects} imply that a bundle’s utility is not necessarily equal to the sum of its standalone components; instead, the combination of specific products may amplify or diminish the perceived value.
For example, pairing a printer with compatible ink cartridges often yields additional value relative to purchasing the items separately (positive synergy), whereas bundling two devices with overlapping functionality may reduce incremental benefit (negative synergy). 
Prior studies typically incorporate such effects through pairwise or higher-order interaction terms \citep{Benson_Kumar_Tomkins_2018,ouyang2023semiparametricdiscretechoicemodels,tulabandhula2023multipurchasebehaviormodelingestimation}. In line with this literature, we adopt a pairwise interaction structure, which offers a flexible yet tractable representation of inter-product synergies.
We next investigate the estimation problem when bundle choices may exhibit such synergy effects.

\textbf{Vectorized representation of bundles.} 
To express the utility model compactly, we first introduce an equivalent vectorized representation of bundles. 
Let $\bm x_j\in \left\{0,1\right\}^I$ denote bundle $j$, where the $i$-th entry equals 1 if product $i\in[I]$ is included in the bundle. 
Customer $n$ is faced with a menu of bundles $\bm X_n\in\{0,1\}^{I\times J_n}$, where each column of $\bm X_n$ is a binary vector denoting an offered bundle and $J_n$ is the number of bundles shown to customer $n$.  
The corresponding bundle prices are collected in $\bm p_n=(p_n^1,\dots,p_n^{J_n})$, which is observed in the data.
Customer $n$ chooses $c_n\in[J_n]\cup\{0\}$, where $c_n = 0$ denotes the no-purchase option.
The dataset therefore consists of tuples $\{(\bm X_n, \bm p_n,c_n)\}_{n=1}^N$.
Note that customers do not necessarily observe the same menu, nor do they see the complete set of possible bundles. 
This vectorized representation generalizes the formulation in Section~\ref{sec:base-model} and provides a convenient way to incorporate synergy effects.

Given a vector of individual product valuation $\bm v$, the utility of bundle $\bm x_j$ to customer $n$ is specified as
\begin{equation}\label{eq:bundle-utility-synergy}
    u_{nj}\triangleq \bm x_j^\top \bm v-p_n^j+\bm x_j^\top \bA \bm x_j,
\end{equation}
where the \emph{upper triangular matrix} $\bA\in\R^{I\times I}$ captures the pairwise synergy between products that appear in the same bundle. 
In particular, if products $i_1$ and $i_2$ are both present in bundle $j$, then the bundle utility includes an additional term $A_{i_1 i_2}$. The sign of $A_{i_1 i_2}$ reflects whether the two products are complementary ($A_{i_1 i_2}>0$) or substitutable ($A_{i_1 i_2}<0$). We impose $A_{ii}=0$ for all $i$ so that $\bA$ only encodes interactions between distinct products.
When $\bA\equiv\bm 0$, the model reduces to the baseline additive specification in Section~\ref{sec:base-model}.

\Copy{rev:compare_Sig_and_A}{
\begin{rem}
The covariance matrix $\bsig$ of $\bm v$ captures correlation in customers’ valuations across products,
whereas the synergy matrix $\bA$ captures interaction effects within a bundle that directly enter the bundle utility. 
For example, a positive covariance $\sigma_{12}$ indicates that customers who value product $1$ highly also tend to value product $2$ highly.
A positive synergy parameter $A_{12}$, on the other hand, implies that including products $1$ and $2$ in the \emph{same} bundle generates additional utility beyond the additive sum of their valuations.
\end{rem}
}

\textbf{IC polyhedra with synergy.} 
Under the synergy specification \eqref{eq:bundle-utility-synergy}, the IC polyhedra in \eqref{eq:ic-polytope} can be rewritten in a vectorized form, comparing the utility of each bundle to that of the chosen one and incorporating the synergy terms that arise whenever two products appear together in a bundle.
In particular, the bundle-utility vector for customer $n$ is given by
\begin{equation}\label{eq:bundle-utility-synergy-vec}
    \bm u_n(\bm v) = \bXn^\top \bm v-\bm p_n+ \diag(\bm X_n ^\top \bA\bm X_n)\in\R^{J_n},
\end{equation}
where $\diag(\cdot)$ extracts the diagonal of a square matrix as a column vector. Then for customer $n$, we can define the corresponding IC region in the bundle utility space by
\begin{equation}
    U_n^{c_n} = \left\{
    \begin{array}{ll}
    \{ \bm u\in \R^{J_n}
    | \; 
    \bm u\le \bm 1_{J_n}u_{nc_n},\; u_{nc_n}\ge 0
    \}, & c_n\neq0;\\
    \{ \bm u\in \R^{J_n}
    | \;
    \bm u\le \bm 0_{J_n}\}, & c_n=0.
    \end{array}
    \right.
\end{equation}
where $\bm 1_n$ and $\bm 0_n$ denote $n$-dimensional column vectors of ones and zeros, respectively. 
A valuation vector $\bm v$ satisfies the IC constraints if and only if its induced
utility vector $\bm u_n(\bm v)$ lies in this set. Hence, the corresponding IC polyhedra in the product valuation space is
\begin{equation}\label{eq:ic-polytope-synergy}
    K_n^{c_n}\triangleq\left\{\bm v \in \mathbb R^{I} 
    \left| \; 
    \bm u_n(\bm v)\in U_n^{c_n}\right.
    \right\}.
\end{equation}
\Copy{rev:difficulty_estimate_A}{
The set $K_n^j$ collects all the vectors $\bm v$ for which the bundle $j$ (including $j=0$ for the outside option) is incentive compatible for customer $n$.
Equivalently, it is the set of all $\bm v$ whose induced bundle-utility vector $\bm u_n(\bm v)$ satisfies the inequalities defining $U_n^j$.
A key difference from Section~\ref{sec:base-model} is that the IC polyhedron $K_n^{j}$ in \eqref{eq:ic-polytope-synergy} now  explicitly depends on the synergy matrix $\bA$ through the term $\diag(\bm X_n^\top \bA\bm X_n)$.  
As a result, the boundary of $K_n^j(\bA)$ changes whenever $\bA$ changes. 
This introduces a \emph{fundamental challenge}: in the estimation problem of Section~\ref{sec:base-model}, the IC polyhedra are directly obtained from the data and do not change with the parameters $(\bmu,\bsig)$. 
With the synergy effect, the likelihood function or the $Q$ function \eqref{eq:Q-func} integrates over a region that depends on $\bA$.
This dependence is also non-differentiable,
which becomes computationally prohibitive in the M-step: how can we maximize over $\bA$ when the objective $Q$ function depends on $\bA$ in such an intractable and discrete form.
}

We next articulate how to adapt the computational framework introduced in Section~\ref{sec:base-model} to this setting.
\paragraph{E-step.} 
Following the standard EM framework, we evaluate the expectation of the complete-data log-likelihood with respect to the conditional distribution of the latent variables $\bm v$. 
The likelihood can be written as 
$\Lscr(\bm\theta;\bm D) 
= \prod_{n=1}^N\int_{K_n^{c_n}(\mathbf{A})} f(\bm v|\bm\mu,\bm\Sigma) \dv 
=\prod_{n=1}^N\int  \mI \{\bm v \in K_n^{c_n}(\mathbf{A})\} f(\bm v|\bm\mu,\bm\Sigma) \dv$,
where $f(\cdot)$ denotes the PDF of the multivariate normal distribution.
The complete-data log-likelihood can therefore be written as
\begin{equation}\label{eq:llh_complete_synergy}
\begin{aligned}
    \ell(\bm\theta;\bm D,\bm Z)
    =\sum_{n=1}^N \log f(\bm v_n\mid \bmu,\bsig)
    +\sum_{n=1}^N \log \mI\{\bm v_n\in K_n^{c_n}(\bA)\}.
\end{aligned}
\end{equation}
In the E-step, the conditional law of $\bm v_n$ is no longer a truncated Gaussian over a fixed region. Thus, a direct M-step based on \eqref{eq:llh_complete_synergy} is difficult because the term $\mI\{\bm v_n\in K_n^{c_n}(\bA)\}$ is non-differentiable in $\bA$, and the feasible region itself changes with $\bA$.
We address the non-differentiability by a differentiable surrogate that approximates this objective.
\Copy{rev:transformation_failure}
{
}
In particular, we approximate the hard IC condition by a product of sigmoid terms:
    \begin{equation}\label{eq:sigmoid_approx}
    \mI\{\bm v\in K_n^{c_n}(\bA)\}
    \approx
    \prod_{j=1}^{J_n}\sigma\!\left(\frac{\Delta U_{nj}(\bm v;\bA)}{\lambda}\right),
    \end{equation}
    where $\sigma(\cdot) = (1 + e^{-\cdot})^{-1}$ is the sigmoid function and $\lambda>0$ is a smoothing parameter,
    and
    $\Delta U_{nj}  
    = \mI\{c_n\neq 0\}\,u_{n c_n} 
    -
    \mI\{c_n=0\ \text{or}\ j\neq c_n\}\,u_{nj}$.
    Essentially, we turn the IC indicator $u_{nc_n}\ge u_{nj}$ to a soft margin $\sigma((u_{nc_n}-u_{nj})/\lambda)$, which
    measures how much the observed choice $c_n$ is preferred to alternative $j$ under $(\bm v,\bA)$.
    This approximation gives a differentiable surrogate likelihood, so $\bA$ can be updated in the M-step using gradient-based optimization.
    As $\lambda\to 0$, one can show that the surrogate approaches the original indicator.

\begin{sfnadded}
\Copy{rev:sigmoid}{
}
\end{sfnadded}

\begin{sfnadded}
    Consequently, under the proposed smoothing approximation with synergy matrix $\bA$, the indicator function in E-step is replaced by the sigmoid approximation in \eqref{eq:sigmoid_approx}. 
    As a result, the hard truncation of the feasible region is replaced by a continuous weighting term. 
    The smoothed likelihood is therefore approximated by $\Lscr(\bm\theta;\bm D)\approx \prod_{n=1}^N\int \prod_{j=1}^{J_n} \sigma\left(\frac{\Delta U_{nj}(\bm v; \mathbf{A})}{\lambda}\right) f(\bm v|\bm\mu,\bm\Sigma) \mathrm{d}\bm v. $
The complete-data log-likelihood can therefore be written as
\begin{equation}\label{eq:likelihood_sigmoid_complete-data}
\begin{aligned}
    \ell(\bm\theta;\bm D,\bm Z)=\sum_{n=1}^N\left(\log f(\bm v_n | \bmu, \bsig) +\sum_{j=1}^{J_n} \log \sigma\left(\frac{\Delta U_{nj}(\bm v_n; \bA)}{\lambda}\right)\right).
\end{aligned}
\end{equation}
Substituting this conditional density into the expectation yields the smoothed $Q'$ function
\begin{equation}\label{eq:smoothed_Q}
\begin{aligned}
Q'(\bm\theta \mid \bm\theta^{(t)})
&=\sum_{n=1}^{N}\int q_n(\bm v;\bm\theta^{(t)})
\left(
\log f(\bm v \mid \bmu,\bsig)+\sum_{j=1}^{J_n}\log \sigma\left(
\frac{\Delta U_{nj}(\bm v;\bA)}{\lambda}
\right)
\right)
\mathrm{d}\bm v\\
\end{aligned},
\end{equation}
where $q_n(\bm v; \bm\theta^{(t)})=  
\left( f(\bm v \mid \bmu^{(t)}, \bsig^{(t)}) \prod_{j=1}^{J_n}\sigma(\frac{\Delta U_{nj}(\bm v; \bA^{(t)})}{\lambda}) 
\right)
\big/ 
\left(
{\int f(\bm\xi \mid \bmu^{(t)}, \bsig^{(t)})\prod_{j=1}^{J_n}\sigma(\frac{\Delta U_{nj}(\bm\xi; \bA^{(t)})}{\lambda})  \mathrm{d}\bm\xi}
\right).$

Compared to the EM formulation in Section \ref{sec: The_EM_Algorithm}, the resulting $Q'$ function differs in two important aspects. First, the truncation of the utility distribution is replaced by a smooth sigmoid term. Second, the log-likelihood now contains an additional contribution from the smoothed IC constraints, which introduces explicit dependence on the synergy parameter $\bA$ inside the integrand. This modification enables gradient-based optimization with respect to $\bA$ while preserving the overall EM structure.

Similar to the base model, substituting the Monte Carlo approximation into Eq.~(\ref{eq:smoothed_Q}) yields the $\hat {Q}'$ function
\begin{equation}\label{eq:smoothed_Qhat}
\begin{aligned}
\hat {Q}'(\bm\theta \mid \bm\theta^{(t)})&=\sum_{n=1}^N\sum_{l=1}^L
\overline\omega_{nl}\left(
\log f(\bm v_n^{(l)}|\bmu,\bsig)+\log \omega(\bm v_n^{(l)};\bA)
\right)\\
&=C-\frac{N}{2}\log|\bsig|+\sum_{n=1}^N\sum_{l=1}^L
\overline\omega_{nl}\left(
-\frac{1}{2}(\bm v_n^{(l)}-\bmu)^\top\bsig^{-1}(\bm v_n^{(l)}-\bmu)+\log \omega(\bm v_n^{(l)};\bA)
\right),
\end{aligned}
\end{equation}
where $C$ is a constant with respect to $\bm \theta$ and
\begin{equation}\label{eq:sigmoid_vals}
    \omega(\bm v;\bA) \triangleq\prod_{j=1}^{J_n}\sigma\left(
    \frac{ \Delta U_{nj}(\bm v; \bA) }{\lambda}
    \right),\quad
    \overline\omega_{nl} \triangleq \frac{ \omega(\bm v_{n}^{(l)};\bA^{(t)})} {\sum_{l=1}^L \omega(\bm v_{n}^{(l)};\bA^{(t)})}.
\end{equation}

\paragraph{M-step.} 
In the M-step, we maximize 
$\hat {Q'}(\bm\theta\mid\bm\theta^{(t)})$ 
with respect to $(\bmu,\bA,\bsig)$. The objective function is concave in each parameter block.
First, the Gaussian log-likelihood term is concave in $\bmu$ and in $\bsig^{-1}$, which follows from standard properties of the multivariate normal distribution.
Second, Lemma~\ref{lem:concavity_wrt_A} below establishes that the contribution involving $\bA$ is also concave. The detailed proof is deferred to the Appendix \ref{app:prof_concavity_A}.
\begin{lem}[Concavity with respect to $\bA$]\label{lem:concavity_wrt_A}
For fixed $(\bmu,\bsig)$, the smoothed objective function $\hat {Q'}(\bm\theta\mid\bm\theta^{(t)})$ is concave with respect to the synergy matrix $\bA$.
\end{lem}

Unlike $\bmu$ and $\bsig$, which admit closed-form updates from the first-order conditions, the synergy matrix $\bA$ does not admit a closed-form solution in the M-step and must be estimated using gradient-based optimization, where the gradient is given in Theorem~\ref{thm:grad_A}.

\begin{thm}[Gradient of the smoothed objective with respect to $\bA$]\label{thm:grad_A}
The gradient of the smoothed objective $\hat{Q'}$ with respect to the synergy matrix $\bA$ is given by
\begin{equation}
\nabla_{\bA}\hat{Q'}
= \frac{1}{\lambda}\sum_{n=1}^N \left(\sum_{j=1}^{J_n} \underbrace{\frac{\partial \Delta U_{nj}(\bA)}{\partial \bA}}_{\text{constant in }\bm v_n^{(l)}} \sum_{l=1}^L \overline\omega_{nl}\!\left(1-\sigma\!\left(\frac{\Delta U_{nj}(\bm v_n^{(l)};\bA)}{\lambda}\right)\right)\right),
\end{equation}
where
\begin{equation}
\frac{\partial \Delta U_{nj}(\bm v;\bA)}{\partial \bA}
= \mI\{c_n\neq 0\}\bm x_{nc_n}\bm x_{nc_n}^\top
- \mI\{c_n=0\ \text{or}\ j\neq c_n\}\bm x_{nj}\bm x_{nj}^\top.
\end{equation}
\end{thm}

The proof of Theorem~\ref{thm:grad_A}, the closed-form updates for $\bmu$ and $\bsig$, and the corresponding update formulas under the importance sampling framework are provided in Appendix~\ref{app:synergy}.

\end{sfnadded}

\begin{sfnadded}
{
}
\end{sfnadded}

\begin{sfnadded}
{
}
\end{sfnadded}

{
}

\subsection{Censored Demand}\label{sec:censored-demand}
In practice, transaction data are often censored because the firm records purchases but does not observe customers who visit without buying.
We next extend our framework to account for this demand censoring.

For ease of exposition, suppose the firm offers a single price menu $\bm p = (p^{1},\dots,p^J)$ to all customers.
The observed data consist of $(p^{1},\dots,p^J,c_n)$ for each purchasing customer $n$, where $c_n>0$ for $n=1,\dots,N$.
These data can be converted to the IC polyhedra $\{R^0,R^1,\dots,R^J\}$ defined in \eqref{eq:ic-polytope}.
Because the price menu is fixed, these regions are common across customers.
As in the base model, we treat each customer valuation $\bm v_n$, $n=1,\dots,N$, as missing data.

Because of the demand censoring, the total number of customers that have visited the store, $N'\ge N$, is also unobserved, as well as the valuations of the censored customers $\bm v_n$, $n=N+1,\dots,N'$.
Therefore, the missing data is $\bm Z\triangleq \left\{N', \bm v_{1},\dots,\bm v_{N'}\right\}$.
It is known that $\bm v_n\in R^0$ for the censored customers $n=N+1,\dots,N'$.
Moreover, we record the total number of customers that buy bundle $j$ as $N_j$, where $\sum_{j=1}^J N_j=N$.
The observed data can thus be encoded as $\bm D \triangleq \left\{N_1,\dots,N_J\right\}$.
Below we describe the implementation of the EM algorithm to estimate $\bm \theta= (\bm \mu, \bm \Sigma)$.


\begin{sfnadded}
    Relative to the base model in Section~\ref{sec:base-model}, the E-step now includes an additional outer expectation over the unobserved total number of customers, $N'\mid \bm D,\bm\theta^{(t)}$.
    Under $\bm\theta^{(t)}$, a customer is censored with probability $\mbp(R^0\mid \bmu^{(t)},\bsig^{(t)})$, the probability of the no-purchase region. Thus, conditional on $N'$, observing $N$ purchases corresponds to $N$ successes in $N'$ Bernoulli trials with success probability $1-\mbp(R^0\mid \bmu^{(t)},\bsig^{(t)})$.
    
    Once the distribution of $N'$ is identified, the E-step is approximated by Monte Carlo simulation as in Section~\ref{sec: The_EM_Algorithm}: in each instance, we first draw $N'^{(l)}$ from a negative binomial distribution, then draw the observed customer valuations from truncated Gaussians on the corresponding IC polyhedra $R^j$, and the censored customer valuations from a truncated Gaussian on $R^0$.
    The M-step can then be adapted using a $\hat Q$ function computed from the simulated samples above, and the parameters are updated by maximizing this $\hat Q$.
\end{sfnadded}

{
}
Then Algorithm~\ref{alg:em-basic} can be adapted to the censored-demand data.
  To conclude this section, we discuss how the EM algorithm can be generalized when customers may observe different price menus provided by the firm, for example, based on the arrival time in a day.
Suppose that the firm offers $m$ price menus to customers.
The transaction data consists of $N^{m}$ observed customer purchases from each price menu and their choices.
Note that the IC polyhedra differ across the price menus.
Using the same procedure stated above, we can use Monte Carlo to simulate the number of total customers $N'^m$ \emph{for each price menu} and simulate the valuations of the customers accordingly.
They can be used in the M-step to update the parameters.
\begin{sfnadded}
The detailed derivation of the E-step, the negative binomial form for $N'\mid \bm D, \bm\theta^{(t)}$, the Monte Carlo procedure, and the resulting M-step update formulas are provided in Appendix~\ref{app:censored-demand}.
\end{sfnadded}

{
}

{
}

%% file: 4_Numerical_Base.tex
\section{Numerical Experiments}\label{sec:numerical}
\begin{sfnadded}
\Copy{rev:numerical_overview}{
    In this section, we conduct a series of numerical experiments to evaluate the performance of the proposed EM algorithm.
\blue
{
We begin by studying convergence and estimation accuracy under the base model in Section~\ref{subsec:base_model}.
We then compare our approach with a Bayesian benchmark and a variant of the MNL model.
Next, we extend the analysis to more general settings in Sections~\ref{subsec:numerical_censored},
\ref{subsec:numerical_gmm}, and
\ref{subsec:numerical_synergy}, including censored observations, Gaussian mixture models, and models with bundle-specific synergy effects. 
Finally, we evaluate the proposed framework using real-world data from JD.com \citep{shen2020jd}.
}
}

\Copy{rev:numerical_setup}{
The experimental setup for the base model is described below.
We consider different combinations of the number of products $I$ and the number of customers $N$.
For each $(I, N)$, we generate multiple datasets under different parameter settings.
The offered set is randomly drawn from all the available bundle collections.
We generate 10 price menus. In each menu, individual product prices follow $p_i\sim\mathcal U[\mu_i-3, \mu_i+3]$ for the corresponding mean valuation $\mu_i$.
For each customer, we randomly select one of the price menus. 
For bundle prices, we calculate the sum of individual prices within the bundle and generate a discount from a uniform distribution $d_j\sim \mathcal U[0,2]$. 
For example, consider a bundle including products $\{1,2,3\}$, with respective prices 5, 6, 7. 
If the random discount is 2, then the bundle price is $5+6+7-2 = 16$. 
Each consumer's valuation vector is drawn from a \emph{multivariate Gaussian distribution}, with the mean drawn from a uniform distribution
$\bm \mu \sim \mathcal{U}[6,12]^I$, 
and the covariance $\bm \Sigma = \Lambda\Lambda^\top$, where the component of $\Lambda$ follows uniform distribution $\mathcal{U}[-2,2]$.}
\end{sfnadded}
To assess estimation accuracy, we use the $\ell_1$-error
$(\|\bmu^{(t+1)}-\bmu^*\|_1+\|\bsig^{(t+1)}-\bsig^*\|_1)/(I^2 + I)$,
which measures the average absolute deviation of the estimated parameters from their true values.

\begin{lightgray}
{
}  
\end{lightgray}

{
}

\subsection{EM Algorithm in the Base Model} \label{subsec:base_model}
{
We first evaluate the convergence of Algorithm~\ref{alg:em-imp} by tracking the log-likelihood and estimation errors across iterations. We also investigate how the estimation accuracy scales with the number of products and customers.

}

\paragraph{EM convergence.} We investigate the convergence of Algorithm~\ref{alg:em-imp} against the number of iterations.  
In this experiment, we consider two products ($I=2$) and one bundle. We randomly generate a sample of $N=1000$ transactions and initialize the algorithm from three different points with varying distances from the true parameters: within one standard deviation (close), two standard deviations (midway), and five standard deviations of the true mean (distant).
In each EM iteration, we calculate the log-likelihood using the estimated parameters and define the error as the difference between two subsequent values. 
We terminate the algorithm when either the error falls below $0.005$ or the number of EM iterations reaches $500$.

The left panel of Figure~\ref{fig:byinitials}
shows that the average log-likelihood of Algorithm~\ref{alg:em-imp} converges toward that under the true parameter across the three initializations.
The right panel of Figure~\ref{fig:byinitials} shows the trajectory of the $\ell_1$-error. When the initial point is far from the true parameters, the $\ell_1$-error may initially increase before decreasing.
\begin{figure}[ht]
    \begin{center}
    {%
        \begin{tikzpicture}
        \begin{axis}[
            width=0.46\linewidth, height=5.5cm,
            xlabel=Iterations,
            ylabel=Average log-likelihood,
            xmin=0, xmax=20,
            ymin=-2.6, ymax=-0.45,
            xtick={0,5,10,15,20},
            legend pos=south east,
            legend style={font=\small},
            grid=major, grid style={dashed,gray!30},
        ]
        \addplot[mark=none, solid, line width=1.5pt, color=red]
            table[x=t, y=distant, col sep=comma]{avg_llh_byinitials_I2N1000_base.csv};
        \addplot[mark=none, dashed, line width=1.5pt]
            table[x=t, y=midway, col sep=comma]{avg_llh_byinitials_I2N1000_base.csv};
        \addplot[mark=none, densely dotted, line width=1.5pt, color=green!70!black]
            table[x=t, y=close, col sep=comma]{avg_llh_byinitials_I2N1000_base.csv};
        \addplot[domain=0:20, color=black, dash dot, line width=1.2pt]{-0.5694};
        \legend{Distant, Midway, Close, Exact}
        \end{axis}
        \end{tikzpicture}%
    }%
    {%
        \begin{tikzpicture}
        \begin{axis}[
            width=0.46\linewidth, height=5.5cm,
            xlabel=Iterations,
            ylabel=$\ell_1$-error,
            xmin=0, xmax=20,
            ymin=0, ymax=6.5,
            legend pos=north east,
            legend style={font=\small},
            grid=major, grid style={dashed,gray!30},
        ]
        \addplot[mark=none, solid, line width=1.5pt, color=red]
            table[x=t, y=distant, col sep=comma]{avg_l1error_byinitials_I2N1000_base.csv};
        \addplot[mark=none, dashed, line width=1.5pt]
            table[x=t, y=midway, col sep=comma]{avg_l1error_byinitials_I2N1000_base.csv};
        \addplot[mark=none, densely dotted, line width=1.5pt, color=green!70!black]
            table[x=t, y=close, col sep=comma]{avg_l1error_byinitials_I2N1000_base.csv};
        \legend{Distant, Midway, Close}
        \end{axis}
        \end{tikzpicture}%
    }
    \end{center}
    \caption{Convergence of log-likelihood and $\ell_1$-error during the EM algorithm for different initial estimates. The experiment is conducted for $I=2$ products and $N=1000$ transactions. \label{fig:byinitials}}
\end{figure}


{
}

{
}

{
}

\begin{sfnadded}
\Copy{rev:numerical_summary}{
We further investigate the impact of the number of transactions on the performance of the EM algorithm.
We fix the number of products to $I=2$ and generate datasets with varying numbers of customer transactions
$N \in \{3000, 5000, 10000\}$.
The results in Figure~\ref{error2} show that as the sample size increases, the estimation accuracy improves.
}

  \begin{figure}[ht]
      \begin{center}
      {%
          \begin{tikzpicture}                                                       
          \begin{axis}[name=llhmain,
              width=0.46\linewidth, height=5.5cm,
              xlabel=Iterations,
              ylabel=Average log-likelihood,
             xmin=50, xmax=100,
            ymin=-0.65, ymax=-0.55,
            xtick={50,60,70,80,90,100},
              legend pos=south east,
              legend style={font=\scriptsize},
              grid=major, grid style={dashed,gray!30},
          ]
          \addplot[mark=none, solid,         line width=1.2pt, color=red]
              table[x=t, y=N1000,  col sep=comma]{avg_llh_byN_I2_base_noA3.csv};
          \addplot[mark=none, dashed,        line width=1.2pt, color=blue]
              table[x=t, y=N2000,  col sep=comma]{avg_llh_byN_I2_base_noA3.csv};
          \addplot[mark=none, solid,        line width=1.2pt, color=teal]
              table[x=t, y=N3000,  col sep=comma]{avg_llh_byN_I2_base_noA3.csv};
          \addplot[mark=none, densely dotted,line width=1.2pt, color=green!70!black]
              table[x=t, y=N10000, col sep=comma]{avg_llh_byN_I2_base_noA3.csv};
        \addplot[domain=50:100, color=black, dash dot, line width=1pt]{-0.5627};
          \draw[black!60, dashed, thin] (axis cs:65,-0.58) rectangle (axis cs:75,-0.56);
          \legend{$N{=}1000$, $N{=}2000$, $N{=}3000$, $N{=}10000$, Exact}
          \end{axis}
          \begin{axis}[
              at={(llhmain.south west)}, anchor=south west,
              xshift=9mm, yshift=8mm,
              width=3.4cm, height=2.4cm,
              xmin=65, xmax=75, ymin=-0.58, ymax=-0.56,
              xtick={65,70,75}, ytick={-0.58,-0.57,-0.56},
              tick label style={font=\tiny},
              axis background/.style={fill=white},
              grid=major, grid style={dashed,gray!20},
          ]
          \addplot[mark=none, solid,         line width=1pt, color=red]
              table[x=t, y=N1000,  col sep=comma]{avg_llh_byN_I2_base_noA3.csv};
          \addplot[mark=none, dashed,        line width=1pt, color=blue]
              table[x=t, y=N2000,  col sep=comma]{avg_llh_byN_I2_base_noA3.csv};
        \addplot[mark=none, solid,        line width=1pt, color=teal]
              table[x=t, y=N3000,  col sep=comma]{avg_llh_byN_I2_base_noA3.csv};
          \addplot[mark=none, densely dotted,line width=1pt, color=green!70!black]
              table[x=t, y=N10000, col sep=comma]{avg_llh_byN_I2_base_noA3.csv};
           \addplot[domain=65:75, color=black, dash dot, line width=0.8pt]{-0.5627};
          \end{axis}
          \end{tikzpicture}%
      }%
      {%
          \begin{tikzpicture}
          \begin{axis}[name=l1main,
              width=0.46\linewidth, height=5.5cm,
              xlabel=Iterations,
              ylabel=$\ell_1$-error,
              xmin=0, xmax=50,
              ymin=0.02, ymax=45,
              legend pos=north east,
              legend style={font=\scriptsize},
              grid=major, grid style={dashed,gray!30},
          ]
          \addplot[mark=none, solid,         line width=1.0pt, color=red]
              table[x=t, y=N1000,  col sep=comma]{avg_l1error_byN_I2_base_noA3.csv};
          \addplot[mark=none, dash dot,      line width=1.0pt, color=blue]
              table[x=t, y=N2000,  col sep=comma]{avg_l1error_byN_I2_base_noA3.csv};
          \addplot[mark=none, solid,         line width=1.0pt, color=teal]
              table[x=t, y=N3000,  col sep=comma]{avg_l1error_byN_I2_base_noA3.csv};
          \addplot[mark=none, densely dotted,line width=1.2pt, color=green!70!black]
              table[x=t, y=N10000, col sep=comma]{avg_l1error_byN_I2_base_noA3.csv};
          \draw[black!60, dashed, thin] (axis cs:20,10) rectangle (axis cs:30,20);
          \legend{1000,2000,3000,10000}
          \end{axis}
          \begin{axis}[
              at={(l1main.south west)}, anchor=south west,
              xshift=5mm, yshift=5mm,
              width=3.4cm, height=2.4cm,
              xmin=20, xmax=30, ymin=10, ymax=20,
              xtick={20,25,30}, ytick={10,15,20},
              tick label style={font=\tiny},
              axis background/.style={fill=white},
              grid=major, grid style={dashed,gray!20},
          ]
          \addplot[mark=none, solid,         line width=1pt, color=red]
              table[x=t, y=N1000,  col sep=comma]{avg_l1error_byN_I2_base_noA3.csv};
          \addplot[mark=none, dash dot,      line width=1pt, color=blue]
              table[x=t, y=N2000,  col sep=comma]{avg_l1error_byN_I2_base_noA3.csv};
          \addplot[mark=none, solid,         line width=1pt, color=teal]
              table[x=t, y=N3000,  col sep=comma]{avg_l1error_byN_I2_base_noA3.csv};
          \addplot[mark=none, densely dotted,line width=1pt, color=green!70!black]
              table[x=t, y=N10000, col sep=comma]{avg_l1error_byN_I2_base_noA3.csv};
          \end{axis}
          \end{tikzpicture}%
      }
      \end{center}
      \caption{Convergence of log-likelihood and $\ell_1$-error during the EM algorithm for varying numbers of observations. The experiment is conducted for $I=2$ products. \label{error2}}
  \end{figure}


We provide a more comprehensive comparison for different numbers of products and observations in Table~\ref{tab:l1error_by_IN}. 
For each setting, we initialize the estimator from five standard deviations from the true mean (distant point). 
We report the final $\ell_1$-error after convergence for $I\in\{2,3,4,5,6\}$ and $N\in \{1000, 1500,2000,2500,3000, 5000, 10000\}$. Taken together, these results show that 
Algorithm~\ref{alg:em-imp} exhibits stable and consistent improvement as the sample size increases, demonstrating its scalability in both dimensions.

    


\begin{sfnadded}

We additionally report the computational running time of the algorithm. All experiments were conducted on a Linux server equipped with 64 physical CPU cores (128 logical cores) running Python 3.10.19.
To provide a hardware-independent measure of computational cost, we report the total CPU time, defined as the sum of the CPU seconds consumed by all worker processes. 
Most experimental settings finish within several minutes to tens of minutes, although a small number of instances require longer runtime. Since the computational cost depends on the realized observations and the corresponding convergence behavior of the EM iterations, the reported runtimes exhibit variability across datasets.

\begin{table}
      \centering
      \caption{$\ell_1$-error and running time (minutes) of the EM algorithm.
      The numbers in parentheses report the standard deviations across five datasets per $(I, N)$.}
      \label{tab:l1error_by_IN}
      \fontsize{7.5}{9}\selectfont   
      \setlength{\tabcolsep}{2pt}
      \begin{tabular}{c|rr|rr|rr|rr|rr}
          \toprule
          & \multicolumn{2}{c|}{$I=2$}
          & \multicolumn{2}{c|}{$I=3$}
          & \multicolumn{2}{c|}{$I=4$}
          & \multicolumn{2}{c|}{$I=5$}
          & \multicolumn{2}{c}{$I=6$} \\
          $N$
          & $\ell_1$-error & Time
          & $\ell_1$-error & Time
          & $\ell_1$-error & Time
          & $\ell_1$-error & Time
          & $\ell_1$-error & Time \\
          \midrule
          1000  & 0.4636 (0.22) & 15.3 (3.8)   & 0.2683 (0.21) & 31.3 (8.8)    & 0.4668 (0.28) & 53.8 (3.8)   & 0.6259 (0.23) & 64.8 (2.5)   & 1.5513 (2.05) & 77.8 (6.7)   \\
          1500  & 0.3346 (0.14) & 23.9 (5.0)   & 0.2650 (0.23) & 40.5 (20.6)   & 0.4229 (0.08) & 81.7 (5.2)   & 0.3859 (0.09) & 102.5 (7.5)  & 0.7978 (0.71) & 120.4 (13.3) \\
          2000  & 0.3483 (0.19) & 30.4 (7.0)   & 0.2837 (0.27) & 52.3 (27.3)   & 0.3559 (0.15) & 108.6 (8.0)  & 0.3560 (0.14) & 134.0 (8.5)  & 0.6450 (0.56) & 157.7 (15.6) \\
          2500  & 0.3102 (0.18) & 35.2 (8.3)   & 0.2538 (0.20) & 64.8 (32.2)   & 0.3184 (0.07) & 135.3 (9.7)  & 0.3662 (0.19) & 169.0 (11.0) & 0.5773 (0.50) & 198.2 (18.3) \\
          3000  & 0.3352 (0.12) & 44.2 (11.7)  & 0.2586 (0.27) & 73.5 (33.3)   & 0.2966 (0.10) & 161.3 (10.7) & 0.3389 (0.16) & 206.1 (13.7) & 0.4844 (0.46) & 238.0 (21.2) \\
          5000  & 0.3891 (0.45) & 70.9 (16.3)  & 0.2192 (0.24) & 105.2 (59.0)  & 0.2386 (0.07) & 271.8 (17.7) & 0.2568 (0.12) & 341.5 (21.1) & 0.4129 (0.49) & 395.5 (34.3) \\
          10000 & 0.2389 (0.24) & 139.4 (33.3) & 0.1419 (0.09) & 200.0 (117.0) & 0.1657 (0.08) & 533.8 (37.6) & 0.1555 (0.11) & 666.1 (72.6) & 0.4450 (0.68) & 779.1 (71.0) \\
          \bottomrule
      \end{tabular}
  \end{table}

\end{sfnadded}

\paragraph{Comparison to Bayesian methods.}\label{ref:bayesian-computation}
We compare 
Algorithm~\ref{alg:em-imp} with the method introduced in \cite{jedidi2003measuring}. 
The latter conducts posterior inference by combining data augmentation, Gibbs sampling, and Metropolis--Hastings steps within a general MCMC framework.
The key idea is to replace a single difficult draw from the joint posterior by a sequence of easier draws from the full conditional distributions of parameter blocks, thereby generating a Markov chain whose stationary distribution is the target joint posterior.
We adopt their methodology by factoring in the posterior probability $\mbp\left(\bmu^{(t)}, \bm \Sigma^{(t)} \mid R_n^{c_n}\right) \propto \mbp\left(R_n^{c_n} \mid \bmu^{(t)}, \bm \Sigma^{(t)}\right) \mbp\left(\bmu^{(t)}\right) \mbp\left(\bm \Sigma^{(t)}\right)$. 
We refer to Appendix \ref{HBM} for further details on this implementation.



\Copy{rev:bayesian-performance}{To evaluate the performance of the EM algorithm relative to the Bayesian approach, we run experiments across different numbers of products, $I\in\{2,3,4,5,6\}$, with $N=1000$ training transactions.
We compute the test log-likelihood for both methods on a test set with $N=2000$.
Both algorithms are initialized with the same parameters for three standard deviations from the true mean (midway point)
and run for $1000$ iterations.
We define the log-likelihood score as
$1 - (\mathcal L_{\text{model}} - \mathcal L_{\text{exact}})/\mathcal L_{\text{exact}}$,
so a value closer to $100\%$ indicates a better fit.
Table~\ref{tab:mh_em_summary_full} summarizes the performance of the two methods.
Both approaches achieve high log-likelihood scores. 
The EM algorithm consistently produces smaller $\ell_1$-errors across all settings. 
For the MH benchmark, the posterior distribution is approximated using Markov chain samples after burn-in. The $\ell_1$-error is computed using the estimated posterior mean of $\bmu$ and $\bsig$.
The difference is more pronounced in higher-dimensional settings. These results suggest that similar likelihood values do not necessarily imply similar recovery of the underlying valuation parameters. One possible explanation is that the likelihood surface under bundle-choice observations may contain multiple regions with similar likelihood values, making posterior exploration challenging.
In contrast, the EM method directly exploits the latent-valuation structure through likelihood-based updates, which appears to provide more stable parameter recovery in our experiments.}


\begin{table}[htbp]
\centering
\caption{Comparison of MH and EM algorithms on test log-likelihood score and $\ell_1$-error.}
\label{tab:mh_em_summary_full}
\small
\begin{tabular}{c rr rr}
\hline
 & \multicolumn{2}{c}{Log-likelihood Score} & \multicolumn{2}{c}{$\ell_1$-error}  \\
\cline{2-3}\cline{4-5}
$I$ & Bayesian & EM & Bayesian & EM \\
\hline
2 & 98.06\% & 98.66\%  & 0.2203 & 0.1289  \\
3 & 99.63\% & 99.46\% & 0.1722 & 0.1187   \\
4 & 98.65\% & 98.26\%  & 0.1191 & 0.0957   \\ 
5 & 98.07\% & 99.42\%  & 0.1323 &  0.0969 \\ 
6 & 96.35\% & 98.19\% & 0.1429 & 0.1069  \\
\hline
\end{tabular}
\end{table}

\paragraph{Comparison to the MNL model.}
We conduct numerical experiments comparing our Gaussian-based model with the multinomial logit (MNL) model, one of the most widely used benchmarks in discrete choice analysis.
\Copy{rev:MNL_setting}{
To set up the MNL model in the bundle setting, we treat each bundle as a separate product in the choice probability.
The expected utility of the bundle is the sum of the included products\citep{jedidi2003measuring,ma2016constructing}.
For example, with two products $\{A,B\}$ with expected utilities $\mu_A$ and $\mu_B$ and bundle price $p_{AB}$,
the MNL choice probability of the bundle is $\exp(\mu_A+\mu_B-p_{AB})/(1+\exp(\mu_A-p_A)+\exp(\mu_B-p_B)+\exp(\mu_A+\mu_B-p_{AB}))$.
}
\blue{We refer to Appendix \ref{app:mnl} for further details.}

We generate the training data according to the base model.
Thus, the MNL model is misspecified.
We estimate the parameters using our method and the standard maximum-likelihood estimator for the MNL model. 
To evaluate the goodness-of-fit, we construct a test dataset of $m$ menus, where each menu consists of a bundle-offering matrix $\bXn$ and an associated bundle-price vector $\bm p_n$, using the same procedure as the training data.
We let $m=100$.
For each menu, we compute the ground-truth choice probability $\mbp^{\text{true}}$
and measure the root mean squared error (RMSE) of an estimator
$$
\text{RMSE}=\frac 1m\sum_{n=1}^m\sqrt{
\frac 1{J_n+1}
\sum_{j=1}^{J_n+1}
\left( \mbp_{nj}^{\text{true}}-\mbp_{nj}^{\text{model}} \right)^2
}
$$
over the $J_n$ products/bundles in a menu and $m$ menus in the test set.
We report the out-of-sample log-likelihood and the RMSE in Table~\ref{tab:gaussianmnl_summary}.

\begin{table}[htbp]
\centering
\caption{Comparison of EM algorithm and the (misspecified) MNL model in terms of the log-likelihood and RMSE in the testing set. The experiment is conducted for different product dimensions $I$ and $N=2500$ training transactions.}
\label{tab:gaussianmnl_summary}
\small
\begin{tabular}{c rr rr}
\hline
 & \multicolumn{2}{c}{Log-likelihood Score} & \multicolumn{2}{c}{RMSE} \\
\cline{2-3}\cline{4-5}
$I$ & MNL & EM & MNL & EM \\
\hline
  2 & 85.09\% & 99.78\% & 0.11334 & 0.01205\\
  3 & 86.17\% & 99.58\% & 0.08597 & 0.01013\\
  4 & 72.06\% & 99.81\% & 0.10385 & 0.01133\\
  5 & 78.22\% & 99.05\% & 0.09833 & 0.00684\\
  6 & 79.44\% & 99.45\% & 0.09722 & 0.02022\\
\hline
\end{tabular}
\end{table}

Because of misspecification, particularly the failure to incorporate the correlation between the valuations of bundles and the included products, the MNL model has lower predictive performance.
Indeed, the MNL model treats the (Gumbel) random utilities of bundles and the included products as independent.
This result illustrates the limitation of adapting MNL directly to the bundle setting.



\end{sfnadded}

{
}

{
}

{
}

%% file: 4_Numerical.tex
\subsection{Product Synergy}\label{subsec:numerical_synergy}
\begin{sfnadded}
In this subsection, we test the algorithm in Section~\ref{sec:synergy}.  
The experimental setup is similar to that of the base model (see Section~\ref{sec:numerical}), including the product set, offered bundles, price menus, and customers' individual-product valuation distribution.
We consider $I \in\{2, 3, 4, 5, 6\}$ products and $N \in\{1000, 1500, 2000, 2500, 3000, 5000, 10000\}$ transactions.
The key difference from the base model is that each consumer's utility now includes pairwise product interactions represented by the synergy matrix $\bA$.
Specifically, the upper-triangular entries of $\bm A$ are randomly sampled from a uniform distribution $\mathcal{U}[-3, 3]$, while the diagonal and lower-triangular entries are fixed to zero. 

We run Algorithm~\ref{alg:em-synergy} on these datasets to jointly estimate the parameters $(\bmu, \bsig, \bA)$. 
We use the $\ell_1$-error
$\left(
\|\bmu^{(t+1)}-\bmu^*\|_1+\|\bA^{(t+1)}-\bA^*\|_1+\|\bsig^{(t+1)}-\bsig^*\|_1
\right)/
\left(\left(3I^2 + I\right)/2\right)$ to measure estimation accuracy and track algorithmic convergence.

\paragraph{EM Convergence.}
Figure~\ref{fig:l1error_synergy} illustrates the convergence of Algorithm~\ref{alg:em-synergy} under different problem settings.
Algorithm~\ref{alg:em-synergy} exhibits stable convergence in the reported instances.
Although instances with larger $I$ require more iterations to reach convergence, the overall convergence pattern is similar across product dimensions.
Larger samples lead to smaller estimation errors and faster convergence.
{Panel~\ref{fig:EMperformance_trajectory} illustrates the trajectory of $(\bmu^{(t)},\bA^{(t)})$ for an instance of $I=2$ products and $N=10000$ transactions during Algorithm~\ref{alg:em-synergy}.
Starting from a distant initialization, the algorithm rapidly moves into a neighborhood of the true parameter values and then gradually refines the estimates until convergence. 
This pattern indicates that, in this instance, the algorithm moves toward the region around the true parameters and then exhibits stable local refinement.
}
\begin{figure}[ht]
    \begin{center}
    \subfigure[$N=10{,}000$ observations\label{fig:l1errorbyproduct}]{%
        \begin{tikzpicture}
        \begin{axis}[
            width=0.34\linewidth, height=5cm,
            xlabel=\# iterations,
            ylabel=$\ell_1$-error,
            xmin=100, xmax=1000,
            ymin=0, ymax=50,
            xtick={100,300,500,700,900},
            legend pos=north east,
            legend style={font=\tiny},
            grid=major, grid style={dashed,gray!30},
        ]
        \addplot[mark=none, solid, line width=1.5pt, color=red]
            table[x=t, y=I2, col sep=comma]{avg_l1_byI_N10000_every50.csv};
        \addplot[mark=none, dashed, line width=1.5pt]
            table[x=t, y=I3, col sep=comma]{avg_l1_byI_N10000_every50.csv};
        \addplot[mark=none, densely dotted, line width=1.5pt, color=green!70!black]
            table[x=t, y=I4, col sep=comma]{avg_l1_byI_N10000_every50.csv};
        \addplot[mark=none, dash dot, line width=1.5pt, color=orange!80!black]
            table[x=t, y=I5, col sep=comma]{avg_l1_byI_N10000_every50.csv};
        \addplot[mark=none, loosely dashed, line width=1.5pt, color=purple!80!black]
            table[x=t, y=I6, col sep=comma]{avg_l1_byI_N10000_every50.csv};
        \legend{$I=2$, $I=3$, $I=4$, $I=5$, $I=6$}
        \end{axis}
        \end{tikzpicture}%
    }%
    \subfigure[$I=2$ products\label{fig:l1errorbytran}]{%
        \begin{tikzpicture}
        \begin{axis}[
            width=0.34\linewidth, height=5cm,
            xlabel=\# iterations,
            xmin=100, xmax=1000,
            ymin=0, ymax=50,
            xtick={100,300,500,700,900},
            legend pos=north east,
            legend style={font=\tiny},
            grid=major, grid style={dashed,gray!30},
        ]
        \addplot[mark=none, solid, line width=1pt, color=red]
            table[x=t, y=N10000, col sep=comma]{avg_l1_byN_I2_every50.csv};
        \addplot[mark=square*, densely dashed, line width=0.8pt, mark repeat=4, mark options={solid,scale=0.5},mark phase=1,]
            table[x=t, y=N5000, col sep=comma]{avg_l1_byN_I2_every50.csv};
        \addplot[mark=triangle*, densely dotted, line width=0.8pt, color=orange!80!black, mark repeat=4, mark options={solid,scale=0.5},mark phase=2,]
            table[x=t, y=N2500, col sep=comma]{avg_l1_byN_I2_every50.csv};
        \addplot[mark=diamond*,dash pattern=on 8pt off 2pt on 1pt off 2pt, line width=0.8pt, color=brown!80!black, mark repeat=4, mark options={solid,scale=0.5},mark phase=3,]
            table[x=t, y=N1000, col sep=comma]{avg_l1_byN_I2_every50.csv};
        \legend{$N=10000$, $N=5000$, $N=2500$, $N=1000$}
        \end{axis}
        \end{tikzpicture}%
    }%
    \subfigure[Trajectory of $(\bmu^{(t)},\bA^{(t)})$\label{fig:EMperformance_trajectory}]{%
        \includegraphics[width=0.30\linewidth]{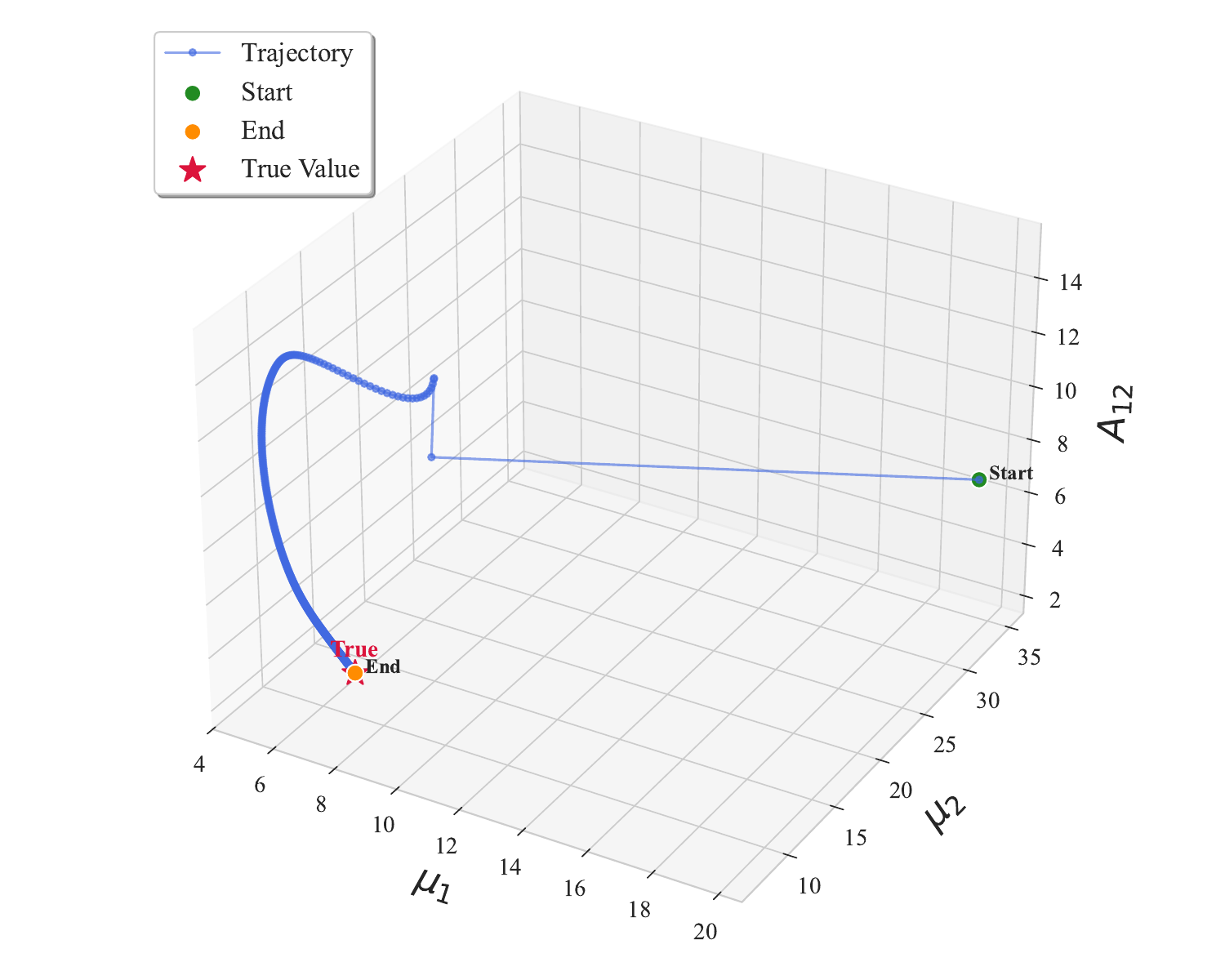}%
    }
    \end{center}
    \caption{The $\ell_1$-errors and the parameter trajectory in the iterations of Algorithm~\ref{alg:em-synergy} when there is product synergy.\label{fig:l1error_synergy}}
  \end{figure}

\paragraph{EM Performance.}
Panel~\ref{fig:EMperformance} illustrates recovery of $(\bmu,\bsig)$, which determine the distribution of the two individual-product valuations.
The contours of the estimated and ground-truth distributions are largely aligned, suggesting that the algorithm recovers both the marginal distribution of the utilities and the cross-product correlations.
{
We also study the approximation quality of the sigmoid smoothing in \eqref{eq:sigmoid_approx} relative to the original Monte Carlo EM algorithm, Algorithm~\ref{alg:em-imp}. When the synergy matrix $\bA=\bm 0$, the extended model reduces exactly to the base model introduced in Section~\ref{sec:base-model}. This setting therefore provides a natural benchmark to evaluate the approximation error introduced by the sigmoid method.
To evaluate the approximation quality of the sigmoid method, we generate synthetic datasets with $N=5000$ transactions consisting of two products and one bundle under the base model with $\bA=\bm 0$. Both Algorithm~\ref{alg:em-imp} and Algorithm~\ref{alg:em-synergy} are then applied to estimate $(\bmu,\bsig)$.
The data-generating process follows the setting of the base model described in Section~\ref{sec:numerical}. 
Since the true data-generating process contains no bundle synergy effects, any performance difference between the two methods reflects the approximation error caused by replacing the indicator function with the sigmoid smoothing.
Both algorithms are initialized from the same starting point to ensure a fair comparison. 
Figure~\ref{fig:2D_Gaussian_noA} illustrates the estimation results for the case $I=2$ and $N=5000$. 
The figure plots the ground-truth distribution together with the distributions implied by the parameters estimated by MCEM and the sigmoid-based EM algorithm. 
The two estimated distributions are close to the ground-truth distribution, suggesting that the sigmoid approximation introduces little additional estimation error in this setting.
}
\begin{figure}[htbp]
    \begin{center}
    \subfigure[Dataset with $A_{12}=2.2492$ and $\widehat A_{12}=2.3129$\label{fig:EMperformance}]
    {%
        \begin{tikzpicture}
        \begin{axis}[
            width=0.46\linewidth, height=0.35\linewidth,
            xlabel={$\mu_1$}, ylabel={$\mu_2$},
            xmin=0,  xmax=14,
            ymin=7,  ymax=16,
            grid=both, grid style={dashed, gray!40},
            tick label style={font=\small},
            label style={font=\large},
            legend cell align=left,
            legend style={font=\small, draw=black, fill=white,
                          fill opacity=0.85, text opacity=1,
                          at={(0.98,0.98)}, anchor=north east},
        ]
        \addplot[color=blue, very thick, solid, no marks,
                 samples=200, smooth, domain=0:360, variable=\t]
            ({6.5463 + 1*1.75230*cos(\t)},
             {11.5616 + 1*(-0.79142)*cos(\t) + 1*0.90225*sin(\t)});
        \addlegendentry{Ground Truth}
        \addplot[color=blue, very thick, solid, no marks, forget plot,
                 samples=200, smooth, domain=0:360, variable=\t]
            ({6.5463 + 3*1.75230*cos(\t)},
             {11.5616 + 3*(-0.79142)*cos(\t) + 3*0.90225*sin(\t)});
        \addplot[color=red, very thick, dashed, no marks,
                 samples=200, smooth, domain=0:360, variable=\t]
            ({6.5689 + 1*1.73756*cos(\t)},
             {11.4730 + 1*(-0.77672)*cos(\t) + 1*0.75280*sin(\t)});
        \addlegendentry{EM Estimation}
        \addplot[color=red, very thick, dashed, no marks, forget plot,
                 samples=200, smooth, domain=0:360, variable=\t]
            ({6.5689 + 3*1.73756*cos(\t)},
             {11.4730 + 3*(-0.77672)*cos(\t) + 3*0.75280*sin(\t)});
        \end{axis}
        \end{tikzpicture}%
    }%
    \subfigure[Dataset with $\bA=\bm 0$\label{fig:2D_Gaussian_noA}]
  {%
      \begin{tikzpicture}
      \begin{axis}[
          width=0.46\linewidth, height=0.35\linewidth,
          xlabel={$\mu_1$}, ylabel={$\mu_2$},
          xmin=4,  xmax=16,
          ymin=3,  ymax=16,
          grid=both, grid style={dashed, gray!40},
          tick label style={font=\small},
          label style={font=\large},
          legend cell align=left,
          legend style={font=\small, draw=black, fill=white,
                        fill opacity=0.85, text opacity=1,
                        at={(0.98,0.98)}, anchor=north east},
      ]
      \addplot[color=blue, very thick, solid, no marks,
               samples=200, smooth, domain=0:360, variable=\t]
          ({9.5461 + 1*1.52022*cos(\t)},
           {8.9266 + 1*(-0.41178)*cos(\t) + 1*1.50949*sin(\t)});
      \addlegendentry{Ground Truth}
      \addplot[color=blue, very thick, solid, no marks, forget plot,
               samples=200, smooth, domain=0:360, variable=\t]
          ({9.5461 + 3*1.52022*cos(\t)},
           {8.9266 + 3*(-0.41178)*cos(\t) + 3*1.50949*sin(\t)});

      \addplot[color=green!60!black, very thick, dashed, no marks,
               samples=200, smooth, domain=0:360, variable=\t]
          ({9.4996 + 1*1.55740*cos(\t)},
           {8.8956 + 1*(-0.39515)*cos(\t) + 1*1.51088*sin(\t)});
      \addlegendentry{EM Estimation}
      \addplot[color=green!60!black, very thick, dashed, no marks, forget plot,
               samples=200, smooth, domain=0:360, variable=\t]
          ({9.4996 + 3*1.55740*cos(\t)},
           {8.8956 + 3*(-0.39515)*cos(\t) + 3*1.51088*sin(\t)});

      \addplot[color=red, very thick, dashed, no marks,
               samples=200, smooth, domain=0:360, variable=\t]
          ({9.5017 + 1*1.52299*cos(\t)},
           {8.8842 + 1*(-0.41353)*cos(\t) + 1*1.49863*sin(\t)});
      \addlegendentry{Sigmoid\_EM Estimation}
      \addplot[color=red, very thick, dashed, no marks, forget plot,
               samples=200, smooth, domain=0:360, variable=\t]
          ({9.5017 + 3*1.52299*cos(\t)},
           {8.8842 + 3*(-0.41353)*cos(\t) + 3*1.49863*sin(\t)});
      \end{axis}
      \end{tikzpicture}%
  }
    \end{center}
    \caption{Estimation performance of the Gaussian distribution $\Nscr(\bmu,\bsig)$ with and without product synergy.\label{fig:2D_Gaussian_combined}}
  \end{figure}

\end{sfnadded}

\begin{lightgray}
{
}
\end{lightgray}

\begin{sfnadded}
To further investigate the statistical properties of Algorithm~\ref{alg:em-synergy}, we evaluate the $\ell_1$-error across $N\in \{1000,2500,5000,10000\}$ and $I\in \{2,3,4,5,6\}$.
As reported in Table~\ref{tab:synergy_l1error}, the $\ell_1$-error generally decreases as the number of transactions $N$ increases for a fixed product dimension $I$.
For larger $I$, the estimation problem becomes more challenging because the number of valuation, covariance, and synergy parameters increases.


\begin{table}[htbp]
  \centering
  \caption{The $\ell_1$-error of Algorithm~\ref{alg:em-synergy} under bundle synergy.}
  \label{tab:synergy_l1error}
  \begin{tabular}{crrrrr}
    \toprule
    $N$ & $I=2$ & $I=3$ & $I=4$ & $I=5$ & $I=6$ \\
    \midrule
    1000  & 0.2440 & 0.2448 & 0.1050 & 0.4645 & 0.3095 \\
    2500  & 0.1380 & 0.1764 & 0.0957 & 0.3664 & 0.2386 \\
    5000  & 0.0790 & 0.0957 & 0.1111 & 0.1929 & 0.1939 \\
    10000 & 0.0689 & 0.0830 & 0.0452 & 0.0329 & 0.0516 \\
    \bottomrule
  \end{tabular}
\end{table}

{
}
\end{sfnadded}

{
}

\subsection{Censored Data}\label{subsec:numerical_censored}
In this section, we evaluate how well the proposed method recovers model parameters when no-purchase observations are censored from the sales data.
\begin{sfnadded}
Starting from the data generation process for the base model experiments, we remove all no-purchase observations, resulting in a censored dataset.
The proportion of removed no-purchase transactions ranges from approximately 10.84\% to 17.54\% across different datasets.
We fix the number of products at $I=2$ and the number of uncensored transactions at $N=2500$. We generate five datasets with different parameters $(\bmu,\bsig)$ and estimate the parameters using the EM method for the censored data and the complete data.
We compute the log-likelihood for both methods on a test set with $N=2000$.
\begin{table}[htbp]
\centering
\caption{Comparison of EM algorithms on censored and complete data for different datasets.}\label{tab:censored}
\begin{tabular}{c|rr|rr}
\toprule
& \multicolumn{2}{c|}{$\ell_1$-error} & \multicolumn{2}{c}{Log-likelihood Score} \\
\text{Dataset} & Censored & Complete & Censored & Complete \\
\midrule
1 & 0.0819 & 0.0329 & 99.55\%  &100.03\% \\
2 & 0.2583 & 0.1672 & 99.95\% & 99.50\% \\
3 & 0.9363 & 0.0908 & 97.71\% & 99.84\% \\
4 & 0.4899 & 0.1577 & 99.29\% & 99.73\% \\
5 & 0.0556 & 0.0402 & 99.78\% & 99.31\% \\
\bottomrule
\end{tabular}
\end{table}




\begin{figure}[ht]                                             
      \begin{center}                                
      \begin{tikzpicture}                             
      \begin{axis}[
          width=0.6\linewidth, height=5.5cm,
          xlabel=\# iterations,
          ylabel=Average log-likelihood,
          xmin=0, xmax=100,
          ymin=-1.05, ymax=-0.30,
          xtick={0,20,40,60,80,100},
          legend pos=south east,
          legend style={font=\small},
          grid=major, grid style={dashed,gray!30},
      ]
      \addplot[mark=none, solid, line width=1.5pt, color=red]
          table[x=t, y=censored_em, col sep=comma]%
{avg_llh_bymethod_censored_multi_fixdiscount10.csv};
      \addplot[mark=none, densely dotted, line width=1.5pt, color=green!70!black]
          table[x=t, y=em_complete, col sep=comma]%
          {avg_llh_bymethod_censored_multi_fixdiscount10.csv};
      \addplot[domain=0:100, color=black, dash dot, line width=1.2pt]{-0.3584};
      \legend{Censored EM, EM with complete data, Exact}
      \end{axis}
      \end{tikzpicture}
      \end{center}
      \caption{Convergence of the average log-likelihood for the censored and complete-data EM algorithms with $N=2500$ total observations and $I=2$ products.\label{fig:censored_llh}}
  \end{figure}

The results in Table~\ref{tab:censored} show that 
the censored EM algorithm remains close to the complete-data benchmark in test log-likelihood, although its parameter error is higher, as expected from the information loss induced by censoring.
Across all datasets, both methods achieve high test log-likelihood scores, while the censored-data estimates have larger $\ell_1$-errors than the complete-data estimates.
We show the converging trajectories of log-likelihood for both methods in Figure~\ref{fig:censored_llh}. 
Taken together, these findings suggest that censoring has a more visible effect on parameter recovery than on out-of-sample likelihood in these experiments.
\end{sfnadded}

{
}

\subsection{Real Data} \label{subsec:realdata}
{
}

\begin{sfnadded}
    This subsection evaluates the performance of our proposed EM framework by comparing several variants of our model, including the base EM, 
    Gaussian mixture model (GMM), and the bundle synergy model, against two benchmark approaches: the multinomial logit (MNL) model and an MH-based Bayesian method.

    We conduct the empirical analysis using the JD.com dataset \citep{shen2020jd}, which provides detailed transaction and clickstream information. The dataset consists of 486,928 transactions involving 9,159 unique stock-keeping units (SKUs) shipped to 60 districts in March 2018. Each transaction record includes product prices and various promotional mechanisms, such as direct discounts, bundle discounts, and gift incentives. In addition, the dataset contains customers’ click histories across SKUs during the same period, which allows us to infer customers’ consideration sets. Notably, only approximately 1\% of customers in the dataset purchase bundles under promotional schemes.
\end{sfnadded}


\Copy{rev:realdata_processing1}{
To conduct our analysis, we focus on distribution center 5, which has the highest transaction volume during the first week of March 2018 ($03/01-03/08$). We select the top 10 products that are part of bundle sales promotions and identify bundles by considering the combinations of products purchased in a single transaction that incur a bundle sales promotion.
}
\Copy{rev:realdata_processing2}{
To form users' consideration sets and identify the number of ``no-purchase customers," we use click data from customers who reside close to distribution center 5.
For customers who purchase a single product or a bundle, we collect their clicks from the start time of the study until the timestamp of their purchase. For customers who end up purchasing another product, we collect all their clicks during the study time period. We exclude customers who do not click on any of the selected products and consider the remaining customers as the ``no-purchase" group. We assume that customers observe the combination of single products they click on if the corresponding bundle is offered by JD.com.
}
\Copy{rev:realdata_price}{
To construct prices for purchased customers, we subtract the direct discount from the original price of each single product purchased by the customer. For bundles, we sum the original prices of the component products and subtract the applicable direct discounts and bundle promotions. However, for no-purchase customers, we do not have a price reference. To address this, we calculate the average original prices, direct discounts, and bundle promotions per day and use them to impute prices for the relevant customer choices. Overall, we have 5,004 transaction records, with approximately 50\% of the dataset representing no-purchase, 15\% representing bundle purchases, and 36\% representing single product purchases.
}

\begin{sfnadded}
\Copy{rev:covariance_matrix}{
For computational tractability, we restrict the covariance matrices to be diagonal in the real-data experiment.
Thus, this specification does not estimate cross-product valuation correlations in the real data.
}
To obtain initial values, we randomly sample 1,000 observations from the training set and run each algorithm for 50 iterations. We then use the estimated parameters as the starting point for the main algorithms.
In the main training phase, we stop when the change in average log-likelihood between consecutive iterations is below $10^{-6}$.
\Copy{rev:80-20-train_test}{
Table~\ref{tab:model_comparison} reports the mean and standard deviation over $5$ independent random 80/20 out-of-sample train-test splits.
}
Top-$X$ accuracy measures the fraction of test observations for which the actual purchase is among the model's top-$X$ predicted alternatives.
    \begin{table}
        \centering
        \caption{Performance in the real data.}
        \label{tab:model_comparison}
        \begin{tabular}{lccccc}
        \toprule
        Model/Method & Log-likelihood & RMSE & Top-1 Accuracy & Top-3 Accuracy & Top-5 Accuracy \\
        \midrule
        MH      & -1.2346 (0.3187) & 0.0524 (0.0016) & 0.6068 (0.0097) & 0.8987 (0.0024) & 0.9438 (0.0059) \\
        MNL     & -4.3827 (0.2050) & 0.0706 (0.0090) & 0.6493 (0.0144) & 0.9168 (0.0043) & 0.9562 (0.0027) \\
        Base EM    & -0.8774 (0.1640) & 0.0315 (0.0025) & 0.6737 (0.0133) & 0.9430 (0.0053) & 0.9753 (0.0041) \\
        Synergy & -0.8245 (0.0363) & \textbf{0.0315 (0.0025)} & \textbf{0.6763 (0.0123)} & 0.9421 (0.0046) & \textbf{0.9794 (0.0018)} \\
        GMM     & \textbf{-0.7733 (0.0217)} & 0.0355 (0.0041) & 0.6662 (0.0234) & \textbf{0.9471 (0.0012)} & 0.9768 (0.0023) \\
        \bottomrule
        \end{tabular}
    \end{table}
    
Table~\ref{tab:model_comparison} shows that all three proposed models (Base, Synergy, and GMM) outperform the MH and MNL benchmarks across the reported evaluation metrics, indicating that the proposed framework captures consumer choice behavior more effectively in this dataset.
Among the proposed models, the Base model already achieves strong predictive performance and provides a significant improvement over the benchmark methods. The Synergy and GMM models further extend the Base model by incorporating additional behavioral structures, leading to modest but consistent improvements in selected performance metrics. 
In particular, the GMM model achieves the best out-of-sample log-likelihood and Top-3 accuracy, while the Synergy model attains the highest Top-1 and Top-5 accuracy. 
These results suggest that the additional flexibility introduced by the Synergy and GMM formulations can capture additional patterns in consumer choice behavior that are not fully represented in the Base model.
Nevertheless, the relatively small performance gap among the three proposed models indicates that the Base model alone is already capable of explaining much of the observed choice behavior.
\end{sfnadded}

%% file: Appendix_Bundle.tex
\newpage
	%
	   \begin{center}
	  {\bf\large
	 Appendix to \\``Learning Customer Preferences from Bundle Sales Data"}
	\end{center}

	\renewcommand\thesection{\Alph{section}}

	\setcounter{equation}{0}
	\setcounter{section}{0}
	\setcounter{lem}{0}
	\setcounter{defn}{0}
	\setcounter{thm}{0}
	\setcounter{prop}{0}

	\numberwithin{lem}{section}
	\numberwithin{defn}{section}
	\numberwithin{equation}{section}
	\numberwithin{thm}{section}
	\numberwithin{prop}{section}

{
}

\section{Detailed EM Algorithm for Model Extensions}\label{app:extensions}
This appendix provides the detailed derivations of the EM algorithm for the model extensions in Section~\ref{sec:extension}.

\subsection{Product Synergy in a Bundle}\label{app:synergy}
We use the same notation as in Section~\ref{sec:synergy}: $\bm D = \{(p_n^1,\dots,p_n^{J_n},c_n)\}_{n=1}^N$ denotes the observed data, and $\bm\theta=(\bmu,\bsig,\bA)$ denotes the parameters to be estimated, where $\bA$ is the bundle synergy matrix.

\paragraph{E-step.} 
Following the standard EM framework, we evaluate the expectation of the complete-data log-likelihood with respect to the conditional distribution of the latent variables $\bm v$. 
Compared with the base model, the key difference arises from the replacement of the indicator function defining the IC polyhedron. 
The likelihood can be written as 
$\Lscr(\bm\theta;\bm D) 
= \prod_{n=1}^N\int_{K_n^{c_n}(\mathbf{A})} f(\bm v|\bm\mu,\bm\Sigma) \mathrm{d}\bm v 
=\prod_{n=1}^N\int  \mI \{\bm v \in K_n^{c_n}(\mathbf{A})\} f(\bm v|\bm\mu,\bm\Sigma) \mathrm{d}\bm v$,
where $f(\cdot)$ denotes the PDF of the multivariate normal distribution.
The complete-data log-likelihood can therefore be written as
\begin{equation}\label{eq:likelihood_synergy_complete-data}
\begin{aligned}
    \ell(\bm\theta;\bm D,\bm Z)
    =\sum_{n=1}^N \log f(\bm v_n\mid \bmu,\bsig)
    +\sum_{n=1}^N \log \mI\{\bm v_n\in K_n^{c_n}(\bA)\}.
\end{aligned}
\end{equation}

Under the proposed sigmoid smoothing approximation with synergy matrix $\bA$, the indicator function in E-step is replaced by the sigmoid product approximation in Eq.~(\ref{eq:sigmoid_approx}). As a result, the hard truncation of the feasible region is replaced by a continuous weighting term. The conditional density used in the E-step therefore becomes $\Lscr(\bm\theta;\bm D)\approx \prod_{n=1}^N\int \prod_{j=1}^{J_n} \sigma\left(\frac{\Delta U_{nj}(\bm v; \mathbf{A})}{\lambda}\right) f(\bm v|\bm\mu,\bm\Sigma) \mathrm{d}\bm v. $
The complete-data log-likelihood can therefore be written as
\begin{equation}\label{eq:app-likelihood_sigmoid_complete-data}
\begin{aligned}
    \ell(\bm\theta;\bm D,\bm Z)=\sum_{n=1}^N\left(\log f(\bm v_n \mid \bmu, \bsig) +\sum_{j=1}^{J_n} \log \sigma\left(\frac{\Delta U_{nj}(\bm v_n; \bA)}{\lambda}\right)\right).
\end{aligned}
\end{equation}

This replaces the truncated Gaussian posterior in the base model with a smoothly weighted Gaussian density, where the sigmoid terms softly enforce the incentive compatibility constraints. Substituting this conditional density into the expectation yields the smoothed $Q'$ function
\begin{equation}\label{eq:app-smoothed_Q}
\begin{aligned}
Q'(\bm\theta \mid \bm\theta^{(t)})
&=\sum_{n=1}^{N}\int q_n(\bm v;\bm\theta^{(t)})
\left(
\log f(\bm v \mid \bmu,\bsig)+\sum_{j=1}^{J_n}\log \sigma\left(
\frac{\Delta U_{nj}(\bm v;\bA)}{\lambda}
\right)
\right)
\mathrm{d}\bm v\\
\end{aligned},
\end{equation}
where 
\begin{equation}
    q_n(\bm v; \bth^{(t)})= 
    \frac
{
f(\bm v \mid \bmu^{(t)}, \bsig^{(t)}) \prod_{j=1}^{J_n}\sigma(\frac{\Delta U_{nj}(\bm v; \bA^{(t)})}{\lambda}) 
}
{
{\int f(\bm\xi \mid \bmu^{(t)}, \bsig^{(t)})\prod_{j=1}^{J_n}\sigma(\frac{\Delta U_{nj}(\bm\xi; \bA^{(t)})}{\lambda})  \mathrm{d}\bm\xi}
}.
\end{equation}

Relative to the original EM formulation, the resulting $Q'$ function differs in two important aspects. First, the truncation of the latent utility distribution is replaced by a smooth probabilistic weighting determined by the sigmoid terms. Second, the log-likelihood now contains an additional contribution from the smoothed IC constraints, which introduces explicit dependence on the synergy parameter $\bA$ inside the integrand. This modification enables gradient-based optimization with respect to $\bA$ while preserving the overall EM structure.

Similar to the base model, we substituting the Monte Carlo approximation into Eq.~(\ref{eq:app-smoothed_Q}) yields the empirical smoothed $\hat {Q'}$ function
\begin{equation}\label{eq:app-smoothed_Qhat}
\begin{aligned}
\hat {Q'}(\bm\theta \mid \bm\theta^{(t)})&=\sum_{n=1}^N\sum_{l=1}^L
\overline\omega_{nl}\left(
\log f(\bm v_n^{(l)}|\bmu,\bsig)+\log \omega(\bm v_n^{(l)};\bA)
\right)\\
&=C-\frac{N}{2}\log|\bsig|+\sum_{n=1}^N\sum_{l=1}^L
\overline\omega_{nl}\left(
-\frac{1}{2}(\bm v_n^{(l)}-\bmu)^\top\bsig^{-1}(\bm v_n^{(l)}-\bmu)+\log \omega(\bm v_n^{(l)};\bA)
\right),
\end{aligned}
\end{equation}
where $C$ is constant with respect to $\bm \theta$ and
\begin{equation}\label{eq:app-sigmoid_vals}
\begin{aligned}
    \omega(\bm v;\bA) &\triangleq\prod_{j=1}^{J_n}\sigma\left(
    \frac{ \Delta U_{nj}(\bm v; \bA) }{\lambda}
    \right),\\
    \overline\omega_{nl} &\triangleq \frac{ \omega(\bm v_{n}^{(l)};\bA^{(t)})} {\sum_{l=1}^L \omega(\bm v_{n}^{(l)};\bA^{(t)})}.
\end{aligned}   
\end{equation}

\paragraph{M-step.} 
In the M-step, we maximize the smoothed Monte Carlo objective $\hat {Q'}(\bm\theta\mid\bm\theta^{(t)})$ obtained in the E-step with respect to $(\bmu,\bA,\bsig)$. The objective function is concave in each parameter block. 

\textbf{Update for $\bmu$ and $\bsig$.}
Taking the gradient of $\hat{Q'}$ with respect to $\bmu$ gives
\begin{equation}
    \nabla_{\bmu}\hat {Q'} = \sum_{n=1}^N\sum_{l=1}^L \overline\omega_{nl}\bsig^{-1} (\bm v_n^{(l)}-\bmu).
\end{equation}
The first order condition yields the closed-form update of $\bmu$ and $\bsig$
\begin{equation}\label{eq:mut1_sigmoid}
\bmu^{(t+1)}=\sum_{n=1}^N\sum_{l=1}^L\overline\omega_{nl}\bm v_n^{(l)}.
\end{equation}
\begin{equation}\label{eq:Sigmat1_sigmoid}
\bsig^{(t+1)}=\sum_{n=1}^N\sum_{l=1}^L\overline\omega_{nl}(\bm v_n^{(l)}-\bmu^{(t+1)})(\bm v_n^{(l)}-\bmu^{(t+1)})^\top.
\end{equation}

\textbf{Gradient with respect to $\bA$.}
Unlike $\bmu$ and $\bsig$, the synergy matrix $\bA$ does not admit a closed-form solution in the M-step. 
This is because $\bA$ appears inside the nonlinear term 
\begin{equation*}
    \log \sigma\left(\frac{\Delta U_{nj}(\bm v;\bA)}{\lambda}\right),
\end{equation*}
which prevents the first-order optimality condition from yielding an explicit solution for $\bA$. 
Consequently, the maximization with respect to $\bA$ must be performed using gradient-based optimization, where the gradient is given in Theorem~\ref{thm:grad_A}.

In summary, the M-step produces closed-form updates for $\bmu$ and $\bsig$, while $\bA$ is updated using the gradient of the smoothed objective. Algorithm~\ref{alg:em-synergy} summarizes the full EM procedure.
If the importance sampling framework used in the E-step is applied, the parameter updates can be written in terms of the normalized importance weights $w_{nl}$.The normalized weights are defined as
\begin{equation}\label{eq:IS_sigmoid_weights}
    \tilde w_{nl}\triangleq\frac
    {w_{nl}\; \overline\omega_{nl} }
    {\sum_{l=1}^{L} w_{nl}\; \overline\omega_{nl}},
\end{equation}
where $\omega_{nl}$ is the importance sampling density ratio and $w_{nl}$ is the sigmoid feasibility weight.

\begin{equation}\label{eq:muSigA_sigmoid_IS}
\begin{aligned}
    \bmu^{(t+1)} &=
    \sum_{n=1}^{N}\sum_{l=1}^{L}\tilde w_{nl}\bm v_n^{(l)} \\
    \bsig^{(t+1)} &=
    \sum_{n=1}^{N}\sum_{l=1}^{L}\tilde w_{nl} (\bm v_n^{(l)}-\bmu^{(t+1)})(\bm v_n^{(l)}-\bmu^{(t+1)})^\top \\
    \nabla_{\bA}\hat Q' &=
    \frac{1}{\lambda}\sum_{n=1}^{N}\sum_{j=1}^{J_n}\left(
    \frac{\partial \Delta U_{nj}(\bA)}{\partial \bA}
    \sum_{l=1}^{L}\tilde w_{nl}
    \left(
    1-\sigma\left(\frac{\Delta U_{nj}(\bm v_n^{(l)};\bA)}{\lambda}\right)
    \right)
    \right).
\end{aligned}
\end{equation}

\begin{algorithm}[htbp]
    \caption{The EM algorithm for Synergy Model with Importance Sampling and Sigmoid Smoothing}\label{alg:em-synergy}
    {
    \begin{algorithmic}[1]
        \State \textbf{Input:} Observed data $\{(\bXn,\bm p_n, c_n)\}_{n=1}^N$, tolerance $\epsilon$, Monte Carlo sample size $L$, initial parameters $(\bmu^{(0)},\bsig^{(0)}, \bA^{(0)})$,
        annealing parameters $\lambda_0$, $\lambda_{\min}$, decay rate $\rho\in(0,1)$
        \State \textbf{Initialization:} $error \gets \infty $, $t \gets 0$, $\lambda \gets \lambda_0$
        \While{$error>\epsilon$}
            \For{$n=1,\dots,N$} \Comment{\textbf{E-step:} generate samples}
                \State Construct ${K_n^{c_n}(\bA)}$ using \eqref{eq:ic-polytope-synergy}
                \State Choose the proposal {$\Nscr(\bmu'_n,\bsig'_n)$}  
                \For{$l=1, \dots, L$}
                     \State Repeat $\bm v \sim \Nscr(\bmu'_n,\bsig'_n)$ until $\bm v\in {K_n^{c_n}(\bA^{(t)})}$
                    \State $\bm v_{n}^{(l)}\gets \bm v$
                \EndFor
            \EndFor
            \State Compute normalized weights $\tilde w_{nl}$ for $n=1,\dots,N$ and $l=1,\dots,L$ according to \eqref{eq:IS_sigmoid_weights}.
            \State \textbf{M-step:} Update parameters with weights using \eqref{eq:muSigA_sigmoid_IS}
            \State 
            $
                error \gets 
                \|\bmu^{(t+1)}-\bmu^{(t)}\|_1+
                \|\bsig^{(t+1)}-\bsig^{(t)}\|_1+
                \|\bA^{(t+1)}-\bA^{(t)}\|_1
                ,\quad
                t \gets t+1
            $
            \State Annealing update:
            $
            \lambda\gets \max(\lambda_{\min}, \lambda_0 e^{-\rho t})
            $
        \EndWhile
        \State \textbf{Return:} $\bmu^{(t)}, \bsig^{(t)},\bA^{(t)}$
    \end{algorithmic}
    }
\end{algorithm}

\subsection{Censored Demand}\label{app:censored-demand}

This appendix provides the detailed derivation of the EM algorithm for the censored demand extension in Section~\ref{sec:censored-demand}.
We adopt the same notation: $\bm D = \{N_1,\dots,N_J\}$ denotes the observed data, $\bm Z = \{N',\bm v_1,\dots,\bm v_{N'}\}$ denotes the missing data, and $\bm\theta=(\bmu,\bsig)$ denotes the parameters to be estimated.

\paragraph{E-step}
We first express the complete-data log-likelihood as
\begin{equation*}
    \ell(\bm\theta;\bm D, \bm Z) = \sum_{n=1}^{N'} \log f(\bm v_n\mid \bm\mu,\bm\Sigma).
\end{equation*}
We then take the expectation of $\ell(\bm\theta;\bm D, \bm Z)$ conditional on $\bm D$ under the current estimate $\bm\theta^{(t)}$:
\begin{align}\label{eq:e-step-censored}
    Q(\bm \theta\mid \bm \theta^{(t)})
    &\triangleq \E\left[\ell(\bm\theta;\bm D,\bm Z)\,\Big|\,\bm D,\bm \mu^{(t)},\bm \Sigma^{(t)}\right] \notag\\
    &= \E\left[\E\left[\sum_{n=1}^{N'} \log f(\bm v_n\mid \bm \mu,\bm \Sigma)\,\Big|\,\bm D,\bm \theta^{(t)},N'\right]\,\Big|\,\bm D,\bm \mu^{(t)},\bm \Sigma^{(t)}\right] \notag\\
    &= \E\left[\sum_{n=1}^{N'} \int_{\bm v\in R_n^{c_n}} \frac{f(\bm v\mid \bm\mu^{(t)},\bm\Sigma^{(t)})}{\int_{\bm \xi\in R_n^{c_n}} f(\bm\xi\mid \bm\mu^{(t)},\bm\Sigma^{(t)})\,\mathrm{d}\bm\xi} \log f(\bm v\mid \bm\mu,\bm\Sigma)\,\mathrm{d}\bm v\,\Big|\,\bm D,\bm\mu^{(t)},\bm\Sigma^{(t)}\right].
\end{align}
In the second equality, we apply the tower property by first taking expectation of $\bm v_n$ conditional on $N'$, which mirrors the derivation in Section~\ref{sec: The_EM_Algorithm} for the base model.
The outer expectation in \eqref{eq:e-step-censored} is taken with respect to the distribution of $N'\mid \bm D,\bm\theta^{(t)}$, which we derive next.

Under $\bm\theta^{(t)}$, a customer is censored with probability $\mbp(R^0\mid \bm\mu^{(t)},\bm\Sigma^{(t)})$, the probability mass of the IC polyhedron $R^0$.
Therefore, $N'\mid \bm D,\bm\theta^{(t)}$ admits the following probability model:
after $N'$ i.i.d.\ Bernoulli trials with success probability $1-\mbp(R^0\mid \bm\mu^{(t)},\bm\Sigma^{(t)})$ (an uncensored customer counts as a success), exactly $N$ successes are observed.
By Bayes' rule, for $n\ge N$,
\begin{equation*}
    \mbp\!\left(N'=n\,\big|\,\bm D,\bm\theta^{(t)}\right)
    =\mbp\!\left(N'=n\,\big|\,N,\bm\theta^{(t)}\right)
    = \frac{\mbp(N\mid N'=n,\bm\theta^{(t)})\,\mbp(N'=n)}{\sum_{\tilde n=N}^{\infty}\mbp(N\mid N'=\tilde n,\bm\theta^{(t)})\,\mbp(N'=\tilde n)}.
\end{equation*}

\Copy{rev:improper_prior}{
{
This probability is unspecified unless we impose an assumption on the distribution of $N'$. 
Since $N'$ represents the total number of customers, while $N$ is the observed number of purchasers, we must have $N' \in \{N, N+1, N+2, \ldots\}$.
To reflect the absence of any prior preference over the total number of customers, we assign equal mass to all feasible values of $N'$ in this domain, i.e.,
$
\mbp(N'=n_1) = \mbp(N'=n_2), \forall n_1, n_2  \in \{N, N+1, N+2, \ldots\}.
$
Note that this assignment does not define a proper probability distribution, since the total mass over this infinite set does not sum to one. Instead, it serves as a minimal and non-informative assumption that does not favor any particular value of $N'$.
Importantly, even under this extremely weak and non-normalized assumption, the resulting posterior distribution remains well-defined.
}
}
Therefore, the above term can be expressed as
\begin{align}
     &\frac{{n \choose N} \mbp\left(R^0| \bm \mu^{(t)},\bm \Sigma^{(t)}\right)^{n-N}\left(1-\mbp\left(R^0| \bm \mu^{(t)},\bm \Sigma^{(t)}\right)\right)^N}{\sum_{\tilde{n}=N}^{+\infty}{\tilde{n} \choose N} \mbp\left(R^0| \bm \mu^{(t)},\bm \Sigma^{(t)}\right)^{\tilde{n}-N}(1-\mbp(R^0| \bm \mu^{(t)},\bm \Sigma^{(t)}))^N}\notag \\
&\quad\quad\quad\quad\quad\quad\quad\quad\quad\quad\quad\quad\quad\quad\quad\quad= {n \choose N} \mbp\left(R^0| \bm \mu^{(t)},\bm \Sigma^{(t)}\right)^{n-N}\left(1-\mbp\left(R^0| \bm \mu^{(t)},\bm \Sigma^{(t)}\right)\right)^{N+1}.\label{eq:neg-binomial}
\end{align}
Note that it has the same distribution as the negative binomial distribution with success rate $1-\mbp\left(R^0| \bm \mu^{(t)},\bm \Sigma^{(t)}\right)$ and $N+1$ successes.
Recall that the negative binomial distribution with $N+1$ successes describes the probability mass function of the number of trials in independent Bernoulli trials until the $(N+1)$th success is observed.
The use of the improper prior equates the total number of customers to that right before the $(N+1)$th customer who makes a purchase.

With \eqref{eq:neg-binomial}, the expectation in \eqref{eq:e-step-censored} is approximated by Monte Carlo simulation:
\begin{enumerate}
    \item Consider $L$ instances. In instance $l$, generate $N^{\prime(l)}$ from the negative binomial distribution \eqref{eq:neg-binomial}.
    \item Generate $N_j$ samples from $\Nscr(\bm\mu^{(t)},\bm\Sigma^{(t)})$ conditional on $\bm v\in \mathcal P_\rho$ for $j=1,\dots,J$, denoted $\{\bm v^{(l)}_{j,s}\}_{s=1}^{N_j}$, and $N^{\prime(l)}-N$ samples from $\Nscr(\bm\mu^{(t)},\bm\Sigma^{(t)})$ conditional on $\bm v\in R^0$, denoted $\{\bm v^{(l)}_{0,s}\}_{s=1}^{N^{\prime(l)}-N}$.
    \item Approximate \eqref{eq:e-step-censored} by
    \begin{equation}\label{eq:approx-Q-censor}
        \hat Q(\bm\theta\mid \bm\theta^{(t)})
        = \frac{1}{L}\sum_{l=1}^L \left(\sum_{j=1}^J \sum_{s=1}^{N_j}\log f(\bm v_{j,s}^{(l)}\mid \bm\mu,\bm\Sigma) + \sum_{s=1}^{N^{\prime(l)}-N}\log f(\bm v_{0,s}^{(l)}\mid \bm\mu,\bm\Sigma)\right).
    \end{equation}
\end{enumerate}
The information of $\bm\theta^{(t)}$ has been fully absorbed by the simulated $\{N'^{(l)}\}$ and $\{\bm v_{j,s}^{(l)}\}$, so \eqref{eq:approx-Q-censor} is a function of $(\bm\mu,\bm\Sigma)$ to be optimized in the M-step.

\paragraph{M-step}
Given $\hat Q(\bm\theta\mid \bm\theta^{(t)})$ in \eqref{eq:approx-Q-censor}, the M-step parallels the M-step in the base model.
The updates $\bmu^{(t+1)}$ and $\bsig^{(t+1)}$ are the sample mean and sample covariance of the simulated valuations:
\begin{align*}
    \bmu^{(t+1)} 
    &= \frac{1}{\sum_{l=1}^L N'^{(l)}}\sum_{l=1}^L \left(\sum_{j=1}^J \sum_{s=1}^{N_j}\bm v_{j,s}^{(l)} + \sum_{s=1}^{N^{\prime(l)}-N}\bm v_{0,s}^{(l)}\right), \\
    \bsig^{(t+1)} 
    &= \frac{1}{\sum_{l=1}^L N'^{(l)}}\sum_{l=1}^L \Bigg(\sum_{j=1}^J \sum_{s=1}^{N_j}(\bm v_{j,s}^{(l)}-\bmu^{(t+1)})(\bm v_{j,s}^{(l)}-\bmu^{(t+1)})^\top \\
    &\qquad\qquad\qquad\qquad\qquad + \sum_{s=1}^{N^{\prime(l)}-N}(\bm v_{0,s}^{(l)}-\bmu^{(t+1)})(\bm v_{0,s}^{(l)}-\bmu^{(t+1)})^\top\Bigg).
\end{align*}
With these E- and M-steps, Algorithm~\ref{alg:em-basic} can be adapted directly to censored-demand data.

\section{Gaussian Mixture Models} \label{app:GMM}
\subsection{Model}
In this section, we generalize the base model in Section \ref{sec:base-model} by allowing each customer's valuation vector $\bm v$ to be drawn independently from an $I$-dimensional Gaussian mixture model with $K$ components:
\begin{equation*}
    \bm v\sim \sum_{k=1}^K \phi_k \Nscr(\bm \mu_k,\bm \Sigma_k),
\end{equation*}
where $\phi_k \geq 0$ and $\sum_{k=1}^{K} \phi_k =1$ are the mixture proportions,
$\Nscr(\bm \mu_k,\bm \Sigma_k)$ is the density of a multivariate Gaussian distribution with mean $\bm \mu_k$ and covariance matrix $\bm \Sigma_k$, and $\bm \theta = (\phi_1, \cdots, \phi_K, \bm\mu_1, \cdots, \bm\mu_K, \bm \Sigma_1, \cdots, \bm \Sigma_K )$ denotes the vector of parameters that we need to estimate.
Such a Gaussian mixture model allows for clustering of consumer valuations and can flexibly approximate a wide range of valuation distributions.

We use $\bm \pi^n \in \{0,1\}^K$ to denote the component association of customer $n$:
$\pi_k^n = 1$ if $\bm v_n$ is generated from $\mathcal{N}(\bm \mu_k, \bm\Sigma_k)$ and zero otherwise.
Note that in this model, for each transaction there are two types of latent quantities: (i) the unobserved valuations, and (ii) the mixture component associated with that valuation.
Therefore, the missing data are $\bm Z \triangleq \{ \bm \pi^1, \dots, \bm \pi^N, \bm v_1, \dots, \bm v_{N} \}$.
We express the complete-data log-likelihood function as
{
\begin{equation*}
     \ell(\bm \theta; \bm D, \bm Z ) =  \sum_{n=1}^N \sum_{k=1}^K \bm \pi_k^n [\log (\bm \phi_k) +\log f(\bm v_n |\bm \mu_k,\bm \Sigma_k)],
\end{equation*}
}
where $f(\cdot)$ denotes the PDF of the multivariate normal distribution.

    Compared with the base model in Section~\ref{sec:base-model}, the conditional expectation of the complete-data log-likelihood now involves an outer expectation over the component memberships $\bm \pi^n\mid \bm D, \bm\theta^{(t)}$.
    Given the choice $c_n$ of customer $n$, the probability that customer $n$ is associated with component $k$ is, by Bayes' rule, proportional to $\bm\phi_k^{(t)}\,\mbp(R_n^{c_n}\mid \bm\mu_k^{(t)},\bm\Sigma_k^{(t)},\pi_k^n=1)$, where $\mbp(R_n^{c_n}\mid \bm\mu_k^{(t)},\bm\Sigma_k^{(t)},\pi_k^n=1)$ is the mass that component $k$ assigns to the IC polyhedron $R_n^{c_n}$.
    Intuitively, components that place more probability mass on the IC polyhedron consistent with the observed choice $c_n$ are more likely to have generated customer $n$.

    Once the posterior of $\bm \pi^n$ is identified, the E-step is approximated by Monte Carlo simulation as in Section~\ref{sec: The_EM_Algorithm}: in each instance, we first estimate the membership posterior $\hat\pi_k^n$ by drawing samples from each component $\Nscr(\bm\mu_k^{(t)},\bm\Sigma_k^{(t)})$ and compute the fraction falling in $R_n^{c_n}$, then draw customer valuations from truncated Gaussians on the corresponding IC polyhedra under each component.
    The M-step can then be adapted using a $\hat Q$ function computed from the simulated samples above, and the parameters $(\bm\phi,\bm\mu,\bm\Sigma)$ are updated by maximizing this $\hat Q$, which yields a weighted version of the standard MLE for Gaussian mixtures.
    In this way, Algorithm~\ref{alg:em-basic} can be adapted to GMM with only minor modifications.
    The detailed derivation of the E-step, the closed form for the membership posterior $\hat\pi_k^n$, the Monte Carlo procedure, and the resulting M-step update formulas are provided in Appendix~\ref{app:GMM}.

\subsection{EM Algorithm}

\paragraph{E-step.}
We first express the complete-data log-likelihood as
\begin{equation*}
    \ell(\bm\theta;\bm D,\bm Z) = \sum_{n=1}^N \sum_{k=1}^K \pi_k^n\big[\log\phi_k + \log f(\bm v_n\mid \bm\mu_k,\bm\Sigma_k)\big],
\end{equation*}
where $f(\cdot)$ is the PDF of the multivariate normal distribution.
We then take the expectation of $\ell(\bm\theta;\bm D, \bm Z)$ conditional on $\bm D$ under the current estimate $\bm\theta^{(t)}$:
\begin{align}\label{eq:GMM-Q}
    Q(\bm\theta\mid \bm\theta^{(t)})
    &\triangleq \E_{\bm Z\mid \bm D,\bm\theta^{(t)}}\!\left[\ell(\bm\theta;\bm D,\bm Z)\,\Big|\,\bm D,\bm\mu^{(t)},\bm\Sigma^{(t)},\bm\phi^{(t)}\right] \notag\\
    &= \E_{\bm\pi\mid \bm D,\bm\theta^{(t)}}\!\left[\E_{\bm v\mid \bm D,\bm\theta^{(t)},\bm\pi}\!\left[\sum_{n=1}^N\sum_{k=1}^K \pi_k^n\log\phi_k + \pi_k^n\log f(\bm v_n\mid \bm\mu_k,\bm\Sigma_k)\,\Big|\,\bm D,\bm\theta^{(t)}\right]\right] \notag\\
    &= \sum_{n=1}^N \sum_{k=1}^K \mbp(\pi_k^n=1\mid \bm D,\bm\theta^{(t)})\,\E\!\left[\log\phi_k + \log f(\bm v_n\mid \bm\mu_k,\bm\Sigma_k)\,\Big|\,\bm D,\bm\theta^{(t)},\pi_k^n=1\right],
\end{align}
where the inner conditional expectation is
\begin{equation*}
    \E\!\left[\log f(\bm v_n\mid \bm\mu_k,\bm\Sigma_k)\,\Big|\,\bm D,\bm\theta^{(t)},\pi_k^n=1\right]
    = \int_{\bm v\in R_n^{c_n}}\frac{f(\bm v\mid \bm\mu_k^{(t)},\bm\Sigma_k^{(t)})}{\int_{\bm\xi\in R_n^{c_n}} f(\bm\xi\mid \bm\mu_k^{(t)},\bm\Sigma_k^{(t)})\,\mathrm{d}\bm\xi}\log f(\bm v\mid \bm\mu_k,\bm\Sigma_k)\,\mathrm{d}\bm v.
\end{equation*}

To evaluate \eqref{eq:GMM-Q}, we first derive the posterior membership probability. Given the choice $c_n$, by Bayes' rule,
\begin{equation}\label{eq:posterior}
    \hat{\pi}_k^n \triangleq \mbp(\pi_k^n=1\mid \bm D,\bm\theta^{(t)})
    = \frac{\bm\phi_k^{(t)}\,\mbp(R_n^{c_n}\mid \bm\mu_k^{(t)},\bm\Sigma_k^{(t)},\pi_k^n=1)}{\sum_{k'=1}^K \bm\phi_{k'}^{(t)}\,\mbp(R_n^{c_n}\mid \bm\mu_{k'}^{(t)},\bm\Sigma_{k'}^{(t)},\pi_{k'}^n=1)}.
\end{equation}
The polyhedral mass $\mbp(R_n^{c_n}\mid \bm\mu_k^{(t)},\bm\Sigma_k^{(t)},\pi_k^n=1)$ in \eqref{eq:posterior} can be estimated by Monte Carlo simulation: generate $L$ samples from $\Nscr(\bm\mu_k^{(t)},\bm\Sigma_k^{(t)})$ and compute the fraction falling in $R_n^{c_n}$.

The expectation in \eqref{eq:GMM-Q} is then approximated by the following Monte Carlo procedure:
\begin{enumerate}
    \item For each component $k=1,\dots,K$, generate $L$ samples $\bm v_k^{(l)}$, $l=1,\dots,L$, from $\Nscr(\bm\mu_k^{(t)},\bm\Sigma_k^{(t)})$ and use them to estimate the posterior membership probabilities $\hat\pi_k^n$ via \eqref{eq:posterior}:
    \begin{equation*}
        \hat\mbp(R_n^{c_n}\mid \bm\mu_k^{(t)},\bm\Sigma_k^{(t)}) = \frac{1}{L}\sum_{l=1}^L \mathbb{I}\{\bm v_k^{(l)}\in R_n^{c_n}\}.
    \end{equation*}
    \item For all $n=1,\dots,N$ and $k=1,\dots,K$, generate $L$ samples $\bm v_{nk}^{(l)}$, $l=1,\dots,L$, from $\Nscr(\bm\mu_k^{(t)},\bm\Sigma_k^{(t)})$ conditional on $\bm v_{nk}^{(l)}\in R_n^{c_n}$, using Algorithm~\ref{alg:acc-rej}.
    \item Approximate \eqref{eq:GMM-Q} by
    \begin{equation}\label{eq:Estep-gmm}
        \hat Q(\bm\theta\mid \bm\theta^{(t)})
        = \sum_{n=1}^N \sum_{k=1}^K \hat\pi_k^n \log\phi_k + \frac{1}{L}\sum_{l=1}^L\sum_{n=1}^N\sum_{k=1}^K \hat\pi_k^n \log f(\bm v_{nk}^{(l)}\mid \bm\mu_k,\bm\Sigma_k).
    \end{equation}
\end{enumerate}

\paragraph{M-step.}
Given $\hat Q(\bm\theta\mid \bm\theta^{(t)})$ in \eqref{eq:Estep-gmm}, we maximize over $\bm\theta=(\bm\phi,\bm\mu,\bm\Sigma)$.
For the mixture proportions, applying the constraint $\sum_k\phi_k=1$ to the first term in \eqref{eq:Estep-gmm} yields
\begin{equation*}
    \phi_k^{(t+1)} = \frac{1}{N}\sum_{n=1}^N \hat\pi_k^n.
\end{equation*}
For the component means and covariances, the second term in \eqref{eq:Estep-gmm} is a weighted Gaussian log-likelihood with weights $\hat\pi_k^n$, so the updates are weighted versions of the standard MLE:
\begin{align*}
    \bm\mu_k^{(t+1)} &= \frac{\sum_{l=1}^L\sum_{n=1}^N \hat\pi_k^n\,\bm v_{nk}^{(l)}}{L\sum_{n=1}^N \hat\pi_k^n}, \\
    \bm\Sigma_k^{(t+1)} &= \frac{\sum_{l=1}^L\sum_{n=1}^N \hat\pi_k^n\,(\bm v_{nk}^{(l)}-\bm\mu_k^{(t+1)})(\bm v_{nk}^{(l)}-\bm\mu_k^{(t+1)})^\top}{L\sum_{n=1}^N \hat\pi_k^n}.
\end{align*}
With these E-steps and M-steps, Algorithm~\ref{alg:em-basic} can be adapted directly to the Gaussian mixture setting.

\subsection{Numerical Experiments}\label{subsec:numerical_gmm}
In this experiment, we study the convergence of the EM algorithm for the Gaussian mixture model with two components and compare its performance with that of the base EM algorithm.
We generate synthetic datasets using a procedure similar to that in the base-model experiments.
Specifically, each consumer's product valuation vector is drawn from a two-component Gaussian mixture distribution.
The mixture weight $\phi$ is randomly generated, while the remaining experimental settings follow the base-model experiments.
To systematically evaluate the model, we fix the number of products at $I=2$ and the transaction size at $N=2500$. 
We generate five datasets with different parameters $(\bmu,\bsig)$ and estimate each dataset using both the GMM and the base model.


 \begin{figure}[ht]               
      \begin{center}
      \begin{tikzpicture}                      
      \begin{axis}[                           
          width=0.6\linewidth, height=5.5cm,            
          xlabel=\# iterations,                   
          ylabel=Average log-likelihood,
          xmin=1, xmax=200,
          ymin=-0.90, ymax=-0.55,
          xtick={0,50,100,150,200},
          legend pos=south east,
          legend style={font=\small},
          grid=major, grid style={dashed,gray!30},
      ]
      \addplot[mark=none, solid, line width=1.5pt, color=red]
          table[x=t, y=gmm_em, col sep=comma]%
          {avg_llh_bymethod_GMM8.csv};
      \addplot[mark=none, densely dotted, line width=1.5pt, color=green!70!black]
          table[x=t, y=base_em, col sep=comma]%
          {avg_llh_bymethod_GMM8.csv};
      \addplot[domain=10:200, color=black, dash dot, line width=1.2pt]{-0.5914};
      \legend{GMM EM, Base EM, Exact}
      \end{axis}
      \end{tikzpicture}
      \end{center}
      \caption{Convergence of the average log-likelihood for the GMM and base EM algorithms with $N=2500$ transactions and $I=2$ products.\label{fig:gmm_llh}}
  \end{figure}

\begin{table}[htbp]
    \centering
    \caption{Comparison between GMM and base EM on RMSE and log-likelihood for different parameters.}\label{tab:gmm}
    \begin{tabular}{c|cc|cc}
    \hline
    & \multicolumn{2}{c|}{RMSE} & \multicolumn{2}{c}{Log-likelihood Score} \\
    \text{Dataset} & GMM & EM & GMM & EM \\
    \hline
    1 & 0.0110 & 0.0167 & 100.16\% & 99.88\% \\
    2 & 0.0096 & 0.0272 & 100.21\% & 99.43\% \\
    3 & 0.0106 & 0.0444 & 100.18\% & 96.58\% \\
    4 & 0.0221 & 0.0816 & 100.31\% & 93.94\% \\
    5 & 0.0072 & 0.0780 & 100.12\% & 97.08\% \\
    \hline
    \end{tabular}
    
    \vspace{0.3cm}
    \parbox{0.95\linewidth}{
    \footnotesize
    Notes:
    \begin{enumerate}[(1)]
        \item Log-likelihood Score is defined as
    $1 - (\mathcal L_{\text{model}} - \mathcal L_{\text{exact}})/\mathcal L_{\text{exact}}$,
    so a value closer to $100\%$ indicates that the fitted model is closer to the oracle test log-likelihood.
    \end{enumerate}
    }
\end{table}

As shown in Figure~\ref{fig:gmm_llh}, the GMM model provides a substantially better fit to the observed data compared to the single-component baseline model, which fails to capture the underlying heterogeneity.
Table~\ref{tab:gmm} shows that the GMM consistently achieves lower RMSE and higher test log-likelihood scores than the single-component base EM model.
This improvement is expected because the data are generated from a two-component mixture, whereas the base model imposes a single Gaussian distribution.
The results highlight the value of the GMM extension when the underlying valuation distribution contains latent customer heterogeneity.
    

{
}

\section{Hierarchical Bayesian Model} \label{HBM}
{
}

In this section, we introduce the Metropolis-Hastings algorithm for estimating the posterior distribution. 
We assume a normal proposal distribution and uniform prior distributions for both $\bmu$ and a covariance factor $\bm S$, where $\bsig=\bm S\bm S^\top$.

\begin{algorithm} \label{algo:MH}
\caption{Metropolis--Hastings}
\begin{algorithmic}[1]
       \State \textbf{Input:} 
\State \quad $\{R_n^{c_n}\}_{n=1}^N$: Observed transaction data (bundle choices)
\State \quad $\bmu^{(0)}, \bm S^{(0)}$: Initial parameters
\State \quad $\overline{t}$: Number of MCMC iterations
\State \quad $\pi(\bmu), \pi(\bm S)$: Prior distributions
\State \quad $\delta_\mu, \delta_S$: Proposal standard deviations

\State \textbf{Initialize:} $t \gets 1$
\State \quad $\bsig^{(0)} \gets \bm S^{(0)}(\bm S^{(0)})^\top$

\While{$t \leq \overline{t}$}

    \State \textbf{1. Generate proposal parameters}
    \State $\bm\epsilon_{\bmu} \sim \Nscr\left(\bm 0, \delta^2_{\bmu} I\right)$, $ \bm\epsilon_{\bm S} \sim \Nscr\left(\bm 0, \delta^2_{S} I\right)$
    \State $\bmu^* = \bm\mu^{(t-1)} + \bm\epsilon_{\bmu}$
    \State $\bm S^* = \bm S^{(t-1)} + \bm\epsilon_{\bm S}$
    \State $\bsig^* = \bm S^* (\bm S^*)^\top$
    
    \State \textbf{2. Compute acceptance ratio}
    \State $\mathcal L(\bmu, \bsig) = \prod_{n=1}^N \mbp(R_n^{c_n} \mid \bmu, \bsig)$
    \State $\alpha = \min \left\{
    1, \frac
    { \mathcal L(\bmu^*,\bsig^*) \pi(\bmu^*)\pi(\bm S^*) }
    { \mathcal L(\bmu^{(t-1)}, \bsig^{(t-1)}) \pi(\bmu^{(t-1)}) \pi(\bm S^{(t-1)})}
    \right\}$

    \State \textbf{3. Accept or reject}
    \State $\beta \sim \mathcal{U}(0,1)$
    \If{$\beta \le \alpha$}
        \State $\bmu^{(t)} \gets \bmu^*, \quad \bm S^{(t)} \gets \bm S^*, \quad \bsig^{(t)} \gets \bsig^*$
    \Else
        \State $\bmu^{(t)} \gets \bmu^{(t-1)}, \quad \bm S^{(t)} \gets \bm S^{(t-1)}, \quad \bsig^{(t)} \gets \bsig^{(t-1)}$
    \EndIf

    \State $t \gets t + 1$

\EndWhile

\State \Return $\bmu^{(t)}, \bsig^{(t)}$
    \end{algorithmic} 
\end{algorithm}

{
}

\section{Multinomial Logit Model for Bundle Choice}\label{app:mnl}
This appendix describes the MNL model used in the numerical experiments.  
The purpose of this benchmark is to evaluate how a classical discrete-choice model performs when the true data-generating process follows the proposed Gaussian bundle-choice model.
The resulting estimator therefore serves as a misspecified benchmark against the proposed Gaussian-based estimator.

Suppose there are $I$ products. The true latent product valuation vector is generated from a multivariate Gaussian distribution: $\bm v_n\sim \Nscr(\bmu,\bsig)$, where $\bsig$ is a diagonal matrix. 
Although the data are generated from a Gaussian model, the MNL estimator assumes the standard multinomial logit structure.
Specifically, the MNL model assumes the utility specification
\begin{align}
U_{n0}^{\text{MNL}} &= \varepsilon_{n0}, \\
U_{nj}^{\text{MNL}} &= \bm x_{nj}^\top \bmu - p_{nj} + \varepsilon_{nj},
\qquad j=1,\dots,J_n,
\end{align}
where $\bmu\in\mathbb{R}^I$ is the vector of product-level mean valuations to be estimated, and
$\varepsilon_{nj} \stackrel{\text{i.i.d.}}{\sim} \mathrm{Gumbel}(0,1).$
Therefore, under the MNL model, bundle utilities are simply the sums of constituent product mean valuations plus an independent Gumbel error term.

Under the MNL assumptions, the standard logit formula gives:
\begin{align}
\mbp(c_n = 0) &=
\frac{1}
{1+\sum_{j=1}^{J_n}\exp\!\left(\bm x_{nj}^\top \bmu - p_{nj}\right)},\\
\mbp(c_n = j) &=
\frac{\exp\!\left(\bm x_{nj}^\top \boldsymbol{\mu} - p_{nj}\right)}
{1+\sum_{j'=1}^{J_n}\exp\!\left(\bm x_{nj'}^\top \bmu - p_{nj'}\right)}.
\end{align}

The MNL benchmark estimates only the product-level mean valuations $\bmu$ by maximizing the log-likelihood implied by the logit probabilities:
\[
\widehat{\bmu}_{\mathrm{MNL}} = \arg\max_{\bmu}
\sum_{n=1}^N \log \mbp(c_n \mid \bmu).
\]
Optimization is performed using the L-BFGS-B algorithm.

%% file: EC_Bundle.tex
		\newpage
\setcounter{page}{1}
\renewcommand{\thepage}{EC.\arabic{page}}

	   \begin{center}
	  {\bf\large
	 Online Supplement to \\``Learning Customer Preferences from Bundle Sales Data"}
	\end{center}

	\renewcommand\thesection{\Alph{section}}

	\setcounter{equation}{0}
	\setcounter{section}{0}
	\setcounter{lem}{0}
	\setcounter{defn}{0}
	\setcounter{thm}{0}
	\setcounter{prop}{0}

	\numberwithin{lem}{section}
	\numberwithin{defn}{section}
	\numberwithin{equation}{section}
	\numberwithin{thm}{section}
	\numberwithin{prop}{section}

\section{Proofs} \label{sec:A}

\subsection{The Derivation of the M-step \label{sec:M-step}}
    \begin{align}
    Q\left(\bm \theta|\bm \theta^{(t)}\right)&\triangleq \E\left[l(\bm\theta;\bm D,\bm Z)\bigg|\bm D,\bm \mu^{(t)},\bm \Sigma^{(t)}\right]\notag\\
                                  &= \sum_{n=1}^{N} \int_{\bm v\in R_n^{c_n}} \frac{f\left(\bm v|\bm \mu^{(t)},\bm \Sigma^{(t)}\right)}{\int_{R_n^{c_n}}f\left(\bm \xi|\bm \mu^{(t)},\bm \Sigma^{(t)}\right)\mathrm{d}\bm \xi }  \log f(\bm v|\bm \mu,\bm \Sigma)\mathrm{d}\bm v.
\end{align}
To find a new estimation for $\bth$, the \emph{M-step} in the EM algorithm maximizes $Q\left(\bth|\bth^{(t)}\right)$ and let
$\bth^{(t+1)}=\argmax_{\bth}Q\left(\bth|\bth^{(t)}\right)$. Thus, we first solve \begin{align}
\partial \E\left[l(\bm\theta;\bm D,\bm Z)\mid\bm D,\bm \mu^{(t)},\bm \Sigma^{(t)}\right]/\partial \bmu = 0, 	\label{eq:deriv-mu}
 \end{align}
and then we solve
	\begin{align}
	\partial \E\left[l(\bm\theta;\bm D,\bm Z)\mid\bm D,\bm \mu^{(t)},\bm \Sigma^{(t)}\right]/\partial \bm \Sigma = 0. \label{eq:deriv-sigma}
	\end{align}
Equality \eqref{eq:deriv-mu} equates to
\begin{align}
\frac{\partial \E\left[l(\bm\theta;\bm D,\bm Z)\mid\bm D,\bm \mu^{(t)},\bm \Sigma^{(t)}\right]}{\partial \bm \mu} =& \frac{\partial }{\partial \bm \mu}  \sum_{n=1}^{N} \int_{\bm v\in R_n^{c_n}} \frac{f\left(\bm v|\bm \mu^{(t)},\bm \Sigma^{(t)}\right)}{\int_{R_n^{c_n}}f(\bm D|\bm  \mu^{(t)},\bm \Sigma^{(t)})\mathrm{d}\bm \xi } \log f(\bm v|\bm \mu,\bm \Sigma)\mathrm{d}\bm v \notag \\
=&\sum_{n=1}^{N} \int_{\bm v\in R_n^{c_n}}  \frac{f\left(\bm v|\bm \mu^{(t)},\bm \Sigma^{(t)}\right)}{\int_{R_n^{c_n}}f\left(\bm \xi|\bm \mu^{(t)},\bm \Sigma^{(t)}\right)\mathrm{d}\bm \xi }  \frac{\partial }{\partial \bm \mu} \log f(\bm v|\bm \mu,\bm \Sigma)\mathrm{d}\bm v\notag\\
=&\sum_{n=1}^{N} \int_{\bm v\in R_n^{c_n}}  \frac{f\left(\bm v|\bm \mu^{(t)},\bm \Sigma^{(t)}\right)}{\int_{R_n^{c_n}}f\left(\bm \xi|\bm \mu^{(t)},\bm \Sigma^{(t)}\right)\mathrm{d}\bm \xi }  \frac{1}{f(\bm v|\bm \mu,\bm \Sigma)} \bm \Sigma^{-1}(\bm v - \bm \mu) f(\bm v|\bm \mu,\bm \Sigma) \mathrm{d}\bm v \notag\\
= &\sum_{n=1}^{N} \int_{\bm v\in R_n^{c_n}}  \bm \Sigma^{-1}(\bm v - \bm \mu) \frac{f\left(\bm v|\bm \mu^{(t)},\bm \Sigma^{(t)}\right)}{\int_{R_n^{c_n}}f\left(\bm \xi|\bm \mu^{(t)},\bm \Sigma^{(t)}\right)\mathrm{d}\bm \xi } \mathrm{d}\bm v \notag\\
=&\sum_{n=1}^{N} \int_{\bm v\in R_n^{c_n}}  \bm v \frac{f\left(\bm v|\bm \mu^{(t)},\bm \Sigma^{(t)}\right)}{\int_{R_n^{c_n}}f\left(\bm \xi|\bm \mu^{(t)},\bm \Sigma^{(t)}\right)\mathrm{d}\bm \xi } \mathrm{d}\bm v \notag \\ &- \sum_{n=1}^{N} \int_{\bm v\in R_n^{c_n}} \bm \mu \frac{f\left(\bm v|\bm \mu^{(t)},\bm \Sigma^{(t)}\right)}{\int_{R_n^{c_n}}f\left(\bm \xi|\bm \mu^{(t)},\bm \Sigma^{(t)}\right)\mathrm{d}\bm \xi } \mathrm{d}\bm v \notag\\
=&\sum_{n=1}^{N} \int_{\bm v\in R_n^{c_n}}  \bm v \frac{f\left(\bm v|\bm \mu^{(t)},\bm \Sigma^{(t)}\right)}{\int_{R_n^{c_n}}f\left(\bm \xi|\bm \mu^{(t)},\bm \Sigma^{(t)}\right)\mathrm{d}\bm \xi } \mathrm{d}\bm v  -  N \bm \mu =0. \label{eq:Q-firstorder-mu}
\end{align}
Thus, we have
\begin{align}
 \bm \mu = \frac{\sum_{n=1}^{N} \int_{\bm v\in R_n^{c_n}}  \bm v  \frac{f\left(\bm v|\bm \mu^{(t)},\bm \Sigma^{(t)}\right)}{\int_{R_n^{c_n}}f\left(\bm \xi|\bm \mu^{(t)},\bm \Sigma^{(t)}\right)\mathrm{d}\bm \xi } \mathrm{d}\bm v}{ N} = \frac{\sum_{n=1}^{N} \bm \E[\bm v |\bm \mu^{(t)},\bm \Sigma^{(t)},   R_n^{c_n}]}{N}.
\end{align}
Equality \eqref{eq:deriv-sigma} equates to
\begin{align}
\frac{\partial  \E\left[l(\bm\theta;\bm D,\bm Z)\mid\bm D,\bm \mu^{(t)},\bm \Sigma^{(t)}\right]}{\partial \bm \Sigma} &=\sum_{n=1}^{N}\frac{\partial }{\partial \bm \Sigma}  \int_{\bm v\in R_n^{c_n}}  \frac{f\left(\bm v|\bm \mu^{(t)},\bm \Sigma^{(t)}\right)}{\int_{R_n^{c_n}}f(\bm \xi|\bm  \mu^{(t)},\bm \Sigma^{(t)})\mathrm{d}\bm \xi } \log f(\bm v|\bm \mu,\bm \Sigma) \mathrm{d}\bm v \notag \\
&=  \sum_{n=1}^{N} \int_{\bm v\in R_n^{c_n}} \frac{f\left(\bm v|\bm \mu^{(t)},\bm \Sigma^{(t)}\right)}{\int_{R_n^{c_n}}f(\bm \xi|\bm  \mu^{(t)},\bm \Sigma^{(t)})\mathrm{d}\bm \xi }(-\frac{1}{2} \bm \Sigma^{-1} + \frac{1}{2} \bm \Sigma^{-1} (\bm v- \bm \mu) (\bm v- \bm \mu) ^\top \bm \Sigma^{-1}) \mathrm{d}\bm v \label{eq:Q-firstorder-sigma} \\
&=-\frac{1}{2} \bm \Sigma^{-1} \sum_{n=1}^{N} \int_{\bm v\in R_n^{c_n}}  \frac{f\left(\bm v|\bm \mu^{(t)},\bm \Sigma^{(t)}\right)}{\int_{R_n^{c_n}}f(\bm \xi|\bm  \mu^{(t)},\bm \Sigma^{(t)})\mathrm{d}\bm \xi }  \mathrm{d}\bm v \notag \\
&\quad +\frac{1}{2} \bm \Sigma^{-1}\sum_{n=1}^{N} \int_{\bm v\in R_n^{c_n}}  \frac{f\left(\bm v|\bm \mu^{(t)},\bm \Sigma^{(t)}\right)}{\int_{R_n^{c_n}}f(\bm \xi|\bm  \mu^{(t)},\bm \Sigma^{(t)})\mathrm{d}\bm \xi }    (\bm v- \bm \mu) (\bm v- \bm \mu) ^\top \bm \Sigma^{-1}  \mathrm{d}\bm v = 0. \notag
\end{align}
Thus, we have
\begin{align*}
 \bm \Sigma &= \frac{ \sum_{n=1}^{N} \int_{\bm v\in R_n^{c_n}} (\bm v- \bm \mu) (\bm v- \bm \mu) ^\top  \frac{f\left(\bm v|\bm \mu^{(t)},\bm \Sigma^{(t)}\right)}{\int_{R_n^{c_n}}f(\bm \xi|\bm  \mu^{(t)},\bm \Sigma^{(t)})\mathrm{d}\bm \xi }  \mathrm{d}\bm v }{N}
\end{align*}

\subsection{Identifiability}\label{appendix:identifiability}

\proof{Proof of Proposition \ref{prop:identifiability}:}

    We divide the proof into two parts.

    \paragraph{Necessity.}
    We first show that having at least two distinct price points for each product is necessary.
    Suppose, to the contrary, that there exists a product $i$ that is observed only at a single price $p_i$ under all separate-selling menus. We construct two distinct parameter vectors that generate exactly the same region probabilities.

    Fix the parameters of all products other than $i$, and assume that product $i$ is independent of the remaining products, that is,
    \[
        \Sigma_{ij}=0 \qquad \text{for all } j \neq i.
    \]
    Let $\mbp_i\triangleq\mbp(_i \le p_i)$.
    Since only one threshold $p_i$ is observed for product $i$, the observable information about its marginal distribution is reduced to the single equation
    \[
        \mbp_i = \Phi\!\left(\frac{p_i-\mu_i}{\sqrt{\Sigma_{ii}}}\right).
    \]
    For any scalar $\tau>0$, define
    \[
        \mu_i(\tau) \triangleq p_i - \tau \Phi^{-1}(\mbp_i),
        \qquad
        \Sigma_{ii}(\tau)\triangleq \tau^2.
    \]
    Then
    \[
         \Phi\!\left(\frac{p_i-\mu_i(\tau)}{\sqrt{\Sigma_{ii}(\tau)}}\right) = \mbp_i
        \qquad \text{for every } \tau>0.
    \]
    Choose two distinct values $\tau_1 \neq \tau_2$, and let $\bm\theta^{(1)}$ and $\bm\theta^{(2)}$ be the corresponding full parameter vectors obtained by replacing only $(\mu_i,\Sigma_{ii})$ with $(\mu_i(\tau_1),\Sigma_{ii}(\tau_1))$ and $(\mu_i(\tau_2),\Sigma_{ii}(\tau_2))$, respectively, while keeping the remaining coordinates of $\bmu$ and $\bsig$ fixed.
    
    Because product $i$ is independent of the remaining products and appears only through the single threshold $p_i$, the probability of every region induced by the separate-selling menus factors into a term depending on $\mbp_i$ and a term depending only on the remaining products. Therefore $\bm\theta^{(1)}$ and $\bm\theta^{(2)}$ induce exactly the same collection of region probabilities, although $\bm\theta^{(1)} \neq \bm\theta^{(2)}$. This contradicts identifiability.

    Hence, at least two distinct price points for each product are necessary.

    \paragraph{Sufficiency.}
    We now show that the menu design in Proposition~\ref{prop:identifiability} is sufficient for identification.

    \paragraph{Step 1: identification of the diagonal terms.}
    Fix any product $i\in\{1,\dots,I\}$. Under the regular menu $\bm p^{(0)}$ and the corresponding single-sale menu $\bm p^{(i)}$, product $i$ is offered at two distinct price levels $p_i < p_i'$. In the separate-selling context, the observable data allow us to estimate the probability mass of the random vector $V_i \sim \Nscr(\mu_i,\Sigma_{ii})$ over the intervals induced by these two prices. Let
    \[
    \mathcal P_{i,1} \triangleq (-\infty,p_i], \qquad
    \mathcal P_{i,2} \triangleq(p_i,p_i'], \qquad
    \mathcal P_{i,3} \triangleq (p_i',+\infty).
    \]
    By observing the fraction of customers who do not purchase product $i$ at prices $p_i$ and $p_i'$, we can consistently estimate
    \[
        \mbp_i \triangleq \mbp\left(V_i\in \mathcal P_{i,1}  \right),
        \qquad
        \mbp_i' \triangleq  \mbp\left(V_i\in \left( \mathcal P_{i,1}\cup \mathcal P_{i,2} \right)  \right).
    \]
    Since $p_i<p_i'$, we have $\mathcal P_{i,1}\subsetneq \left( \mathcal P_{i,1}\cup \mathcal P_{i,2} \right)$, and because the Gaussian density is strictly positive on $\R$, it follows that
    \[
        \mbp_i < \mbp_i'.
    \]
    The identifiability of $(\mu_i,\Sigma_{ii})$ is equivalent to showing that the mapping from $(\mu_i,\Sigma_{ii})$ to the observed probabilities $(\mbp_i,\mbp_i')$ is injective. By the Gaussian CDF $\Phi(\cdot)$,
    \begin{equation}
        \mbp_i = \Phi\!\left(\frac{p_i-\mu_i}{\sqrt{\Sigma_{ii}}}\right),
        \qquad
        \mbp_i' = \Phi\!\left(\frac{p_i'-\mu_i}{\sqrt{\Sigma_{ii}}}\right).
    \end{equation}
    Since $\Phi(\cdot)$ is a strictly increasing bijection from $\R$ to $(0,1)$ and $\mbp_i < \mbp_i'$, the values $\Phi^{-1}(\mbp_i)$ and $\Phi^{-1}(\mbp_i')$ are uniquely determined by the observed data. Therefore the system can be rearranged as
    \begin{equation}
        \begin{cases}
            \mu_i + \Phi^{-1}(\mbp_i)\sqrt{\Sigma_{ii}} = p_i,\\
            \mu_i + \Phi^{-1}(\mbp_i')\sqrt{\Sigma_{ii}} = p_i'.
        \end{cases}
    \end{equation}
    Equivalently,
    \begin{equation}
        \begin{bmatrix}
            1 & \Phi^{-1}(\mbp_i)\\
            1 & \Phi^{-1}(\mbp_i')
        \end{bmatrix}
        \begin{bmatrix}
            \mu_i\\
            \sqrt{\Sigma_{ii}}
        \end{bmatrix}
        =
        \begin{bmatrix}
            p_i\\
            p_i'
        \end{bmatrix}.
    \end{equation}
    To establish identifiability, we verify that this linear system has a unique solution. Consider the coefficient matrix
    \[
        \mathcal A_i \triangleq
        \begin{bmatrix}
            1 & \Phi^{-1}(\mbp_i)\\
            1 & \Phi^{-1}(\mbp_i')
        \end{bmatrix}
    \]
    and the augmented matrix
    \[
        \left[\ \mathcal A_i\ |\ \bm b_i\ \right] \triangleq
        \left[
        \begin{array}{cc|c}
            1 & \Phi^{-1}(\mbp_i) & p_i\\
            1 & \Phi^{-1}(\mbp_i') & p_i'
        \end{array}
        \right],
    \]
    where $\bm b_i=(p_i,p_i')^\top$.
    First,
    \[
        \det(\mathcal A_i)=\Phi^{-1}(\mbp_i')-\Phi^{-1}(\mbp_i)\neq 0,
    \]
    because $\mbp_i<\mbp_i'$ and $\Phi^{-1}$ is strictly increasing. Hence
    \[
        \operatorname{rank}(\mathcal A_i)=2.
    \]
    Since the augmented matrix has the same number of rows, its rank cannot exceed $2$, and because it contains $\mathcal A_i$ as a submatrix, its rank is also $2$. Therefore
    \[
        \operatorname{rank}(\mathcal A_i)=\operatorname{rank}([\mathcal A_i\mid \bm b_i])=2.
    \]
    Thus the linear system has a unique solution. Hence $\mu_i$ and $\Sigma_{ii}$ are uniquely identified.

    Since $i$ was arbitrary, all entries of $\bmu$ and all diagonal entries of $\bsig$ are identified.

    \paragraph{Step 2: identification of the off-diagonal terms.}
    Fix any pair $1 \le i < j \le I$. Under the regular menu, the joint no-purchase probability
    \[
        \mbp_{ij} \triangleq \mbp(V_i \le p_i,\ V_j \le p_j)
    \]
    is observed from the data.

    By Step~1, $\mu_i,\mu_j,\Sigma_{ii},\Sigma_{jj}$ are already known. Define
    \[
        c_i \triangleq \frac{p_i-\mu_i}{\sqrt{\Sigma_{ii}}},
        \qquad
        c_j \triangleq \frac{p_j-\mu_j}{\sqrt{\Sigma_{jj}}},
        \qquad
        \rho_{ij} \triangleq \frac{\Sigma_{ij}}{\sqrt{\Sigma_{ii}\Sigma_{jj}}}.
    \]
    Then
    \[
        \mbp_{ij} = \Phi_2(c_i,c_j;\rho_{ij}),
    \]
    where $\Phi_2(\cdot,\cdot;\rho)$ denotes the CDF of the standard bivariate normal distribution with correlation $\rho$.

    For fixed $(c_i,c_j)$, the map
    \[
        \rho \mapsto \Phi_2(c_i,c_j;\rho)
    \]
    is strictly increasing on $(-1,1)$. Indeed, by Plackett's identity,
    \[
        \frac{\partial}{\partial \rho}\Phi_2(c_i,c_j;\rho)=\phi_2(c_i,c_j;\rho)>0,
    \]
    where $\phi_2(\cdot,\cdot;\rho)$ is the corresponding bivariate normal density.
    Therefore $\mbp_{ij}$ uniquely determines $\rho_{ij}$, and hence
    \[
        \Sigma_{ij} = \rho_{ij}\sqrt{\Sigma_{ii}\Sigma_{jj}}
    \]
    is uniquely determined.

    Since the pair $(i,j)$ was arbitrary, every off-diagonal entry $\Sigma_{ij}$ is identified.

    \paragraph{Conclusion.}
    Step~1 identifies all entries of $\bmu$ and all diagonal entries of $\bsig$. Step~2 identifies all off-diagonal entries of $\bsig$. Therefore the full parameter pair $(\bmu,\bsig)$ is uniquely determined by the probabilities of the regions induced by the $I+1$ separate-selling menus. This proves sufficiency.

    Combining the necessity and sufficiency parts, we conclude that under separate selling, having at least two distinct price points for each product is necessary, and the menu design in Proposition~\ref{prop:identifiability} is sufficient, for the identification of $(\bmu,\bsig)$.\halmos
    \endproof

\begin{sfnadded}

\end{sfnadded}

\subsection{Proofs for Section~\ref{sec:EM-conv}: EM Convergence}\label{appendix:EM-conv}
    \subsubsection{Preliminaries}
    Recall the definition of the population-level $Q$-function:
    \begin{equation}\label{eq:pop-Q}
    Q(\bmu'\mid \bmu) =  \sum_{\rho=1}^{P}  
\left(\int_{\bm \xi \in {\mathcal P_\rho}} f(\bm \xi \mid \bmu^*) \, \mathrm d\bm \xi \right)
\int_{\bm v \in {\mathcal P_\rho}} 
\frac{f(\bm v \mid \bmu)}
{\int_{\bm \xi \in {\mathcal P_\rho}} f(\bm \xi \mid \bmu) \, \mathrm d\bm \xi}
\log f(\bm v \mid \bmu') \, \dv.
    \end{equation}
    By definition, $q(M(\bmu)) = Q(M(\bmu)\mid \bmu^*)$. Therefore,
    \begin{align}
    q(M(\bmu)) &= Q(M(\bmu) \mid \bmu^*) \notag \\
&= \sum_{\rho=1}^{P}  \int_{\bm v \in {\mathcal P_\rho}} f(\bm v \mid \bmu^*) \log f(\bm v \mid M(\bmu)) \, \dv \notag \\
&= \mathbb{E}_{\bmu^*}\left[\log f(\bm V \mid M(\bmu))\right]. \label{q-func-simp}
\end{align}
    We rewrite $Q(M(\bmu)\mid \bmu)$ as
    \begin{align}
   Q(M(\bmu)\mid \bmu) 
&= \sum_{\rho=1}^{P}  
\left(\int_{\bm \xi \in {\mathcal P_\rho}} f(\bm \xi \mid \bmu^*) \, \mathrm d\bm \xi \right)
\int_{\bm v \in {\mathcal P_\rho}} 
\frac{f(\bm v \mid \bmu)}
{\int_{\bm \xi \in {\mathcal P_\rho}} f(\bm \xi \mid \bmu) \, \mathrm d\bm \xi}
\log f(\bm v \mid M(\bmu)) \, \dv \notag \\
&= \sum_{\rho=1}^{P}  
\frac{\int_{\bm \xi \in {\mathcal P_\rho}} f(\bm \xi \mid \bmu^*) \,\mathrm d\bm \xi}
{\int_{\bm \xi \in {\mathcal P_\rho}} f(\bm \xi \mid \bmu) \, \mathrm d\bm \xi}
\int_{\bm v \in {\mathcal P_\rho}} f(\bm v \mid \bmu) \log f(\bm v \mid M(\bmu)) \, \dv.
\label{Q-func-simp}
\end{align}
For the Gaussian density, we have
\begin{equation}
\nabla_{\bmu} \log f(\bm v \mid \bmu) 
= (\bsig^*)^{-1} (\bm v - \bmu).
\end{equation}
Using the above identity, we obtain
\begin{align}
\nabla Q(M(\bmu)\mid \bmu)
&=\sum_{\rho=1}^{P}  
\frac{\int_{\bm \xi \in {\mathcal P_\rho}} f(\bm \xi \mid \bmu^*) \, \mathrm d\bm \xi}
{\int_{\bm \xi \in {\mathcal P_\rho}} f(\bm \xi \mid \bmu) \, \mathrm d\bm \xi}
\int_{\bm v \in {\mathcal P_\rho}} f(\bm v \mid \bmu)
(\bsig^*)^{-1} (\bm v - M(\bmu)) \, \dv \notag \\
&= \sum_{\rho=1}^{P}  
\frac{\int_{\bm \xi \in {\mathcal P_\rho}} f(\bm \xi \mid \bmu^*) \, \mathrm d\bm \xi}
{\int_{\bm \xi \in {\mathcal P_\rho}} f(\bm \xi \mid \bmu) \, \mathrm d\bm \xi}
\int_{\bm v \in {\mathcal P_\rho}} f(\bm v \mid \bmu)
(\bsig^*)^{-1} \bm v \, \dv
- (\bsig^*)^{-1} M(\bmu).
\label{deltatheta}
\end{align}
Similarly, for $q(\bmu) = Q(\bmu \mid \bmu^*)$, we have
\begin{align}
\nabla q(\bmu)
= \mathbb{E}_{\bmu^*}\left[(\bsig^*)^{-1}(\bm V - \bmu)\right]
= (\bsig^*)^{-1}(\bmu^* - \bmu).
\label{deltathetastar}
\end{align}

\subsubsection{EM Convergence}
\begin{lem} \label{lem:concave}
The function $q(\bmu) = \mathbb{E}_{\bmu^*}[\log f(\bm V \mid \bmu)]$ is strongly concave. 
In particular, for all $\bmu_1, \bmu_2$, we have
\begin{align*}
q(\bmu_1) - q(\bmu_2) - \langle \nabla q(\bmu_2), \bmu_1 - \bmu_2 \rangle
\le -\frac{1}{2 \lambda_{\max}(\bsig^*)} \|\bmu_1 - \bmu_2\|_2^2.
\end{align*}
\end{lem}
\proof{Proof of Lemma \ref{lem:concave}:}
Recall that
\[
q(\bmu) = \mathbb{E}_{\bmu^*}[\log f(\bm V \mid \bmu)].
\]
For the Gaussian density, we have
\[
\log f(\bm V \mid \bmu)
= -\frac{1}{2} (\bm V - \bmu)^\top (\bsig^*)^{-1} (\bm V - \bmu) + C,
\]
where $C$ is a constant independent of $\bmu$.

Therefore,
\[
q(\bmu)
= -\frac{1}{2} \mathbb{E}_{\bmu^*}\left[(\bm V - \bmu)^\top (\bsig^*)^{-1} (\bm V - \bmu)\right] + C.
\]

Taking gradient and Hessian with respect to $\bmu$, we obtain
\[
\nabla q(\bmu) = (\bsig^*)^{-1}(\bmu^* - \bmu),
\qquad
\nabla^2 q(\bmu) = -(\bsig^*)^{-1}.
\]

Since $(\bsig^*)^{-1}$ is positive definite, we have
\[
\nabla^2 q(\bmu) \preceq -\frac{1}{\lambda_{\max}(\bsig^*)} I.
\]

The result then follows from the standard characterization of strong concavity.
\halmos \endproof

\begin{lem}
	\label{lem:FOScondition}
	Suppose Assumption~\ref{minum-eig-assumption} holds. 
Then there exists a neighborhood $\mathbb{B}_2(r;\bmu^*)$ such that for all $\bmu \in \mathbb{B}_2(r;\bmu^*)$,
\begin{align}
\| \nabla Q(M(\bmu) \mid \bmu) - \nabla Q(M(\bmu) \mid \bmu^*) \|_2 
\le \frac{1-\epsilon/2}{\lambda_{\min}(\bsig^*)} \|\bmu - \bmu^*\|_2.\label{gamma-fos}
\end{align}
\end{lem}

\proof{Proof of Lemma \ref{lem:FOScondition}.} 
Using \eqref{deltatheta}, we can express the gradient difference as
\begin{align}
\nabla Q(M(\bmu)\mid \bmu) - \nabla Q(M(\bmu)\mid \bmu^*)
= (\bsig^*)^{-1} \left( \sum_{\rho=1}^{P} \mathbb{P}_{\bmu^*}(\mathcal P_\rho) \, g_j(\bmu) - \bmu^* \right),
\end{align}
where
\[
g_j(\bmu) := \mathbb{E}_{\bmu}[\bm V \mid \bm V \in \mathcal P_\rho].
\]
A direct calculation shows that
\[
\nabla g_j(\bmu) = \operatorname{Var}_{\bmu}(\bm V \mid \mathcal P_\rho)(\bsig^*)^{-1}.
\]

Applying a first-order Taylor expansion around $\bmu^*$,
\[
g_j(\bmu) = g_j(\bmu^*) + \operatorname{Var}_{\bmu^*}(\bm V \mid \mathcal P_\rho)(\bsig^*)^{-1}(\bmu - \bmu^*) + r_j(\bmu),
\]
where $\|r_j(\bmu)\| = O(\|\bmu - \bmu^*\|^2)$.

Note that $\sum_{\rho=1}^J \mathbb P_{\bmu^*}(\mathcal P_\rho) g_j(\bmu^*) = \mbe_{\bmu^*}[\bm V] = \bmu^*$. Summing the Taylor expansion over $j$ weighted by $\mathbb P_{\bmu^*}(\mathcal P_\rho)$, and using the law of total covariance
\begin{align}
\sum_{\rho=1}^{P} \mathbb{P}_{\bmu^*}(\mathcal P_\rho)\operatorname{Var}_{\bmu^*}(\bm V \mid \mathcal P_\rho)
= \bsig^* - \operatorname{Var}_{\bmu^*}(\mathbb{E}_{\bmu^*}[\bm V \mid R']),
\end{align}
where $R'$ is the categorical region label introduced in Section~\ref{sec:EM-conv}, we obtain
\begin{align}
\nabla Q(M(\bmu)\mid \bmu) - \nabla Q(M(\bmu)\mid \bmu^*)
= (\bsig^*)^{-1} \left( \bsig^* - \operatorname{Var}_{\bmu^*}(\mathbb{E}_{\bmu^*}[\bm V \mid R']) \right)(\bsig^*)^{-1}(\bmu - \bmu^*) + O(\|\bmu - \bmu^*\|^2).
\end{align}

Recall the whitening transformation $\bm V' = (\bsig^*)^{-1/2}(\bm V - \bmu^*)$. The conditional means satisfy $\mbe_{\bmu^*}[\bm V\mid R'] = (\bsig^*)^{1/2}\mbe[\bm V'\mid R']+\bmu^*$, so
\[
\operatorname{Var}_{\bmu^*}(\mathbb{E}_{\bmu^*}[\bm V \mid R'])
= (\bsig^*)^{1/2} \operatorname{Var}(\mathbb{E}[\bm V' \mid R']) (\bsig^*)^{1/2}.
\]
Substituting back yields
\begin{align}
\nabla Q(M(\bmu)\mid \bmu) - \nabla Q(M(\bmu)\mid \bmu^*)
= (\bsig^*)^{-1/2} \left( I - \operatorname{Var}(\mathbb{E}[\bm V' \mid R']) \right) (\bsig^*)^{-1/2}(\bmu - \bmu^*) + O(\|\bmu - \bmu^*\|^2).
\end{align}

By the law of total variance, $\operatorname{Var}(\bm V') = \operatorname{Var}(\mathbb{E}[\bm V'\mid R']) + \mbe[\operatorname{Var}(\bm V'\mid R')]$, and since $\operatorname{Var}(\bm V')=I$ and $\mbe[\operatorname{Var}(\bm V'\mid R')] \succeq 0$, we have $0 \preceq \operatorname{Var}(\mathbb E[\bm V'\mid R']) \preceq I$. Combined with Assumption~\ref{minum-eig-assumption},
\[
0 \le \lambda_{\max}\!\left(I - \operatorname{Var}(\mathbb{E}[\bm V' \mid R'])\right) \le 1 - \epsilon.
\]

By sub-multiplicativity of the spectral norm,
\[
\big\|(\bsig^*)^{-1/2}\big(I - \operatorname{Var}(\mathbb E[\bm V'\mid R'])\big)(\bsig^*)^{-1/2}\big\|_2 
\le (1-\epsilon)\,\big\|(\bsig^*)^{-1/2}\big\|_2^{\,2}
= \frac{1-\epsilon}{\lambda_{\min}(\bsig^*)}.
\]

Choose $r$ small enough so that for all $\bmu \in \mathbb B_2(r;\bmu^*)$, the higher-order term satisfies $\|O(\|\bmu-\bmu^*\|^2)\| \le \frac{\epsilon/2}{\lambda_{\min}(\bsig^*)}\|\bmu-\bmu^*\|$. Then
\[
\|\nabla Q(M(\bmu)\mid \bmu) - \nabla Q(M(\bmu)\mid \bmu^*)\|_2
\le \frac{1-\epsilon/2}{\lambda_{\min}(\bsig^*)} \|\bmu - \bmu^*\|_2.
\]
\halmos\endproof

\proof{Proof of Theorem \ref{thm:contra-map}.} 
The result follows from \cite[Theorem 4]{balakrishnan2017statistical}. 
By Lemma~\ref{lem:concave}, the function $q(\bmu)$ is strongly concave with parameter
\[
\lambda = \frac{1}{\lambda_{\max}(\bsig^*)}.
\]
By Lemma~\ref{lem:FOScondition}, the first-order stability condition holds with
\[
\gamma = \frac{1-\epsilon/2}{\lambda_{\min}(\bsig^*)}.
\]
Since $\gamma < \lambda$ for any $0 < \epsilon \le 1$, 
\cite[Theorem 4]{balakrishnan2017statistical} implies that the EM operator $M(\cdot)$ is a contraction mapping on $\mathbb{B}_2(r;\bmu^*)$. 
In particular,
\begin{align}
\| M(\bmu) - \bmu^* \|_2 
\le (1 - \epsilon/2) \|\bmu - \bmu^*\|_2,
\quad \forall \bmu \in \mathbb{B}_2(r;\bmu^*).
\end{align}

Applying this inequality recursively, we obtain for any initialization $\bmu^{(0)} \in \mathbb{B}_2(r;\bmu^*)$,
\[
\|\bmu^{(t)} - \bmu^*\|_2 
\le (1 - \epsilon/2)^t \|\bmu^{(0)} - \bmu^*\|_2,
\]
which establishes the linear convergence of the population EM sequence.
\halmos \endproof


\subsection{Concavity}\label{app:prof_concavity_A}
\proof{Proof of Lemma ~\ref{lem:concavity_wrt_A}.}
Firstly, consider the scalar function
\begin{equation*}
    g(z)=\log\sigma(z),
\end{equation*}
where $\sigma(z)=(1+e^{-z})^{-1}$ is the sigmoid function. 
Since $\sigma'(z)=\sigma(z)(1-\sigma(z))$,
we have
\begin{equation*}
g'(z)=\frac{\sigma'(z)}{\sigma(z)}=1-\sigma(z),
\end{equation*}
and
\begin{equation*}
g''(z)=-\sigma(z)(1-\sigma(z))\le0.
\end{equation*}
Therefore $g(z)$ is concave in $z$.

Then observe that the utility difference $\Delta U_{nj}(\bm v;\bA)$ is affine in $\bA$. In particular,
\begin{equation*}
\frac{\partial \Delta U_{nj}(\bm v;\bA)}{\partial\bA}
=
(1-0^{c_n})\,\bm x_{nc_n}\bm x_{nc_n}^\top
-
\left[0^{c_n}+(1-0^{c_n})(1-0^{j-c_n})\right]\bm x_{nj}\bm x_{nj}^\top ,
\end{equation*}
which does not depend on $\bA$. Hence $\Delta U_{nj}(\bm v;\bA)$ is an affine function of $\bA$.

Since $g(z)$ is concave and $\Delta U_{nj}(\bm v;\bA)$ is affine in $\bA$,the composition rule for concave functions implies that
\begin{equation*}
\log\sigma\left(\frac{\Delta U_{nj}(\bm v;\bA)}{\lambda}\right)
\end{equation*}
is concave in $\bA$.
Since sums and expectations preserve concavity, the objective $\hat {Q'}(\bm\theta|\bm\theta^{(t)})$ is concave in $\bA$.\halmos
\endproof

\subsection{Gradient}\label{app:proof_gradA}
\proof{Proof of Theorem ~\ref{thm:grad_A}}
Consider the Monte Carlo approximation of the smoothed objective
\begin{equation*}
\hat Q'(\bm\theta|\bm\theta^{(t)})
=\sum_{n=1}^{N}\sum_{l=1}^{L}\overline w_{nl}
\left(
\log f(\bm v_n^{(l)}|\bmu,\bsig)+\sum_{j=1}^{J_n}\log
\sigma\left(
\frac{\Delta U_{nj}(\bm v_n^{(l)};\bA)}{\lambda}
\right)
\right).
\end{equation*}
When differentiating with respect to $\bA$, the Gaussian log-likelihood term 
$\log f(\bm v_n^{(l)}|\bmu,\bsig)$ does not depend on $\bA$ and therefore vanishes. 
Hence only the sigmoid contribution needs to be differentiated.

Using the identity
\begin{equation*}
\frac{d}{dz}\log\sigma(z)=1-\sigma(z),
\end{equation*}
and applying the chain rule yields
\begin{equation*}
\frac{\partial}{\partial \bA}\log\sigma\left(
\frac{\Delta U_{nj}(\bm v_n^{(l)};\bA)}{\lambda}
\right)
=\frac{1}{\lambda}\left(
1-\sigma\left(
\frac{\Delta U_{nj}(\bm v_n^{(l)};\bA)}{\lambda}
\right)
\right)
\frac{\partial \Delta U_{nj}(\bm v_n^{(l)};\bA)}{\partial \bA}.
\end{equation*}

Substituting this derivative into the objective and exchanging summation with differentiation gives
\begin{equation*}
\nabla_{\bA}\hat {Q'}
=
\frac{1}{\lambda}
\sum_{n=1}^{N}\sum_{j=1}^{J_n}
\left(
\frac{\partial \Delta U_{nj}(\bA)}{\partial \bA}
\sum_{l=1}^{L}\overline{\omega_{nl}}
\left(
1-\sigma\left(
\frac{\Delta U_{nj}(\bm v_n^{(l)};\bA)}{\lambda}
\right)
\right)
\right).
\end{equation*}

Finally, since $\Delta U_{nj}(\bm v;\bA)$ is affine in $\bA$, its derivative admits the closed form stated in the theorem. This proves the theorem.\halmos
\endproof